\documentclass[10pt,journal,compsoc]{IEEEtran}
\usepackage{amsthm}
\usepackage{thmtools}
\usepackage{mathtools}
\usepackage{bbm}
\usepackage{color}
 \usepackage{graphicx}
\usepackage{booktabs}
\usepackage{threeparttable}
\usepackage{makecell}
\usepackage{wrapfig}
\usepackage{longtable}
\usepackage{caption}
\usepackage{subcaption}
\usepackage{array}
\usepackage{comment}
\usepackage{wrapfig}
\usepackage{longtable}
\usepackage{supertabular}
\usepackage[normalem]{ulem}
\usepackage{amsmath}
\usepackage{amssymb}
\usepackage{hyperref}       
\usepackage{url}            
\usepackage{booktabs}       
\usepackage{amsfonts}       
\usepackage{nicefrac}       
\usepackage{microtype}      
\usepackage{float}
\usepackage{multirow}
\usepackage{multicol}
\usepackage{enumitem}
\usepackage[table]{xcolor}

\usepackage[algo2e,ruled,linesnumbered,norelsize,boxed]{algorithm2e}
\usepackage{subcaption} 
\usepackage{multirow} 

\newtheorem{theorem}{Theorem}

\theoremstyle{definition}

\DeclareTextFontCommand{\textbf}{\bfseries}
\DeclareTextFontCommand{\textit}{\itshape}

\def\argmin{\text{arg\,min }}
\def\Rad{\mathfrak{R}}

\ifCLASSOPTIONcompsoc
  \usepackage[nocompress]{cite}
\else
  \usepackage{cite}
\fi
\ifCLASSINFOpdf
\else
\fi
\begin{document}
%
\title{Promoting Generalization for Exact Combinatorial Solvers via Adversarial Instance Augmentation}
%
%
%
%

\author{Haoyang~Liu,
        Jie~Wang*,~\IEEEmembership{Senior Member,~IEEE,}
        Yufei~Kuang,
        Xijun~Li,
        Yongdong~Zhang,~\IEEEmembership{Senior Member,~IEEE,}
        and~Feng~Wu,~\IEEEmembership{Fellow,~IEEE}
\IEEEcompsocitemizethanks{
\IEEEcompsocthanksitem H. Liu, J. Wang, Y. Kuang, X. Li, Y. Zhang, and F. Wu are with: a) CAS Key Laboratory of Technology in GIPAS, University of Science and Technology of China, Hefei 230027, China; b) Institute of Artificial Intelligence, Hefei Comprehensive National Science Center, Hefei 230091, China.
E-mail: dgyoung@mail.ustc.edu.cn, ,yfkuang@mail.ustc.edu.cn, jiewangx@ustc.edu.cn, zhyd73@ustc.edu.cn, fengwu@ustc.edu.cn.
\IEEEcompsocthanksitem X. Li is with Shanghai Jiaotong University.
E-mail: lixijun@sjtu.edu.com.

Manuscript received October, 2023. * Corresponding author.
}
}

%
%

\markboth{Journal of \LaTeX\ Class Files,~Vol.~14, No.~8, August~2015}%
{Shell \MakeLowercase{\textit{et al.}}: Bare Demo of IEEEtran.cls for Computer Society Journals}
%



\IEEEtitleabstractindextext{%
\begin{abstract}
    Machine learning has been successfully applied to accelerate Mixed-Integer Linear Programming (MILP) solvers.
    However, the learning-based solvers often suffer from severe performance degradation on unseen MILP instances due to the limited number of training instances and diversity of training distributions.
    To tackle this problem, we propose a novel data augmentation approach, called \textbf{Ad}versarial Instance \textbf{A}ugmentation, to promote the data diversity for learning-based branching modules in branch-and-bound (B\&B) \textbf{Solver}s (\textbf{AdaSolver}).
    To generate diverse augmented instances, 
    AdaSolver uses an augmentation policy to augment the structures of the bipartite graphs representing MILP instances. 
    The major technical contribution is that AdaSolver formulates the \textit{non-differentiable} learning problem of the augmentation policy as a contextual bandit problem, enabling efficient \textit{gradient-based} adversarial training for the learning-based solver and augmentation policy. 
    To the best of our knowledge, AdaSolver is the \textit{first} general and effective framework for understanding and improving the generalization of both {imitation-learning-based} and {reinforcement-learning-based} B\&B solvers.
    Experiments demonstrate that AdaSolver leads to a remarkable roughly 35\% improvement in solving time over the B\&B solver across various distributions.
    Moreover, AdaSolver significantly improves the sample efficiency by roughly 40\% reduction in solving time compared to the competitive GNN branching baseline with only 1\% of the training instances.
\end{abstract}

}

\maketitle

\IEEEdisplaynontitleabstractindextext

%
\IEEEpeerreviewmaketitle

\IEEEraisesectionheading{\section{Introduction}\label{sec:introduction}}

%
%
%
%
\IEEEPARstart{C}{ombinatorial} optimization (CO) problem is one of the most fundamental optimization problems that widely originates from scheduling \cite{chen2010integrated}, chip design \cite{ma2019accelerating}, route planning \cite{liu2008tsp}, and other industrial scenarios.
Solving CO problems is often time-consuming and computationally intensive due to their NP-hard nature.
In practice, a large family of CO problems can be formulated as Mixed-Integer Linear Programming (MILP).
Exact MILP solvers, such as the B\&B solvers SCIP \cite{bestuzheva2021scip} and Gurobi \cite{gurobi2021gurobi},  typically rely on custom heuristics that require extensive manual tuning and complex workflows \cite{colorni1996heuristics, hromkovivc2013algorithmics}.
To enhance solver efficiency, recent research has leveraged fast-inference learning-based models to approximate complex heuristics \cite{he2014learning, khalil2016learning, gasse2019exact} or to discover more effective heuristics within existing solvers \cite{NEURIPS2022_756d74cd, wanglearning}. 
These approaches have shown promising improvements in in-distribution efficiency.

However, generalizing learning-based solvers to out-of-distribution (OOD) instances presents significant challenges for real-world applications.
\textbf{First, these solvers are typically trained on a specific dataset, yet they must handle instances of varying sizes.}
The sizes of instances encountered in practice can be much larger than those in the training dataset.
\textbf{Second, environmental changes frequently introduce perturbations to instance structures.}
For instance, in staff scheduling problems, unexpected emergencies may arise, imposing additional constraints that were not considered in previous MILP formulations. 
Consequently, we must account for new types of constraints that reflect these restrictions, altering the problem structures.
\textbf{Third, acquiring or generating additional instances from different distributions or of larger sizes may be impractical due to concerns about information security and privacy, as well as high data acquisition costs } \cite{bengio2021machine, Sakuma2007AGA}.
In scenarios like job-shop scheduling within a company, sensitive business information---such as production capacity and costs---can be inferred from MILP instances. 
As a result, companies may only provide a limited number of anonymous training instances to mitigate the risk of privacy breaches.

The solving time for learning-based solvers in the aforementioned contexts can be considerably longer than that of traditional heuristics (Table \ref{table:performance}), particularly when training instances are limited (Table \ref{table:sample_efficiency_under_less_data}).
Existing research on the generalization capabilities of CO solvers primarily targets problem-specific approximation solvers, such as neural solvers for routing issues, rather than general exact MILP solvers.
For the problem-specific solvers, we can get access to the problem formulation and then employ domain-randomization type approaches by generating a large number of instances from different distributions \cite{wang2023asp, geisler2021generalization}.
However, data generation faces significant challenges in cases where prior knowledge of the problem formulation is unavailable, especially with anonymous datasets. 
Consequently, problem-specific generation algorithms are not applicable in such situations.
This paper concentrates on the restrictive practical setting where training data is drawn from a limited number of domains, making it infeasible to acquire or generate new instances.

To address the aforementioned challenges, we propose \textbf{Ad}versarial Instance \textbf{A}ugmentation for \textbf{Solver} (\textbf{AdaSolver}), a robust optimization framework for learning-based B\&B solvers.
Specifically, we focus on the competitive graph neural network (GNN) approach for learning branching heuristics (please see Section \ref{section:branching} for more details on branching). 
First, AdaSolver represents each MILP instance as a bipartite graph and promotes the diversity of graph structures in training distributions without requiring additional training instances or prior knowledge of problem types.
The framework comprises two main components: a GNN branching policy for selecting branching variables and an adversarial augmentation policy that modifies bipartite graph structures to simulate real-world perturbations.
Second, we formulate the non-differentiable learning problem of the augmentation policy as a contextual bandit problem, which offers two key advantages. (1) The contextual bandit formulation facilitates gradient-based adversarial training for both the branching and augmentation policies.
(2) We introduce an efficient training method adapted from \cite{DBLP:journals/corr/SchulmanWDRK17}, enabling rapid learning with a limited number of instances.
Finally, to ensure that augmented instances remain within acceptable deviations from the training distributions, we employ a discriminator network to filter out unfavorable instances that could disrupt model training.
We apply the AdaSolver framework to both imitation-learning-based (IL-based) and reinforcement-learning-based (RL-based) solvers, resulting in two variants: AdaSolver-IL and AdaSolver-RL.

We evaluate our approach through experiments conducted on four synthesis benchmarks and a real-world dataset featuring large instance sizes, comparing it against baseline methods under both in-distribution and out-of-distribution testing conditions. Our experiments highlight several advantages of AdaSolver. 
(1) \textbf{High generalization ability}: 
AdaSolver achieves an impressive approximately 35\% reduction in solving time compared to the B\&B solver across various distributions. It consistently outperforms both IL-based and RL-based solvers on instances with sizes ranging from five to ten times larger than those in the training set.
(2) \textbf{High sample efficiency}: With just 1\% of the training instances, AdaSolver significantly enhances sample efficiency, resulting in about a 40\% reduction in solving time compared to the competitive GNN branching baseline. This allows for effective training of a branching policy using far fewer training instances while maintaining comparable performance.
To the best of our knowledge, AdaSolver is the first general framework designed to enhance the robustness and generalization capabilities of both IL- and RL-based B\&B solvers in real-world settings.

\section{Related Work}

\subsection{Machine Learning for Combinatorial Optimization}
Learning to optimize combinatorial optimization (CO) problems has garnered increasing interest in the fields of operations research and machine learning \cite{bengio2021machine, lodi2017learning}. 
Existing methods can be broadly categorized into two groups.
\textbf{The first group focuses on solving specific CO problems in an end-to-end manner}. 
Examples include the Traveling Salesperson Problem (TSP) \cite{vinyals2015pointer, bello2016neural, kool2018attention}, the Satisfiability Problem (SAT) \cite{selsam2018learning, yolcu2019learning}, the Vehicle Routing Problem (VRP) \cite{kool2018attention, nazari2018reinforcement}, and the Maximum Cut Problem (MCP) \cite{barrett2020exploratory}.
\textbf{The second group employs learning-based methods to approximate heuristics within exact B\&B solvers}.
Examples include branching \cite{khalil2016learning, gasse2019exact,gupta2020hybrid,gupta2022lookback}, node selection \cite{he2014learning, NEURIPS2022_cf5bb188}, and cut selection \cite{tang2020reinforcement,wanglearning}.
Built on exact solvers, these approaches offer optimal guarantees and can be applied to solve general MILP instances.
This paper specifically focuses on the branching module within B\&B solvers, as the computational efficiency and quality of selected variables during branching significantly affect the overall performance of the B\&B algorithm \cite{achterberg2007constraint}.

Recently, researchers have paid more attention to the generalization and robustness of end-to-end neural solvers. 
For example, various studies have enhanced the generalization of these solvers through techniques like curriculum learning \cite{wang2023asp}, group distributionally robust optimization \cite{2022RoutingDRO}, knowledge distillation \cite{bi2022learning}, and generative adversarial training \cite{xin2022generative}, along with others \cite{geisler2021generalization}. 
However, these methods often require prior knowledge of the problem formulation for generation, and many are primarily tailored to routing problems, which have low-cost instance generation processes.
Some other works, such as \cite{pmlr-v162-duan22b}, propose label-preserving augmentations specifically for SAT problems.
Nonetheless, in practical scenarios where instances are anonymous or difficult to generate, these approaches are not applicable to exact B\&B solvers.
For exact CO solvers, \cite{luroco} introduces an adversarial attack approach to evaluate their robustness. 
Additionally, \cite{Zarpellon_Jo_Lodi_Bengio_2021} seeks to enhance generalization across heterogeneous MILPs in the imitation learning setting. 
However, the design of robust learning-based exact solvers for both IL and RL remains largely unexplored.

\subsection{Machine Learning for Branching}
Existing works on learning to branch can be categorized into two main types: imitation-learning-based (IL-based) and reinforcement-learning-based (RL-based) algorithms \cite{Scavuzzo2024}.
\emph{IL-based methods} assume the presence of an accurate but expensive branching expert and use a machine learning model to replicate the expert's branching strategy.
For example, \cite{khalil2016learning} employs support vector machines (SVM) to imitate a strong branching expert ({see Appendix \ref{section:branching_rules} for details on strong branching}). Other IL-based approaches, such as those in \cite{gasse2019exact} and \cite{nair2020solving}, represent branching states as bipartite graphs and utilize graph neural networks (GNNs) to enhance branching performance. Additionally, \cite{Zarpellon_Jo_Lodi_Bengio_2021} proposes a TreeGate model to enhance the tree search process.
\cite{nair2020solving} uses data augmentation to enhance the solving performance of MILPs. 
\emph{RL-based methods}, on the other hand, do not depend on a predefined branching expert. Instead, they automatically explore branch-and-bound trees to improve heuristics. 
For instance, \cite{10.1007/978-3-030-58942-4_12} is one of the first to apply reinforcement learning for variable selection. Other studies have introduced various techniques to enhance the performance of RL-based branching policies, including evolution strategies \cite{sun2020improving}, retrospective branching trajectories \cite{parsonson2022retro}, an advanced Q-network trained with mixed data \cite{qu2022improved}, and tree Markov decision processes \cite{NEURIPS2022_756d74cd}.

\section{Background}
\subsection{Branch-and-Bound Algorithm}
\label{section:branching}
In practice, most of the CO problems can be formulated as Mixed-Integer Linear Programming (MILP) in the form of
\begin{align}\label{MILP}
 \underset{{\mathbf{x}}\in\mathbb{R}^n}{\min}\,{\mathbf{c}}^\top {\mathbf{x}}, \text{    s.t.    }{\mathbf{A}\mathbf{x}}\le{\mathbf{b}}, {\mathbf{l}}\le{\mathbf{x}}\le{\mathbf{u}},\text{    }{\mathbf{x}}\in\mathbb{Z}^p\times\mathbb{R}^{n-p},
\end{align}
where ${\mathbf{c}}\in\mathbb{R}^n$, $\mathbf{A}\in \mathbb{R}^{m\times n}$, ${\mathbf{b}}\in\mathbb{R}^m$, ${\mathbf{l}}\in(\mathbb{R} \cup\{-\infty\})^{n}$ and ${\mathbf{u}}\in(\mathbb{R} \cup\{+\infty\})^{n}$ denote the lower and upper bounds of the variables.
The branch-and-bound (B\&B) algorithm solves MILP instances with a divide-and-conquer approach.
B\&B recursively performs the following steps.
First, it solves the linear relaxation of a selected subproblem, starting with the original MILP.
It identifies an integer variable $x_i$ with fractional value $x_i^*$ in the current LP solution and generates two subproblems by adding constraints $x_i\le \lfloor x_i^*\rfloor, x_i\ge \lceil x_i^*\rceil$ to the current subproblem, respectively.
B\&B then chooses a new subproblem to explore next.
As the algorithm proceeds, the subproblems form a binary search tree, where the root node represents the original MILP and the left and right children of a subproblem are two new subproblems created in the previous step.
The corresponding subtree is pruned if a subproblem is infeasible or does not contain optimal solutions.

The efficiency of the B\&B algorithm is greatly influenced by the selection of a fractional variable $x_i$, also known as the branching process.
The strong branching policy \cite{applegate1995finding} has been shown to produce the smallest B\&B trees among existing heuristics.
Strong branching selects a branching variable by evaluating the expected bound improvement for each candidate variable.
For more information on branching policy, please refer to Appendix \ref{section:branching_rules}.

\subsection{Instance Bipartite Graph and Branching Samples}
We represent each MILP instance (\ref{MILP}) as a weighted bipartite graph $G = (W\cup V, E)$.
In this representation, the vertex sets $W$ and $V$ correspond to the constraint set and variable set of the MILP, respectively.
Each node in $W$ or $V$ is associated with features $\mathbf{w}$ or $\mathbf{v}$, which contain relevant information about the corresponding constraint or variable.
The adjacency matrix $E$ corresponds to the constraint matrix $\mathbf{A}$. The collection of all such bipartite graphs is denoted as $\mathcal{G}$ and referred to as \textit{instance bipartite graphs}.

Given a MILP instance, the B\&B solver recursively solves a sequence of subproblems, each of which is also a MILP. 
Consequently, we can represent the subproblem at time step $t$ as a bipartite graph analogous to the original MILP instances, referred to as \emph{branching samples} $G_t$.

\begin{figure}[H]
  \centering
  \includegraphics[width=0.4\textwidth]{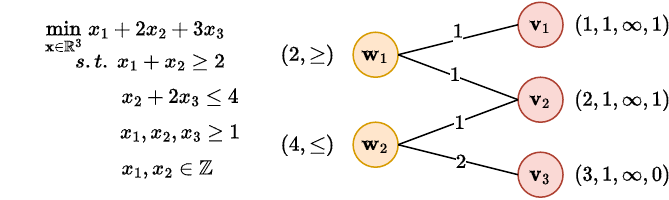}
  \caption{A simple example of an MILP instance bipartite graph. 
  }\label{bipartite_graph}
    \vspace{-3mm}
\end{figure}

\section{IL- and RL-based Branching Formulation}
In practice, we have a finite set of training instances $\{G^{(i)}\}_{i=1}^n$ drawn from the training distribution $p_{\text{tr}}(G)$, while the testing instances come from a different, unseen distribution $p_{\text{te}}(G)$.
The sequential branching decisions made during the B\&B process can be formulated as a Markov decision process (MDP) \cite{he2014learning}, where the solver serves as the environment and the branching module acts as the agent.

We consider the solving process as a finite-horizon MDP, called \textbf{branching MDP}, with a maximum time horizon $T$. 
This MDP is represented by the tuple $(\mathcal{G}, \mathcal{A},\mathcal{T}, R)$, where $\mathcal{G}$ denotes the state space of branching samples, $\mathcal{A}$ represents the action space consisting of candidate branching variables, and $R$ is the reward function.
{ Following \cite{prouvost2020ecole}, the reward function $R$ is defined as the negative primal-dual gap at the corresponding time step, multiplied by a sampled duration. 
The accumulated reward is defined as the negative primal-dual integral (please see Appendix \ref{appendix:PD_metric} for definition).
}
The transition function $\mathcal{T}:\mathcal{G}\times\mathcal{A}\to\mathcal{G}$ is deterministic since the branching process rigorously follows the universal mathematical rules.
A policy $\pi\in\Pi:\mathcal{G}\to\mathcal{A}$ maps an observed state to an action based on the corresponding branching strategy.

At time $t$, the branching module observes a branching sample $G_t\in\mathcal{G}$ corresponding to the current subproblem. 
It selects a fractional variable as action $a_t$ from the branching candidates $\mathcal{A}$ according to the branching policy $\pi$. 
Subsequently, the solver splits the current subproblem using $a_t$ to generate two new subproblems and determines the next subproblem as $G_{t+1}=\mathcal{T}(G_t,a_t)$ for further branching.
The distribution of branching samples obtained by executing the policy $\pi$ on instance $G$ drawn from distribution $p$ is denoted as $p(G_{1:T}|\pi)$, 
{ 
\begin{align*}
  p(G_{1:T}|\pi) = p(G)p(G_1|G,\pi(G))\prod_{t=1}^{T-1}p(G_{t+1}|G_t, \pi(G_t)).
\end{align*}
}
The distribution $p(G_{1:T}|G, \pi)$ represents the marginal distribution of branching samples given an instance $G$,
{ 
\begin{align*}
  p(G_{1:T}|G,\pi) = p(G_1|G,\pi(G))\prod_{t=1}^{T-1}p(G_{t+1}|G_t, \pi(G_t)).
\end{align*}
}
In implementation, the learned branching policy is a parameterized function $\pi_\theta$ with parameters $\theta$.

\subsection{IL-based branching formulation}
To solve the branching MDP mentioned above, we can employ imitation learning or reinforcement learning to train the branching agent.
IL-based branching utilizes a strong branching expert with policy $\pi_E$, assumed to be near-optimal.
We obtain a dataset of branching samples and branching variables $\mathcal{D}_{\text{IL}}=\{(G_t^{(i)}, \pi_E(G_t^{(i)}))\}$ by running the expert on the training instances $\{G^{(i)}\}_{i=1}^n$.
The goal of IL is to replicate the expert's behavior by training the agent to make decisions consistent with those of the expert under the same observations.
The optimization objective can be formulated as follows
\begin{align}\label{formulation:il}
  \min_{\pi_\theta} \mathcal{J}_{p_{\text{tr}}}(\pi_\theta)=\mathbb{E}_{G_t\sim p_{\text{tr}}(G_{1:T}|\pi_E)} \left[\ell(\pi_\theta(G_t),\pi_E(G_t))\right]
\end{align}
where $\ell(\cdot, \cdot)$ is the loss function \cite{gasse2019exact} taking the form of 
{

\begin{equation}
    \begin{aligned}      
\ell(\pi_\theta(G_t),&\pi_E(G_t))= \sum_{a\in \mathcal{A}}-\mathbbm{1}_{a=\pi_E(G_t)}\log p_{\theta}(a|G_t)\\
&-(1-\mathbbm{1}_{a=\pi_E(G_t)}) \log (1-p_{\theta}(a|G_t)),
\end{aligned}
\end{equation}

where the probability $p_\theta$ is the output logit of labeled action $\pi_E(G_t)$ for policy $\pi_{\theta}$, given the input branching sample $G_t$.
The indicator function $\mathbbm{1}_{a=\pi_E(G_t)}$ takes value 1 when $a=\pi_E(G_t)$ and 0 otherwise.
This loss function aims to maximize the predicted probability of the ground-truth branching variables.
}
{ where the probability $p_\theta$ is the output logit of labeled action $\pi_E(G_t)$ for policy $\pi_{\theta}$, given the input branching sample $G_t$.
The loss function maximizes the predicted probability of the ground-truth branching variables.}

\subsection{RL-based branching formulation}
In the RL setting, we do not require an expert to collect branching samples. Instead, the training dataset consists of the instance set $\mathcal{D}_{\text{RL}}=\{G^{(i)}\}_{i=1}^n$.
The branching agent explores the B\&B tree for branching samples using the current policy $\pi_\theta$ and aims to improve the policy by minimizing the negative cumulative reward, 
\begin{align}\label{formulation:rl}
  \min_{\pi_\theta} \mathcal{J}_{p_{\text{tr}}}(\pi_\theta)=\mathbb{E}_{G_t\sim p_{\text{tr}}(G_{1:T}|\pi_\theta)} \left[\sum_{t=0}^{T-1}-\gamma^t R(G_t)\right],
\end{align}
where the reward $R$ is defined as the primal-dual integral (see Appendix \ref{appendix:PD_metric} for the definition).

We summarize the optimization objectives for both the IL and RL settings as follows, 
{
\begin{align*}
&\mathcal{J}_{p_{\text{tr}}}(\pi_\theta)=\begin{cases}
    \mathbb{E}_{{G}_t\sim p_{\text{tr}}({G}_{1:T}|\pi_E)} \left[\ell(\pi_\theta({G}_t),\pi_E({G}_t))\right] & {\text{(Equ \ref{formulation:il})}}\\
    \mathbb{E}_{{G}_t\sim {p_{\text{tr}}}({G}_{1:T}|\pi_\theta)} \left[\sum_{t=0}^{T-1}-\gamma^t R({G}_t)\right] & {\text{(Equ \ref{formulation:rl})}}
\end{cases}
\end{align*}
}
For simplification, we use $F(\pi,G)$ as a unified notation
{
\begin{align*}
    F(\pi_\theta,G)=\begin{cases}
    \mathbb{E}_{G_t\sim {p}(G_{1:T}|G,\pi_E)} \left[\ell(\pi_\theta(G_t),\pi_E(G_t))\right] & \text{(IL)}\\
    \mathbb{E}_{G\sim {p}(G_{1:T}|G,\pi_\theta)} \left[\sum_{t=0}^{T-1}-\gamma^t R(G_t)\right] & 
    \text{(RL)}
    \end{cases}
\end{align*}
}
Thus, we have the overall objective
\begin{align*}
    J_p(\pi) = \mathbb{E}_{G\sim p}F(\pi, G).
\end{align*}


\subsection{Definitions on the optimal policy}
We first define the optimal policy $\pi^*$ for the testing distribution, which minimizes the objective $\mathcal{J}_{p_{\text{te}}}(\pi)$ within the hypothesis policy class $\Pi$:
\begin{align}\label{equ_optimal_policy}
  \pi^*=\underset{\pi_{\theta}\in\Pi}{\argmin}\mathcal{J}_{p_{\text{te}}}(\pi_{\theta}).
\end{align}
However, in practice, we only have access to a finite set of instances $\mathcal{D}_{\text{IL}}=\{(G_t^{(i)}, \pi_E(G_t^{(i)})\}$ for IL setting and $\mathcal{D}_{\text{RL}}=\{G^{(i)}\}$ for RL setting, both sampled from the training distribution.
We estimate the objective using empirical risks to derive the best policy $\hat{\pi}$ on the dataset
\begin{align}
   \hat{\pi}=\underset{\pi_{\theta}\in\Pi}{\argmin}\hat{\mathcal{J}}_{p_{\text{tr}}}(\pi_{\theta}).
\end{align}
We use data augmentation to improve models' generalization ability.
We denote the distribution of augmented instances by $\tilde{p}$ and the best policy on the augmented dataset by $\tilde{\pi}$,
\begin{align}
  \tilde{\pi}=\underset{\pi_{\theta}\in\Pi}{\argmin}\hat{\mathcal{J}}_{\tilde{p}}(\pi_{\theta}),
\end{align}
in which $\hat{\mathcal{J}}_{\tilde{p}}(\pi)$ is the empirical risks calculated using the augmented instances.

\section{AdaSolver}
\begin{figure*}[t]
  \centering
  \includegraphics[width=\textwidth]{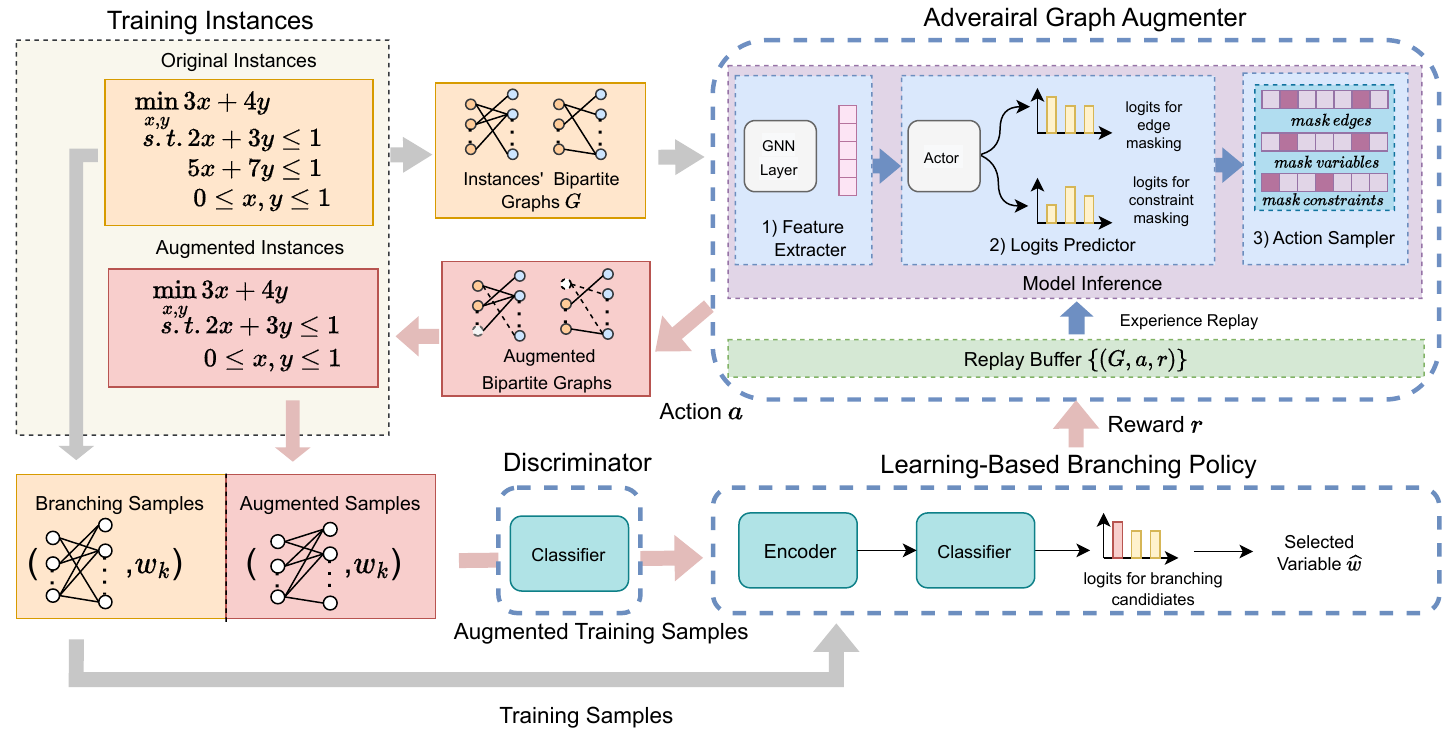}
  \caption{Overview of AdaSolver framework.
  AdaSolver jointly optimizes a learning-based branching policy and an adversarial graph augmentation policy.
  In each training epoch, we transform a subset of MILP instances into instance bipartite graphs and send them to the augmentation policy. Then the augmentation policy decides which nodes and edges to mask. After that, we transform the augmented graphs back to instances.
  We also leverage a discriminator for data selection.
  To train the branching policy, we collect branching samples from the original and selected augmented MILP instances. The branching policy records the average loss over samples generated from each augmented instance and returns the loss to compute reward signals for the augmentation policy. For training efficiency, we use a replay buffer to perform experience replay for the augmentation policy.
  }\label{AdaSolver framework}
    \vspace{-3mm}
\end{figure*}

In this section, we introduce the robust optimization framework for learning-based exact MILP solvers using adversarial instance augmentation, namely \textbf{Ad}versarial Instance \textbf{A}ugmentation for the exact MILP \textbf{Solver} (\textbf{AdaSolver}).
The overview of the proposed framework on the learn-to-branch task is presented in Figure \ref{AdaSolver framework}, which consists of two components: a learning-based branching policy and an adversarial augmentation policy.
In Section \ref{section: derivation}, we outline the derivation for instance augmentation. Section \ref{section: operators} describes the augmentation operations applied to the instance graphs. Sections \ref{section: contextual bandit} and \ref{section: discriminator} detail the adversarial augmentation process.
The adversarial augmentation policy utilizes data augmentation to enhance the diversity of graph structures within the training data. Simultaneously, the learning-based branching policy network aims to minimize training loss by leveraging branching samples from both original and augmented instances, thereby extracting task-relevant features across different distributions.

\subsection{Robust Optimization and Adversarial Augmentation}
\label{section: derivation}
Given a branching policy $\pi$ and a training instance distribution $p_{\text{tr}}(G)$, we formulate the distributionally robust training process as the following optimization problem
\begin{equation}
\label{equ:robustopt}
\begin{aligned}
& \min_{\pi_\theta} \sup_{\tilde{p}}\mathcal{J}_{\tilde{p}}(\pi_\theta)\\
& \text{ s.t. } \mathcal{J}_{\tilde{p}}(\pi_\theta)=\mathbb{E}_{\tilde{G}\sim \tilde{p}}F(\pi_\theta, \tilde{G})
\\
& \qquad\mathcal{W}_{\mathcal{G}}(p_{\text{tr}},\tilde{p})\le \rho.
\end{aligned}
\end{equation}
Here, $\tilde{p}$ is the augmentation distribution, and $\mathcal{W}_{\mathcal{G}}(p_{\text{tr}},\tilde{p})$ is the Wasserstein distance that measures the distance between two distributions $p_{\text{tr}}(G)$ and $\tilde{p}(G)$.
Specifically, we define $$\mathcal{W}_{\mathcal{G}}(p_{\text{tr}},\tilde{p}) =\inf_{\mu}\mathbb{E}_{\mu}\left[c(G,\tilde{G})\right],$$ 
where $\mu$ is the coupling distribution of $G$ and $\tilde{G}$ with marginal distribution $p_{\text{tr}}(G)$ and {$\tilde{p}(\tilde{G})$}.
The cost function $c:\mathcal{G}\times\mathcal{G}\to\mathbb{R}$ measures the discrepancy between two instance graphs or branching samples.
{
In general, the cost function can represent any distance metric within the graph space. 
Following \cite{aia}, the cost function in the representation space corresponds to semantic distances. 
In our implementation, the learned branching policy is represented as a parameterized function $\pi$. We decompose
the branching policy as $\pi = f\circ h$, where $h : \mathcal{G} \to \mathbb{R}^d$ is the graph encoder {with embedding size $d$}, and $f$ is the decoder.
We define the cost function in this representation space as 
\begin{align}c(G, \tilde{G})=\|h(G)-h(\tilde{G})\|_2,\end{align}
which represents the augmentation cost from instance $G$ to $\tilde{G}$.
}

Since computing the supremum over the distributions is intractable within the optimization problem (\ref{equ:robustopt}), we employ a Lagrangian relaxation with a penalty parameter $\beta$:
\begin{align}\label{equ:fomulation}
  &\min_{\pi_\theta} \sup_{\tilde{p}} \mathbb{E}_{\tilde{G}\sim \tilde{p}(G)}F(\pi_\theta,\tilde{G})-\beta\mathcal{W}_{\mathcal{G}}(p_{\text{tr}},\tilde{p}).
\end{align}
To solve the optimization problem (\ref{equ:fomulation}), 
\cite{DBLP:conf/nips/VolpiNSDMS18} and \cite{aia} introduce a surrogate loss function such that 
\begin{align}
\nabla_{\theta}\sup_{\tilde{G}\in\mathcal{G}} [F(\pi_\theta,G)-\beta c(G,\tilde{G})]=\nabla_{\theta}F(\pi_\theta,G^*),
\end{align}
where the instance $G^*$ satisfies 
$$G^*=\text{argmax}_{\tilde{G}\in\mathcal{G}}[F(\pi_{\theta},\tilde{G})-\beta c(G,\tilde{G})].$$
The instance $G^*$ can be considered the worst-case augmented instance for the branching policy in the surrogate function. This implies that computing the gradient only requires evaluating the objective on the augmented instances $G^*$.
Therefore, we focus on developing an effective and efficient augmentation method for MILP instances.

\subsection{Instance Graph Augmentation Operators}
\label{section: operators}
Since the MILP problem can be represented by instance graphs, we propose to augment the instances by masking variable nodes, constraint nodes, and edges in instance bipartite graphs to promote the diversity of graph structures.
These augmentation operators are aligned with real-world scenarios. 
For example, many MILP problems arise from real-world scheduling scenarios, where variables represent resources such as workers, machines, or vehicles, and constraints capture scheduling relationships among these resources.
By using simple and reliable augmentation techniques, we can effectively simulate real-world resource perturbations, such as the inaccessibility of certain materials or the relaxation of specific constraints.
Furthermore, recent studies, including \cite{chen2022representingmilp}, have highlighted the challenges faced by graph neural networks (GNNs) when processing certain ``foldable" MILPs, which are common in real-world applications and exhibit significant structural symmetry.
Specifically, GNNs struggle to distinguish between two different ``foldable" MILPs such that much of the training data is ineffective for training the branching policy. 
This issue further exacerbates the challenges posed by the limited diversity of training data.
By augmenting the graph structures, AdaSolver effectively disrupts the symmetric patterns present in training instances, thereby promoting greater diversity in instance graph structures. As a result, instance augmentation enhances both the generalization ability and sample efficiency of the branching policy, enabling comparable performance with fewer instances, as demonstrated in Section \ref{subsection:sample_efficiency}.

To mitigate any adverse effects on training, we carefully control the application of augmentation operators and adjust the masking proportions, ensuring that the augmented instances retain similar problem structures and computational complexities as the original instances.

\subsection{Contextual Bandit for Adversarial Augmentation}
\label{section: contextual bandit}
The instance augmentation policy, denoted as $\phi:\mathcal{G}\to\mathcal{G}$, maps an instance bipartite graph $G$ to the augmented instance graph $\tilde{G}:=\phi(G)$ by applying augmentation operators.
However, two significant challenges remain to be addressed in this process.
First, applying structural augmentation to a bipartite graph results in a non-differentiable learning objective for the augmentation policy, making it difficult to train using backpropagation and gradient-based updates. Second, the augmentation policy receives considerably less training data (instance graphs) compared to the branching policy (branching samples), leading to low training efficiency for the augmentation policy.
The adversarial augmentation policy operates on MILP instance graphs $\{G^{(i)}\}_i$ to generate augmented instances $\{\phi(G^{(i)})\}_i$, and each augmented instance can generate a sequence of branching samples. 
Therefore, we formulate the instance augmentation process as a contextual bandit problem and adopt a sample-efficient contextual-bandit version of the proximal policy optimization (PPO) algorithm to train the augmentation policy \cite{DBLP:journals/corr/SchulmanWDRK17}.

In this contextual bandit formulation, we treat the solver as the environment and the adversarial augmentation policy as the agent. The contextual bandit problem can be represented as the tuple $({\mathcal{G}}, \widetilde{\mathcal{A}}, \tilde{r})$.
Then we specify the state space ${\mathcal{G}}$, the action space $\widetilde{\mathcal{A}}$ and reward function $\tilde{r}:{{\mathcal{G}}}\times\widetilde{\mathcal{A}}\to\mathbb{R}$ for a MILP instance as follows.
\begin{itemize}
  \item \textbf{The state space} $\mathcal{G}$ is the set of instance graph representations of the MILP instances.
  \item
      \textbf{An action} $\tilde{a}=(\widetilde{V}, \widetilde{E}, \widetilde{W})\in\widetilde{\mathcal{A}}$, consists of subsets of variables $\widetilde{V}$, edges $\widetilde{E}$, and constraints $\widetilde{W}$ to be masked.
      The agent applies action $\tilde{a}$ to an instance graph $G$ to obtain an augmented instance $\phi(G)$, from which branching samples are collected.
  \item {\textbf{The reward} The augmentation reward function $\tilde{r}(G, \tilde{a})$ is defined as the increase in the average loss of the branching policy, given by $F(\pi_\theta,\phi(G))-F(\pi_\theta,G)$, on the branching samples from the augmented instances $\phi(G)$ and the original instances $G$, with two additional regularized terms. 
    The reward takes the form of 
  \begin{equation}
      \begin{aligned}
      \tilde{r}(G, \tilde{a}) =& F(\pi_\theta,\phi(G))-F(\pi_\theta,G)\\
      &-\beta c(G, \phi(G))-I_{\text{fea}}(\phi(G)),
  \end{aligned}
  \end{equation}
  where $\beta=0.01$ is a constant, $I_{\text{fea}}(\phi(G))$ is the indication function taking 0 when $\phi(G)$ is a feasible instance and $\infty$ when $\phi(G)$ is infeasible. 
  }
\end{itemize}
The training process under this contextual bandit formulation is described as follows. We implement an augmentation policy network $\phi_{\eta}:\mathcal{G}\to\mathcal{G}$ with parameters $\eta$. 
This network takes an instance graph $G$ as input and outputs the masking probability for each node and edge, denoted as $p_{\eta}(\tilde{a}|G)$.
An augmented instance is constructed by selecting the nodes and edges most likely to be masked based on the predicted probabilities and a predefined masking proportion.

\subsection{Ensure the Feasibility for Data Augmentation}
\label{section: discriminator}
The adversarial augmentation policy tends to generate instances that significantly differ from the original distribution, leading to some instances being either extremely easy or overly difficult to solve.
This occurs because the training objective of the augmentation policy focuses solely on maximizing the loss of the branching policy on augmented instances without imposing constraints on the augmentation process.
Consequently, the policy may overlook distributional information, resulting in instances that disrupt the underlying structure of the MILP problems.
As seen in Equation (\ref{equ:fomulation}), the cost function $c$ is calculated using embeddings from the optimal policy, which are not available during training. To address this, we approximate the cost function using the embeddings from the current policy, denoted as $h_{\theta_2}(G)$.
We introduce a auxiliary discriminator network $D = g_{\theta_3}\circ h_{\theta_2}$, which shares the encoder network $h_{\theta_2}$ with the policy network and uses a decoder $g_{\theta_3}$ with parameters $\theta_3$.

The graph encoder captures the structural information of the MILP instance, while the decoder predicts the probability that an instance belongs to the original distribution. This allows the decoder to identify and filter out instances that exhibit significant deviations from the original distribution. During training, we minimize the following objective to train the discriminator
\begin{align}\label{equ: discriminator}
  \min_{\theta_3} \mathcal{L}(\theta_3)=\mathbb{E}_{G\sim p_{\text{tr}}}[ D(\phi_\eta(G))] +\mathbb{E}_{G\sim p_{\text{tr}}}[1-D(G)].
\end{align}
When presented with an augmented instance graph, the discriminator decides to include the augmented instance in training the learning-based branching policy if $D(G)>0.5$; otherwise, it discards it. 
{ Additionally, we introduce the regularized reward for the branching policy
\begin{equation*}
    \begin{aligned}
  \tilde{r}(G) = &F(\pi_\theta,\phi_\eta(G))-F(\pi_\theta,G)\\
  &-\beta c(G, \phi_\eta(G))-I_{\text{fea}}(\phi_\eta(G))
\end{aligned}
\end{equation*}
 The cost function term serves to prevent the augmented instances from deviating significantly from the original instances in the representation space. 
The feasibility indication function acts as a strong regularization for feasibility. }

To improve the training efficiency of the augmentation policy, we collect tuples $p_{\eta}(\tilde{a}|G)$ from the history of the augmentation policy for experience replay. We represent the likelihood ratio between the current policy and the old policies sampled from the buffer as
\begin{align}
    \lambda_{\eta}(G)=\frac{p_{\eta}(\tilde{a}|G)}{p_{\eta_{\text{old}}}(\tilde{a}|G)}    
\end{align}
Finally, the training objective for the adversarial augmentation policy is minimizing the following equations:
\begin{equation}\label{PPOobj}
\begin{aligned}
&\mathcal{L}(\eta)=\mathbb{E}_{G\sim p_{\text{tr}}}\left[\min(\lambda_{\eta}(G)\tilde{r}, \text{clip}(\lambda_{\eta}(G),1-\epsilon, 1+\epsilon)\tilde{r}))\right]
\end{aligned}
\end{equation}
where $\epsilon$ is the clip ratio.
The PPO clips $\lambda_{\eta}(G)$ in the objective to avoid substantial policy updates.
The importance of the sample-efficient bandit algorithm in the overall training process is demonstrated in Table \ref{table:ablation}.

We provide the pseudo-code for the AdaSolver framework in Algorithm \ref{alg-training}.

\section{Theoretical Analysis for Adversarial Augmentation Framework}
\label{section:theorem}
In this section, we provide a detailed theoretical analysis of the proposed instance augmentation technique.
We analyze the generalization bound for data augmentation in both IL and RL settings.
We suppose that the augmented instances are drawn from a perturbed distribution $\tilde{p}$.

\subsection{IL Setting}
Since the branching samples can be shuffled and collected by skip sampling, the transition function $\mathcal{T}$ may have little effect on the training dataset.
We thus omit the transition $\mathcal{T}$ in the IL setting and view the collected branching samples drawn independently from the same distribution.
For the hypothesis policy class $\Pi$, we use the classical Rademacher complexity $\Rad_n(\Pi)$
\begin{align*}
  \Rad_n(\Pi) = \mathbb{E}_{\mathcal{D}_{\text{IL}}}\mathbb{E}_{\sigma}\left[\sup_{\pi\in\Pi}\frac{1}{nT}\sum_{i=1}^n\sigma_{i}\sum_{t=0}^{T-1}\ell(\pi(G^{(i)}_t),\pi_{E}({G^{(i)}_t})\right]
\end{align*}
and $\widetilde{\Rad}_n(\Pi)$ under the instance augmentation
\begin{align*}
  \widetilde{\Rad}_n(\Pi) = \mathbb{E}_{\mathcal{D}_{\text{IL}}}\mathbb{E}_{\sigma}\left[\sup_{\pi\in\Pi}\frac{1}{nT}\sum_{i=1}^n\sigma_{i}\sum_{t=0}^{T-1}\ell(\pi(\tilde{G}^{(i)}_t),\pi_{E}({\tilde{G}^{(i)}_t}))\right]
\end{align*}
to bound the generalization gap, where $G\sim p_{\text{tr}}(G_{1:T}|\pi_E)$ and $\tilde{G}\sim \tilde{p}(\tilde{G}_{1:T}|\pi_E)$, { $\sigma=(\sigma_1, \dots, \sigma_n)$  is Rademacher variables with $\sigma_n$ independent uniform random variables taking $\{-1, 1\}$.
We denote $\sqrt{\frac{\log(1/\delta)}{2n}}$ as $\Omega_n$ for simplicity}.

Next, we derive the IL generalization bound between the empirical risks in the training dataset and the expected risk in the testing dataset.
The classical proof of the generalization bound with Rademacher complexity can be found in \cite{MohriRostamizadehTalwalkar18}, and the in-distribution generalization bound for data augmentation is discussed in \cite{Chen20Grouptheoretic}. We now derive the out-of-distribution (OOD) version of the generalization bound under data augmentation in the IL setting (\ref{equ_il_bound2}).
{
We find that the cross-entropy loss for branching has an upper bound in our experiments, and thus the loss is Lipschitz given the fixed value range.  
\begin{theorem}\label{thm_IL_bound}
  Suppose that the function $\ell(\pi_1(\cdot), \pi_2(\cdot))$ is bounded by 1 and is $\frac{1}{2}$-Lipschitz continuous for any policy $\pi_1$ and $\pi_2$.
  With probability at least $1-\delta$ and for any $\pi\in\Pi$, we have the generalization bounds in the out-of-distribution imitation learning setting as follows.
  The generalization bound for empirical risk minimization
    \begin{equation}\label{equ_il_bound1}
    \begin{aligned}
      \mathcal{J}_{p_{\text{te}}}(\hat{\pi})-\hat{\mathcal{J}}_{p_{\text{tr}}}(\hat{\pi})\le&\mathcal{W}_{\mathcal{G}}(p_{\text{tr}}(G_{1:T}|\pi_E), p_{\text{te}}(G_{1:T}|\pi_E))\\
      &+ 2\Rad_{n}(\Pi)+\Omega_n.
    \end{aligned}
    \end{equation}
 The generalization bound for data augmentation
    \begin{equation}\label{equ_il_bound2}
    \begin{aligned}
      \mathcal{J}_{p_{\text{te}}}(\hat{\pi})-\hat{\mathcal{J}}_{\tilde{p}}(\hat{\pi})&\le \mathcal{W}_{\mathcal{G}}(\tilde{p}(G_{1:T}|\pi_E), p_{\text{te}}(G_{1:T}|\pi_E))\\
       &+2\widetilde{\Rad}_n(\Pi)+\Omega_n.
    \end{aligned}
    \end{equation}
  Furthermore, we suppose that the instance augmentation technique reduces the distribution discrepancy between training and testing data, i.e., for some $\alpha<1$, 
  \begin{align*}
      \mathcal{W}_{\mathcal{G}}&(\tilde{p}(G_{1:T}|\pi_E), p_{\text{te}}(G_{1:T}|\pi_E)) \\
      &\le \alpha \mathcal{W}_{\mathcal{G}}(p_{\text{tr}}(G_{1:T}|\pi_E), p_{\text{te}}(G_{1:T}|\pi_E)).
  \end{align*}
   For $\alpha\le1/3$, the generalization bound in Equation \ref{equ_il_bound2} is lower than that in Equation \ref{equ_il_bound1}, that is
  \begin{equation}\label{equ_compare_bound1}
  \begin{aligned}
    &\mathcal{W}_{\mathcal{G}}(\tilde{p}(G_{1:T}|\pi_E), p_{\text{te}}(G_{1:T}|\pi_E))
       +2\widetilde{\Rad}_n(\Pi)\\
       &\le \mathcal{W}_{\mathcal{G}}(p_{\text{tr}}(G_{1:T}|\pi_E), p_{\text{te}}(G_{1:T}|\pi_E))+2\Rad_{n}(\Pi).
  \end{aligned}
  \end{equation}
  
\end{theorem}
}
Thus, Theorem \ref{thm_IL_bound} shows that data augmentation is beneficial for out-of-distribution generalization settings.

\subsection{RL Setting}
We now explore the theory of instance augmentation for OOD RL settings.
The transition function $\mathcal{T}$ in the RL setting is essential and should be considered in the generalization analysis.
The policy to update repeatedly interacts with the environment and collects branching samples  during training, and the transition function plays a vital role in the data collection. 
We denote the Rademacher complexity for policy class $\Pi$ by
\begin{align*}
  \Rad_n(\Pi) = \mathbb{E}_{\mathcal{D}_{\text{RL}}}\mathbb{E}_{\sigma}\left[\sup_{\pi\in\Pi}\frac{1}{nT}\sum_{i=1}^n\sigma_i\sum_{t=0}^{T-1}-\gamma^t R({G}_t^{(i)})\right].
\end{align*}
For instance augmentation,
\begin{align*}
  \widetilde{\Rad}_n(\Pi) = \mathbb{E}_{\mathcal{D}_{\text{RL}}}\mathbb{E}_{\sigma}\left[\sup_{\pi\in\Pi}\frac{1}{nT}\sum_{i=1}^n\sigma_i\sum_{t=0}^{T-1}-\gamma^t R(\tilde{G}_t^{(i)})\right].
\end{align*}
We derive the following generalization bound and generalization error for the RL setting in Theorem 
\ref{thm_rl_error}.

Notice that the sample distribution is closely related to $\pi$ and varies with different $\pi$, posing difficulties in analyzing the generalization bound.
Thus, we consider the reparameterizable RL proposed in \cite{pmlr-v97-wang19o}.
Suppose that the policy $\pi$ is deterministic and given an instance $G$, the solving process is deterministic since the transition $\mathcal{T}$ is also deterministic.

We make the following assumptions.
(\textbf{A1}) The transition functions in the training and testing environment remain the same, as the solver follows rigorous mathematical rules and node selection policies.
{
(\textbf{A2})
The transition function $\mathcal{T}$ is Lipschitz continuous with Lipschitz constant of $L_1$:
\begin{align}
  c(\mathcal{T}(G,\pi(G)),\mathcal{T}(G',\pi(G')))& \le L_1c(G,G'), \forall G, G' \text{ and } \pi. 
\end{align}
The state reward function $R:\mathcal{G}\to\mathbb{R}$ is $L_3$-Lipschitz, i.e., 
\begin{align}
  |R(G)-R(G')| & \le L_3c(G,G'), \forall G, G'.
\end{align}
}
{
\begin{theorem}\label{thm_rl_error}
  Consider the solvers that can be formulated as the reparameterizable RL with horizon $T$.
  Suppose that the expected accumulated negative function $R$ is normalized and bounded by 1.
  The state reward function $R:\mathcal{G}\to\mathbb{R}$ is $L_3$-Lipschitz.
  With probability at least $1-2\delta$, we have
  \begin{equation}\label{equ_rl_bound_thm}
  \begin{aligned}
    \mathcal{J}_{p_{\text{te}}}(\tilde{\pi})&-\mathcal{J}_{p_{\text{te}}}(\pi^*)\le 4\widetilde{\Rad}_n(\Pi)+2\Omega_n \\
    &+2\sum_{t=0}^{T-1}\gamma^tL_3L_1^t\mathcal{W}_{\mathcal{G}}(\tilde{p}, p_{\text{te}}).
  \end{aligned}
  \end{equation}
\end{theorem}}
Similar to Theorem \ref{thm_IL_bound}, the bound (\ref{equ_rl_bound_thm}) is lower than that without augmentation under appropriate conditions.

\begin{table*}[t]
\caption{Statistical information for the four synthesis benchmarks. Notice that the largest sizes of instances (transfer distribution $D_6$) are around five to ten times larger than instances on the training distributions ($D_1$) on all the benchmarks, and are much larger than the previous works. }
\label{table:instance_stat}
\centering
\small{
\vspace{-1mm}
\label{table:dataset}
\centering
\resizebox{0.9\textwidth}{!}{
\begin{tabular}{@{}ccccccccccccc@{}}
\toprule
\toprule
&\multicolumn{6}{c}{Set Covering} & \multicolumn{6}{c}{Combinatorial Auctions}\\
\cmidrule(r){1-7} \cmidrule(lr){8-13}
 & \multicolumn{1}{c}{$D_1$ } & \multicolumn{1}{c}{$D_2$  } & \multicolumn{1}{c}{$D_3$  } & \multicolumn{1}{c}{$D_4$  }& \multicolumn{1}{c}{$D_5$ }& \multicolumn{1}{c}{$D_6$  }& \multicolumn{1}{c}{$D_1$ } & \multicolumn{1}{c}{$D_2$  } & \multicolumn{1}{c}{$D_3$  } & \multicolumn{1}{c}{$D_4$  }& \multicolumn{1}{c}{$D_5$ }& \multicolumn{1}{c}{$D_6$  }\\
 \cmidrule(r){1-7} \cmidrule(lr){8-13}
Constraint number &500 &1000 &2000 &4000 & 3000 &8000 &193.03 &386.25 &576.75 &771.65 & 963.56 &1156.59\\
Variable number&1000 &1000 &1000 &1000 &1000 &1000& 500 &1000 &1500 &2000 &2500 & 3000\\
\bottomrule
\\[-7pt]
\end{tabular}
}
\newline
\newline
\resizebox{0.9\textwidth}{!}{
\begin{tabular}{@{}ccccccccccccc@{}}
\toprule
\toprule
&\multicolumn{6}{c}{Capacitated Facility Location} & \multicolumn{6}{c}{Maximum Independent Set}\\
\cmidrule(r){1-7} \cmidrule(lr){8-13}
 & \multicolumn{1}{c}{$D_1$ } & \multicolumn{1}{c}{$D_2$  } & \multicolumn{1}{c}{$D_3$  } & \multicolumn{1}{c}{$D_4$  }& \multicolumn{1}{c}{$D_5$ }& \multicolumn{1}{c}{$D_6$  }& \multicolumn{1}{c}{$D_1$ } & \multicolumn{1}{c}{$D_2$  } & \multicolumn{1}{c}{$D_3$  } & \multicolumn{1}{c}{$D_4$  }& \multicolumn{1}{c}{$D_5$ }& \multicolumn{1}{c}{$D_6$  }\\
 \cmidrule(r){1-7} \cmidrule(lr){8-13}
Constraint number &10201&20301&40501&60701&80901&161701&1952.34&3946.56 &5942.34 &7936.40 &9934.12 & 11931.65\\
Variable number&10100 &20100 &40100&60100 &80100 & 160100& 500 &1000 &1500 &2000&2500 & 3000\\
\bottomrule
\\[-7pt]
\end{tabular}
}
}
\vspace{-2mm}
\end{table*}

\section{Experiments}
In this part, we conduct comparative experiments on multiple well-recognized benchmarks to demonstrate the effectiveness of our proposed AdaSolver framework in terms of solving efficiency and sample efficiency. 
Additionally, we conduct ablation studies and comprehensive investigations to offer deeper insights into the AdaSolver approach.

\subsection{Experiment Setup}
\subsubsection{Benchmarks}
We evaluate the performance of our proposed AdaSolver on four widely used benchmarks of NP-hard synthesis problem families and a challenging real-world Anonymous dataset.
The synthesis datasets include the set covering, combinatorial auctions, capacitated facility location and maximum independent set problems.
The data generation settings follow the conventions established in \cite{gasse2019exact}, \cite{gupta2020hybrid}, and \cite{gupta2022lookback}.
In each benchmark, we generate 10,000 training instances and 2,000 validation instances for AdaSolver.
For the final evaluation, we generate an in-distribution testing set of 32 instances $D_1$.
Additionally, we generate five transfer sets $D_2-D_6$, each containing 32 instances with larger sizes.
We emphasize that the instance sizes in the transfer datasets increase in order, with the maximum size much larger than those used in previous works.
The Anonymous dataset, a curated subset of the MILPLIB library, is a well-known real-world dataset.
It consists of the large-size instances with 98 for training, 20 for validation and 20 for testing.
On average, the instances in the Anonymous dataset contain 49,603 constraints and 37,881 variables.

We report the instances' size for each synthesis benchmark in Table \ref{table:instance_stat}.
Further details regarding the benchmarks and instance generation for the synthesis datasets can be found in Appendix \ref{appendix:benchmarks}, due to space limitations.

\subsubsection{Metric}
We present the results of the comparative experiments for different branching approaches, measured in terms of average solving time (Time, lower is better) and primal-dual integral (PD integral, lower is better). 
Further details on the PD integral metric can be found in Appendix \ref{appendix:PD_metric}.
Moreover, we provide additional results in terms of another two metrics, i.e., the average branching nodes (Node) and the average primal-dual gap (PD gap, lower is better) in Appendix \ref{appendix:more_results_main_evalutation}.

Following \cite{wanglearning}, we mainly focus on the \emph{Time metric} in distributions where the solver successfully solves all instances within the given time limit, since the solving time serves as a direct measure of solving efficiency.
However, when solvers fail to solve a majority of instances within the time limit, the average solving time becomes almost the same across different methods, making it challenging to determine the best-performing method.
So we focus on another well-recognized metric \emph{primal-dual integral} in these distributions instead.
Given a time limit, the lower primal-dual integral indicates higher-quality feasible solutions found by the solver.
Specifically, we focus on the Time metric for $D_1-D_2$ in the synthesis benchmarks, $D_3$ in the combinatorial auctions and capacitated facility location benchmarks.
For the remaining distributions, we focus on the primal-dual integral.

The Node metric reflects the size of the B\&B searching trees.
When the solver successfully solves an instance, a smaller number of explored nodes may indicate a more efficient search policy.

\subsubsection{Baselines}
Our baselines include a SCIP branching rule and six learning-based branching models containing the competitive IL-based and RL-based approaches.
The SCIP branching rule is the traditional state-of-the-art reliability pseudo-cost (RPB) branching heuristic. 
We then introduce the learning-based branching baselines.
(1) SVMRank \cite{khalil2016learning} employs a support vector machine for learning to select branching variables. 
(2) Trees \cite{hansknecht2018cuts} is a branching policy that utilizes ExtraTrees. 
(3) LMart \cite{Joachims2002OptimizingSE} is based on LambdaMART. 
(4) tMDP+DFS is a reinforcement learning method based on tree MDP \cite{NEURIPS2022_756d74cd}. 
(5) TreeGate is an IL-based branching approach introducing the B\&B tree structure information \cite{Zarpellon_Jo_Lodi_Bengio_2021}.
(6) GNN is the graph neural network approach for learning to branch \cite{gasse2019exact}.
Finally, we build upon the IL-based GNN branching policy (GNN) \cite{gasse2019exact} and the RL-based GNN branching policy (tMDP+DFS) \cite{NEURIPS2022_756d74cd} to develop our AdaSolver framework, resulting in two methods: AdaSolver-IL and AdaSolver-RL.
As mentioned in the original paper of tMDP+DFS \cite{NEURIPS2022_756d74cd}, the RL-based method performs worse than RPB or the IL-based baselines, which coincides with our results in Table \ref{table:performance}. 
However, tMDP+DFS is the newest open-source RL branching method with the best performance we have found so far, and our focus is primarily on the improvement achieved by AdaSolver-RL over tMDP+DFS.

\begin{table*}[!th]
\caption{The generalization results on multiple distributions of the benchmarks. For each benchmark, $D_1-D_6$ are six distributions with increasing instance sizes. We report the mean of solving time and PD integral metrics. The metrics we mainly focus on in each distribution are indicated with $\downarrow$, where lower values indicate better performance. AdaSolver-IL is built on the IL-based GNN baselines, while AdaSolver-RL is built on the RL-based tMDP+DFS baseline. }
\small{
\vspace{-1mm}
\label{table:performance}
\centering
\resizebox{\textwidth}{!}{
\begin{tabular}{@{}ccccccccccccc@{}}
\toprule
\toprule
 & \multicolumn{2}{c}{$D_1$ } & \multicolumn{2}{c}{$D_2$ } & \multicolumn{2}{c}{$D_3$ } & \multicolumn{2}{c}{$D_4$ }& \multicolumn{2}{c}{$D_5$}& \multicolumn{2}{c}{$D_6$}\\ \midrule
Method & Time(s) $\downarrow$ & PD integral & Time(s) $\downarrow$ & PD integral  & Time(s)  & PD integral $\downarrow$ & Time(s) & PD integral $\downarrow$ & Time(s) & PD integral $\downarrow$ & Time(s) & PD integral $\downarrow$\\\cmidrule(r){1-3} \cmidrule(lr){4-5} \cmidrule(l){6-7}\cmidrule(l){8-9}\cmidrule(l){10-11}\cmidrule(l){12-13}
RPB & 10.98  & 114.91 & 87.62 & 666.215& 831.41 & 7152.10 & 900.00 & 11494.93 & 900.00 & 14757.09 & 900.00 & 23564.46\\
\cmidrule(r){1-3} \cmidrule(lr){4-5} \cmidrule(l){6-7}\cmidrule(l){8-9}\cmidrule(l){10-11}\cmidrule(l){12-13}
SVMRank & 11.40  & 116.21 & 158.29  & 989.67& 890.66 &8546.88 & 900.00 & 13444.79 & 900.00 & 17018.42 & 900.00 & 24535.92\\
Trees & 14.86  & 131.23 & 210.26  &1238.87 & 899.87  & 8690.80& 900.00 & 13226.00 & 900.00 & 16772.03 & 900.00 & 24599.38\\
LMART &9.46   &106.92  &127.34   &812.27 & 875.87 &7797.14 & 900.00 & 12927.04 & 900.00 & 16668.55 & 900.00 & 24525.11\\
tMDP+DFS  &20.73 &159.59 &444.38 &2892.12 &900.00 &11559.79 &900.00 &15393.24 &900.00 &18384.06 &900.00 &24387.40\\
TreeGate & 25.16  & 187.99 & 271.82  &1790.70 & 900.00 &10391.53 & 900.00 & 14389.55 & 900.00 & 17747.06 & 900.00 & 24118.51\\
GNN & 9.25  &  116.22 & 95.57  &716.25 & 844.74 & 7310.17&900.00 &11899.40 &900.00&16067.00& 900.00& 23786.51 \\
\cmidrule(r){1-3} \cmidrule(lr){4-5} \cmidrule(l){6-7}\cmidrule(l){8-9}\cmidrule(l){10-11}\cmidrule(l){12-13}
AdaSolver-RL &18.87&152.26&433.97&2885.97&900.00&11330.87&900.00&14466.95&900.00&17550.63&900.00&24319.00\\
AdaSolver-IL & \textbf{6.28} & {80.67} & \textbf{73.78} &{515.92} & 839.80 & \textbf{7079.13}& 900.00 &\textbf{10691.05} &900.00&\textbf{14452.71}&900.00&\textbf{21792.69}\\ \bottomrule
\\[-7pt]
    & \multicolumn{12}{ c }{Set Covering} \\
    \\[-3pt]
\end{tabular}
}
\newline
\vspace{-1mm}
\newline
\resizebox{\textwidth}{!}{
\begin{tabular}{@{}ccccccccccccc@{}}
\toprule
\toprule
& \multicolumn{2}{c}{$D_1$ } & \multicolumn{2}{c}{$D_2$ } & \multicolumn{2}{c}{$D_3$ } & \multicolumn{2}{c}{$D_4$ }& \multicolumn{2}{c}{$D_5$}& \multicolumn{2}{c}{$D_6$}\\ \midrule
Method & Time(s) $\downarrow$ & PD integral  & Time(s) $\downarrow$ & PD integral  & Time(s) $\downarrow$  & PD integral  & Time(s) & PD integral $\downarrow$ & Time(s) & PD integral $\downarrow$ & Time(s) & PD integral $\downarrow$\\\cmidrule(r){1-3} \cmidrule(lr){4-5} \cmidrule(l){6-7}\cmidrule(l){8-9}\cmidrule(l){10-11}\cmidrule(l){12-13}
RPB & 2.84  & 32.20 & 23.20  & 156.62 & 224.66  & {539.25} & 838.77 & 1622.46 & 900.00 & 2313.35 & 900.00 &  2657.51\\\cmidrule(r){1-3} \cmidrule(lr){4-5} \cmidrule(l){6-7}\cmidrule(l){8-9}\cmidrule(l){10-11}\cmidrule(l){12-13}
SVMRank & 2.59  & 26.32 &  51.12& 194.46 & 427.01 & 625.01   & 881.33& 1730.40 & 900.00 & 2293.06 & 900.00 & 2580.31 \\
Trees & 2.76 & 26.44& 87.05 &221.68 & 790.01 & 994.48 & 900.00   & 1782.20 & 900.00 & 2323.16  & 900.00  & 2535.18 \\
LMART & 1.90 & 24.69& 28.08  & 156.10& 287.69 & 446.49   & 862.31 & 1563.85 & 900.00  & 2136.37 & 900.00  & 2455.08  \\
tMDP+DFS &2.18 &24.76  &27.98 &146.15 &312.93 &518.45 &860.76 &1871.73 &900.00 &2664.34 &900.00 &3103.08\\
TreeGate & 5.75  & 37.74 & 84.44  &303.53 & 623.25 & 1168.37 &  891.50 & 2486.65 & 900.00 & 3302.70 & 900.00 & 3771.89\\

GNN & 2.20  & 30.48 & 20.30  &167.60 & 195.68 & 384.17&819.28 &1300.99 &900.00&1931.54&900.00&2293.41\\
\cmidrule(r){1-3} \cmidrule(lr){4-5} \cmidrule(l){6-7}\cmidrule(l){8-9}\cmidrule(l){10-11}\cmidrule(l){12-13}
AdaSolver-RL &2.04 &106.83 & 17.28 &117.28 &323.94 &517.91 &846.32 &1685.69 &900.00 &2379.82 &900.00 &2980.48\\
AdaSolver-IL & \textbf{1.63}  & {22.44} & \textbf{16.34}  & {117.13} & \textbf{178.32}  & 314.32& 817.57 & \textbf{1154.79} & 900.00 & \textbf{1716.55} & 900.00 & \textbf{2107.19}\\ \bottomrule
\\[-7pt]
    & \multicolumn{12}{ c }{Combinatorial Auctions}  \\
    \\[-3pt]
\end{tabular}
}
\newline
\vspace{-1mm}
\newline
\resizebox{\textwidth}{!}{\begin{tabular}{@{}ccccccccccccc@{}}
\toprule
\toprule
& \multicolumn{2}{c}{$D_1$ } & \multicolumn{2}{c}{$D_2$ } & \multicolumn{2}{c}{$D_3$ } & \multicolumn{2}{c}{$D_4$ }& \multicolumn{2}{c}{$D_5$}& \multicolumn{2}{c}{$D_6$}\\ \midrule
Method & Time(s) $\downarrow$ & PD integral  & Time(s) $\downarrow$ & PD integral  & Time(s) $\downarrow$  & PD integral  & Time(s) & PD integral $\downarrow$ & Time(s) & PD integral $\downarrow$ & Time(s) & PD integral $\downarrow$\\\cmidrule(r){1-3} \cmidrule(lr){4-5} \cmidrule(l){6-7}\cmidrule(l){8-9}\cmidrule(l){10-11}\cmidrule(l){12-13}
RPB & 65.67  & {170.34} & 287.52 & {719.26} & 689.45  & 3333.72 & 886.54 & 10762.83 & 892.90 & 20102.78 & 897.51 & 27527.24\\
\cmidrule(r){1-3} \cmidrule(lr){4-5} \cmidrule(l){6-7}\cmidrule(l){8-9}\cmidrule(l){10-11}\cmidrule(l){12-13}
SVMRank & 56.98  & 274.04 & 207.16  &1011.28 &592.15  & 3913.06 & 884.87 & 13898.43 & 873.08 & 22201.25 & 887.00 & 33649.01\\
Trees & 77.41  & 285.65 & 262.54  &1038.46 &652.15  & 3895.60 & 887.00 & 13649.01 & 878.81 & 22433.54 & 889.54 & 34909.74\\
LMART & 51.30  & 281.76 & 201.06  &988.92 & 591.41 & 3927.82 & 884.29 & 13908.60 & 872.67 & 22455.29 & 886.84 & 34976.25\\
tMDP+DFS & 73.35 & 275.38 & 270.71  & 1108.57& 683.42 & 4253.40 & 886.07 & 12950.94 & 888.14 & 23675.80 & 893.70 & 34995.77\\
TreeGate & 47.96  & 180.17 & 192.29  & 757.39& 661.64 &4238.80  & 873.98 & 10553.59 & 886.34 & 19564.65 & 897.63 & 27529.18\\

GNN & 51.63  &290.91  & 235.07  & 1107.29& 660.25 & 4418.54 &854.27& 10297.25 &868.93& 19033.90 & 890.53& 32302.74 \\
\cmidrule(r){1-3} \cmidrule(lr){4-5} \cmidrule(l){6-7}\cmidrule(l){8-9}\cmidrule(l){10-11}\cmidrule(l){12-13}
AdaSolver-RL  & 64.10  & 257.10 & 233.64 & 892.77 & 587.02 & 3332.38 & 852.53 &7779.28& 849.93 & 14727.70 &852.51& \textbf{27298.25}\\
AdaSolver-IL & \textbf{34.51}   & 196.56 & \textbf{147.47}  & 738.36 & \textbf{555.66}  & {3324.63} & 731.81 & \textbf{6644.35} & 819.31 & \textbf{13818.34} &884.48&28231.23\\ \bottomrule
\\[-7pt]
    & \multicolumn{12}{ c }{Capacitated Facility Location} \\
    \\[-3pt]
\end{tabular}
}
\newline
\vspace{-1mm}
\newline
\resizebox{\textwidth}{!}{\begin{tabular}{@{}ccccccccccccc@{}}
\toprule
\toprule
& \multicolumn{2}{c}{$D_1$ } & \multicolumn{2}{c}{$D_2$ } & \multicolumn{2}{c}{$D_3$ } & \multicolumn{2}{c}{$D_4$ }& \multicolumn{2}{c}{$D_5$}& \multicolumn{2}{c}{$D_6$}\\ \midrule
Method & Time(s) $\downarrow$ & PD integral & Time(s) $\downarrow$ & PD integral  & Time(s)  & PD integral $\downarrow$ & Time(s) & PD integral $\downarrow$ & Time(s) & PD integral $\downarrow$ & Time(s) & PD integral $\downarrow$\\\cmidrule(r){1-3} \cmidrule(lr){4-5} \cmidrule(l){6-7}\cmidrule(l){8-9}\cmidrule(l){10-11}\cmidrule(l){12-13}
RPB & 16.46  & 86.57 &302.43  & 1045.88 & 883.79 & 3818.32 & 900.00 & 4634.32 & 900.00 & 5067.01 & 900.00 & 5683.46\\
\cmidrule(r){1-3} \cmidrule(lr){4-5} \cmidrule(l){6-7}\cmidrule(l){8-9}\cmidrule(l){10-11}\cmidrule(l){12-13}
SVMRank & 18.37  & 76.17  &456.19 & 879.49 &883.18  &2539.03 & 900.00 & 3640.36 & 900.00 & 4313.94 &  900.00 & 5105.40\\
Trees & 26.04  & 91.01 & 701.12  &1353.35 & 900.00 & 2828.39 & 900.00 & 3729.98 & 900.00 & 4353.62 &900.00 & 5013.31\\
LMART &12.80   & 61.79 & 445.80  & 826.64& 877.65  & 2475.11 & 900.00 &3570.20 & 900.00 & 4221.15 & 900.00 & 5042.00\\
tMDP+DFS & 9.45  & 54.00 &  294.04 & 644.69& 781.44 & 2201.37& 897.91 & 3628.92 & 900.00 & 4613.32 & 900.00 & 5314.64\\
TreeGate &67.48   &229.32  & 844.92  &2711.83 & 900.00 &3939.96 & 900.00 & 4520.56 & 900.00 & 4906.46 & 900.00 & 5299.72\\

GNN & 11.09  & 65.73 & 254.12  &558.31 & 817.04 & 2284.32 &900.00& 3532.03& 900.00& 4387.27&  900.00& 5168.77\\
\cmidrule(r){1-3} \cmidrule(lr){4-5} \cmidrule(l){6-7}\cmidrule(l){8-9}\cmidrule(l){10-11}\cmidrule(l){12-13}
AdaSolver-RL & 8.43 & 51.46 & 282.54 & 627.81 & 803.75 & \textbf{2117.01} & 879.81 & \textbf{3318.08} & 900.00 &\textbf{ 4148.24} & 900.00 & \textbf{5099.92} \\
AdaSolver-IL & \textbf{7.39}  & {44.56} & \textbf{210.36} & {460.80} & 806.78 & 2355.11&900.00&3591.38 &900.00&4375.48&900.00& 5152.28\\ \bottomrule
\\[-7pt]
    & \multicolumn{12}{ c }{Maximum Independent Set} \\
    \\[-3pt]
\end{tabular}
}
}

\end{table*}

\subsubsection{Implementation}
Throughout the experiments, we utilize the state-of-the-art open-source solver SCIP 7.0.3 as the backend solver. 
The solving time limit on a testing instance is 900s.
Similar to \cite{gasse2019exact, gupta2020hybrid, gupta2022lookback}, we deactivate solver restarts and allow employing the cutting plane generation module only at the root node.
We keep all the other SCIP parameters to their default so as to better observe the improvements achieved by AdaSolver and make comparisons of these approaches as fair.

{ We first train the GNN branching policy with villina IL- and RL-based methods for the first ten epochs and then begin the adversarial training process. 
We train the augmentation network from scratch. }
We use the Adam optimizer \cite{kingma2014adam} to optimize the model parameters.
During the training of AdaSolver, the adversarial augmentation policy randomly selects training instances for augmentation.
We then collect branching samples from the augmented instances.
We choose the augmentation operators that perform the best for each benchmark.
More details for these baselines and our AdaSolver are available in Appendix \ref{appendix:implementation}.

\vspace{-1em}
\subsection{Comparative Experiments on Solving Efficiency}
\subsubsection{Size Generalization}
\label{subsection:size_generalization}

We evaluate the generalization ability of AdaSolver and compare it to other baselines on four benchmarks of generated datasets (see Table \ref{table:performance}).
According to Table \ref{table:performance}, our AdaSolver-IL consistently outperforms the baselines in the four synthesis benchmarks in terms of Time or PD Integral metrics.
AdaSolver-RL achieves a significant improvement over tMDP+DFS in terms of Time or PD Integral metrics.
Please refer to Appendix \ref{appendix:more_results_main_evalutation} for results in terms of Nodes and PD gap metrics.

In the easier distributions ($D_1$, $D_2$ in the four benchmarks and $D_3$ in the combinatorial auctions and maximum independent set benchmarks), AdaSolver-IL achieves the shortest solving time.
The GNN is a competitive baseline that outperforms other learning-based baselines on most of the easier distributions of the four benchmarks.
While LMART is another promising learning-based baseline with stable performance. 
AdaSolver leads to a remarkable \textbf{35.25\%} improvement in solving time over the B\&B solver across various distributions.
Compared to GNN, AdaSolver-IL achieves \textbf{24.60\%} improvement of solving time on the easier distributions. 
AdaSolver-RL improves the solving time of tMDP+DFS by \textbf{10.75\%}.
The results demonstrate that the AdaSolver is effective in improving the performance in both in-distribution and out-of-distribution situations.

In the harder distributions ($D_4-D_6$ in the four benchmarks and $D_3$ in the set covering and capacity facility location benchmarks), AdaSolver still significantly outperforms other baselines with the lowest PD integral, leading to a PD integral improvement of \textbf{9.53\%} over GNN and \textbf{11.68\%} over tMDP+DFS.
This demonstrates that AdaSolver can consistently enhance the generalization ability of the GNN and tMDP+DFS baselines to instances with over five to ten times larger sizes.



\begin{table}[ht]
\caption{{The results in the challenging real-world Anonymous dataset show that AdaSolver-IL achieves the lowest PD gap and PD integral.}}
\vspace{-1mm}
\label{table:real_world_dataset}
\centering
\begin{tabular}{@{}ccccc@{}}
\toprule
\toprule
Method & Time(s) & Nodes  & PD integral $\downarrow$ & PD gap $\downarrow$\\\midrule
RPB & {6831.95} & {355372.63} & {46284.89} & {1.92}\\\midrule
tMDP+DFS & {5691.97} &  {327665.67} & {247486.52} & {1.02} \\
GNN & {5612.58} & {316961.59} & {229794.84} & {1.10} \\\midrule
AdaSolver-RL& {5523.64} & {359847.44} & {229711.85} & {0.99}\\
AdaSolver-IL & {\textbf{5508.74}} & {\textbf{294718.36}}
& {\textbf{207294.27}}& {\textbf{0.92}}\\ \bottomrule
\end{tabular}

\end{table}

\begin{table*}[t]
\caption{Sample efficiency of our proposed AdaSolver-IL. \textit{Method ($n$ instances)} means that the \textit{Method} is trained on $n$ instances. The GNN baseline faces significant performance degradation when the training instances are few (here we use 1\% of the training instances). AdaSolver achieves a 36.20\% improvement in solving time over the competitive GNN branching baseline with only 1\% of the training instances, demonstrating a significant improvement in sample efficiency.}
\small{
\vspace{-1mm}
\label{table:sample_efficiency_under_less_data}
\centering
\resizebox{\textwidth}{!}{
\begin{tabular}{@{}ccccccccccccc@{}}
\toprule
\toprule
& \multicolumn{2}{c}{$D_1$ } & \multicolumn{2}{c}{$D_2$ } & \multicolumn{2}{c}{$D_3$ } & \multicolumn{2}{c}{$D_4$ }& \multicolumn{2}{c}{$D_5$}& \multicolumn{2}{c}{$D_6$}\\ \midrule
Method & Time(s) $\downarrow$ & PD integral & Time(s) $\downarrow$ & PD integral  & Time(s)  & PD integral $\downarrow$ & Time(s) & PD integral $\downarrow$ & Time(s) & PD integral $\downarrow$ & Time(s) & PD integral $\downarrow$\\\cmidrule(r){1-3} \cmidrule(lr){4-5} \cmidrule(l){6-7}\cmidrule(l){8-9}\cmidrule(l){10-11}\cmidrule(l){12-13}
GNN (10000 instances)& 9.25  &  116.22 & 95.57  &716.25 & 844.74 & 7310.17&900.00 &11899.40 &900.00&16067.00& 900.00& 23786.51 \\
GNN (100 instances)& 13.41 & 138.13 & 278.75 & 1694.85 & 900.00 & 10049.36 & 900.00 & 14158.38 & 900.00 & 16706.91 & 900.00 & 24062.04 \\
AdaSolver-IL (100 instances) &7.35 &85.06&237.97 &1289.08&900.00&8840.34&900.00&12381.27&900.00&15240.56&900.00&22754.06\\
 \bottomrule
\\[-7pt]
    & \multicolumn{12}{ c }{Set Covering} \\
    \\[-3pt]
\end{tabular}
}
\newline
\vspace{1mm}
\newline
\resizebox{\textwidth}{!}{
\begin{tabular}{@{}ccccccccccccc@{}}
\toprule
\toprule
& \multicolumn{2}{c}{$D_1$ } & \multicolumn{2}{c}{$D_2$ } & \multicolumn{2}{c}{$D_3$ } & \multicolumn{2}{c}{$D_4$ }& \multicolumn{2}{c}{$D_5$}& \multicolumn{2}{c}{$D_6$}\\ \midrule
Method & Time(s) $\downarrow$ & PD integral & Time(s) $\downarrow$ & PD integral  & Time(s) $\downarrow$ & PD integral & Time(s) & PD integral $\downarrow$ & Time(s) & PD integral $\downarrow$ & Time(s) & PD integral $\downarrow$\\\cmidrule(r){1-3} \cmidrule(lr){4-5} \cmidrule(l){6-7}\cmidrule(l){8-9}\cmidrule(l){10-11}\cmidrule(l){12-13}
GNN (10000 instances) & 2.20  & 30.48 & 20.30  &167.60 & 195.68 & 384.17&819.28 &1300.99 &900.00&1931.54&900.00&2293.41\\
GNN (100 instances)&3.28 & 35.51 & 40.05 & 185.77 & 577.08 & 820.29 & 900.00 & 1981.16 & 900.00 & 2582.09 & 900.00 & 3014.64 \\
AdaSolver-IL (100 instances)&2.62 & 35.69 & 24.54 & 182.54 &217.18 & 428.28 & 830.51 & 1349.77 & 900.00 & 1934.11 & 900.00 & 2326.64 \\
\bottomrule
\\[-7pt]
    & \multicolumn{12}{ c }{Combinatorial Auctions}  \\
    \\[-3pt]
\end{tabular}
}
}
\caption{Ablation study of adversarial augmentation policy, where RM-IL and RM-RL denote the instance augmentation approach with random masking policy for the IL-based GNN and RL-based tMDP+DFS baselines, respectively. {For fair comparison, we have carefully tuned the random masking ratio to make the method competitive}. While RM is sometimes effective in improving the in-distribution performance ($D_1$) of the branching policy, it fails to generalize to larger instances.
These findings emphasize the crucial role of the adversarial augmentation policy.}
\small{
\vspace{-1mm}
\label{table:ablation}
\centering
\resizebox{\textwidth}{!}{
\begin{tabular}{@{}ccccccccccccc@{}}
\toprule
\toprule
& \multicolumn{2}{c}{$D_1$ } & \multicolumn{2}{c}{$D_2$ } & \multicolumn{2}{c}{$D_3$ } & \multicolumn{2}{c}{$D_4$ }& \multicolumn{2}{c}{$D_5$}& \multicolumn{2}{c}{$D_6$}\\ \midrule
Method & Time(s) $\downarrow$ & PD integral & Time(s) $\downarrow$ & PD integral  & Time(s)  & PD integral $\downarrow$ & Time(s) & PD integral $\downarrow$ & Time(s) & PD integral $\downarrow$ & Time(s) & PD integral $\downarrow$\\\cmidrule(r){1-3} \cmidrule(lr){4-5} \cmidrule(l){6-7}\cmidrule(l){8-9}\cmidrule(l){10-11}\cmidrule(l){12-13}
RM-IL&\textbf{6.13} &79.31 &{82.35} &{571.85} &{855.92} & {7284.54} & {900.00} &{15550.73} & 900.00 &18496.45 & 900.00 & 24254.16\\
AdaSolver-IL & 6.28 & 80.67 & \textbf{73.78} &515.92 & 839.80 &\textbf{7079.13}& 900.00 &\textbf{10691.05} &900.00&\textbf{14452.71}&900.00&\textbf{21792.69}\\\midrule
RM &18.75 & 149.36 & 436.97 & 2931.59 & {900.00} & {11641.30} & 900.00 &15478.89&{900.00} & {18638.37} &900.00&25966.91\\
AdaSolver-RL &18.87&152.26&433.97&2885.97&900.00&11330.87&900.00&14466.95&900.00&17550.63&900.00&24319.00\\
 \bottomrule
\\[-7pt]
    & \multicolumn{12}{ c }{Set Covering} \\
    \\[-3pt]
\end{tabular}
}
\newline
\vspace{1mm}
\newline
\resizebox{\textwidth}{!}{
\begin{tabular}{@{}ccccccccccccc@{}}
\toprule
\toprule
& \multicolumn{2}{c}{$D_1$ } & \multicolumn{2}{c}{$D_2$ } & \multicolumn{2}{c}{$D_3$ } & \multicolumn{2}{c}{$D_4$ }& \multicolumn{2}{c}{$D_5$}& \multicolumn{2}{c}{$D_6$}\\ \midrule
Method & Time(s) $\downarrow$ & PD integral & Time(s) $\downarrow$ & PD integral  & Time(s) $\downarrow$ & PD integral & Time(s) & PD integral $\downarrow$ & Time(s) & PD integral $\downarrow$ & Time(s) & PD integral $\downarrow$\\\cmidrule(r){1-3} \cmidrule(lr){4-5} \cmidrule(l){6-7}\cmidrule(l){8-9}\cmidrule(l){10-11}\cmidrule(l){12-13}
RM-IL& 1.70 &22.93 &{22.78} &{126.60} &464.55 &662.24 &829.49 & 1327.54 & {900.00} & {1829.98} &900.00 & 2299.41 \\
AdaSolver-IL & {\textbf{1.63}}  & 22.44 & {\textbf{16.34}}  & 117.13 & {\textbf{178.32}}  & 314.32& 817.57 & {\textbf{1154.79}} & 900.00 & {\textbf{1716.55}} & 900.00 & {\textbf{2107.19}}\\ \midrule
RM&{2.15} & {24.32} & 18.80 & 118.94 & 301.43 & 489.06 & {852.49}  & {1749.38} & 900.00 &2548.19&900.00&3247.74\\
AdaSolver-RL &2.04 &106.83 & 17.28 &117.28 &323.94 &517.91 &846.32 &1685.69 &900.00 &2379.82 &900.00 &2980.48\\
\bottomrule
\\[-7pt]
    & \multicolumn{12}{ c }{Combinatorial Auctions}  \\
    \\[-3pt]
\end{tabular}
}
\newline
\vspace{1mm}
\newline
\resizebox{\textwidth}{!}{\begin{tabular}{@{}ccccccccccccc@{}}
\toprule
\toprule
& \multicolumn{2}{c}{$D_1$ } & \multicolumn{2}{c}{$D_2$ } & \multicolumn{2}{c}{$D_3$ } & \multicolumn{2}{c}{$D_4$ }& \multicolumn{2}{c}{$D_5$}& \multicolumn{2}{c}{$D_6$}\\ \midrule
Method & Time(s) $\downarrow$ & PD integral & Time(s) $\downarrow$ & PD integral  & Time(s) $\downarrow$ & PD integral & Time(s) & PD integral $\downarrow$ & Time(s) & PD integral $\downarrow$ & Time(s) & PD integral $\downarrow$\\\cmidrule(r){1-3} \cmidrule(lr){4-5} \cmidrule(l){6-7}\cmidrule(l){8-9}\cmidrule(l){10-11}\cmidrule(l){12-13}
RM-IL&\textbf{33.24} & 215.65 &\textbf{140.40} & 752.18 & {562.18} &{3521.52} & {847.83} & {9472.67} &871.82 &18585.50 &877.97 & 32716.55\\
AdaSolver & {34.51}   & 196.56 & {147.47}  & 738.36 & {\textbf{555.66}}  & 3324.63 & 731.81 & {\textbf{6644.35}} & 819.31 & {\textbf{13818.34}} &884.48& 28231.23\\ \midrule
RM& {67.41} & {273.37} &  243.00 & 936.84 & 637.58 & 3953.49&877.48&8579.21&892.98&21496.21&893.97&39560.20\\
AdaSolver-RL & 64.10  & 257.10 & 233.64 & 892.77 & 587.02 & 3332.38 & 852.53 &7779.28& 849.93 & 14727.70 &852.51& {\textbf{27298.25}}\\\bottomrule
\\[-7pt]
    & \multicolumn{12}{ c }{Capacitated Facility Location} \\
    \\[-3pt]
\end{tabular}
}
\newline
\vspace{1mm}
\newline
\resizebox{\textwidth}{!}{\begin{tabular}{@{}ccccccccccccc@{}}
\toprule
\toprule
& \multicolumn{2}{c}{$D_1$ } & \multicolumn{2}{c}{$D_2$ } & \multicolumn{2}{c}{$D_3$ } & \multicolumn{2}{c}{$D_4$ }& \multicolumn{2}{c}{$D_5$}& \multicolumn{2}{c}{$D_6$}\\ \midrule
Method & Time(s) $\downarrow$ & PD integral & Time(s) $\downarrow$ & PD integral  & Time(s)  & PD integral $\downarrow$ & Time(s) & PD integral $\downarrow$ & Time(s) & PD integral $\downarrow$ & Time(s) & PD integral $\downarrow$\\\cmidrule(r){1-3} \cmidrule(lr){4-5} \cmidrule(l){6-7}\cmidrule(l){8-9}\cmidrule(l){10-11}\cmidrule(l){12-13}
RM-IL& \textbf{7.26} & 45.85 & {328.74} & {735.87} &871.08 & 2999.36 & 900.00 &  5684.19 & 900.00 &  6331.45& 900.00 &  6833.94\\
AdaSolver-IL & {7.39}  & 44.56 & {\textbf{210.36}} & 460.80 & 806.78 & 2355.11&900.00&3591.38 &900.00&4375.48&900.00& 5152.28\\ \midrule
RM&8.00  & 48.94 & {273.20} & {625.28} & 790.42 & 2153.45 & 890.97 &3686.33& {900.00} & {4429.57} &{900.00}&{5529.82}\\
AdaSolver-RL & 8.43 & 51.46 & 282.54 & 627.81 & 803.75 & \textbf{2117.01} & 879.81 & \textbf{3318.08} & 900.00 & \textbf{4148.24} & 900.00 & \textbf{5099.92}\\\bottomrule
\\[-7pt]
    & \multicolumn{12}{ c }{Maximum Independent Set} \\
    \\[-3pt]
\end{tabular}
}
\vspace{-3mm}
}
\end{table*}
\subsubsection{Perturbation Robustness (Real-World Dataset)}
\label{appendix:real_world_dataset}
To evaluate the performance in real-world applications, we deploy AdaSolver to a real-world Anonymous dataset, a widely used subset of the MIRPLIB.
MIRPLIB is a library of maritime inventory routing problems in global bulk shipping.
The number of constraints in the Anonymous dataset ranges from 3,375 to 159,823, and the number of variables ranges from 1,613 to 92,261.
The wide range of problem sizes, as well as real-world perturbations, pose significant challenges for the learning-based methods.
The results in Table \ref{table:real_world_dataset} demonstrate that AdaSolver is able to improve the performance of both IL- and RL-based solvers.
{AdaSolver-IL achieves the best performance in PD integral and PD gap in a predefined time limit of three hours.
A longer solving time is necessary to see the long-term performance of the solvers. }
The performance of AdaSolver sheds light on the generalizable learning-based solvers in real-world applications.

\subsection{Improving Sample Efficiency of the Branching Policy}
\label{subsection:sample_efficiency}
We compare the performance of AdaSolver-IL and the GNN solver with only 100 training instances, one percent of the instances in Section \ref{subsection:size_generalization}.
The results presented in Table \ref{table:sample_efficiency_under_less_data} highlight a significant performance degradation for the GNN baseline when trained on fewer instances. 
Specifically, there is a \textbf{115.58\%} increase in solving time and a \textbf{25.57\%} increase in PD integral.
In contrast, AdaSolver-IL achieves a notable improvement of \textbf{36.20\%} in solving time and \textbf{16.94\%} in PD integral compared to the GNN baseline, indicating a significant improvement in sample efficiency. 
Real-world scenarios often involve limited training instances, posing challenges for learning-based solvers due to the scarcity of available data.
Consequently, the remarkable performance under such constraints reflects the potential of the practical applicability of AdaSolver.

\subsection{Ablation Studies}
We conduct ablation studies on the synthesis benchmarks to investigate (1) the contributions of the adversarial augmentation policy and (2) the effectiveness of the training methods for the augmentation policy.
\subsubsection{The Contributions of the Augmentation Policy}
We compare the performance of a variant of AdaSolver (RM), which employs a random masking approach instead of the adversarial graph augmentation policy.
RM-IL and RM-RL denote the random masking policy built on the IL-based GNN and RL-based tMDP+DFS baselines, respectively.
Please see Table \ref{table:ablation} for the results in terms of Time and PD integral metrics.
{
In our experiments, we fine-tuned the hyperparameters of RM to enhance its competitiveness. For each distribution, we tested ten different groups of masking probabilities and selected the best outcomes to report.
}
An interesting observation is that RM sometimes exhibits a slight advantage over AdaSolver in terms of solving time on the distribution $D_1$ of the set covering, capacitated facility location, and maximum independent set benchmarks. 
However, AdaSolver consistently outperforms RM in other benchmarks.
RM fails to capture and generate graph structures that enable generalization to larger instances.
These findings emphasize the crucial role of the adversarial graph augmentation policy in identifying worst-case instances that contribute to the generalization of branching policy networks.

\subsection{Extensive Studies for AdaSolver-IL}

\subsubsection{Sensitivity Analysis on the Masking Proportions}
We conduct a sensitivity analysis of AdaSolver-IL's performance on various masking proportions, as presented in Table \ref{table:augmentation_magnitude}.
AdaSolver-IL x represents the method with masking proportion x, while the default masking proportion of AdaSolver-IL is set to 0.01.
For AdaSolver-IL, we choose to mask 1\% edges on the set covering benchmark and mask 1\% variables on the combinatorial auctions benchmark. 
The results demonstrate that the performance of AdaSolver-IL is not significantly affected by different masking proportions. 
AdaSolver-IL consistently outperforms the GNN baselines when the proportion is limited to 4\%. 
We observe that lower masking proportions tend to yield better-learned policies as they promote data diversity without excessively disrupting the problem structures.

\begin{table*}[t]
\caption{ Sensitivity of the masking proportion.  }
\vspace{-1mm}
\label{table:augmentation_magnitude}
\centering
\resizebox{\textwidth}{!}{
\begin{tabular}{@{}ccccccccccccc@{}}
\toprule
\toprule
& \multicolumn{2}{c}{$D_1$ } & \multicolumn{2}{c}{$D_2$ } & \multicolumn{2}{c}{$D_3$ } & \multicolumn{2}{c}{$D_4$ }& \multicolumn{2}{c}{$D_5$}& \multicolumn{2}{c}{$D_6$}\\ \midrule
Method & Time(s) $\downarrow$ & PD integral & Time(s) $\downarrow$ & PD integral  & Time(s) $\downarrow$ & PD integral & Time(s) & PD integral $\downarrow$ & Time(s) & PD integral $\downarrow$ & Time(s) & PD integral $\downarrow$\\\cmidrule(r){1-3} \cmidrule(lr){4-5} \cmidrule(l){6-7}\cmidrule(l){8-9}\cmidrule(l){10-11}\cmidrule(l){12-13}
AdaSolver-IL 0.01&{\textbf{1.63}}  & 22.44 & {\textbf{16.34}}  & 117.13 & {\textbf{178.32}}  & 314.32& 817.57 & {\textbf{1154.79}} & 900.00 & {1716.55} & 900.00 & {2107.19}\\
AdaSolver-IL 0.02& 1.96 &27.68 &19.21 &155.75 &194.64 &375.73 &819.28 &1254.26 &900.00 &1743.97 &900.00 &2051.01\\
AdaSolver-IL 0.03&1.92 &26.14 &17.95 &132.11 &202.41 &360.52 &820.05 &1240.96 &900.00 &1720.49 &900.00 &2054.11\\
AdaSolver-IL 0.04&2.22 &30.55 &20.35 &156.33 &193.72 &361.55 &820.80 &1209.92 &900.00 &\textbf{1700.80}&900.00 &\textbf{1968.02}\\
AdaSolver-IL 0.05&2.36 &32.73 &19.78 &161.43 &196.50 &384.36 &823.93 &1292.45 &900.00 &1769.48 &900.00 &2116.08\\ \bottomrule

    \\[-7pt]
    & \multicolumn{12}{ c }{Combinatorial Auctions} \\
    \\[-3pt]
\end{tabular}
}
\newline
\vspace{1mm}
\newline
\resizebox{\textwidth}{!}{
\begin{tabular}{@{}ccccccccccccc@{}}
\toprule
\toprule
& \multicolumn{2}{c}{$D_1$ } & \multicolumn{2}{c}{$D_2$ } & \multicolumn{2}{c}{$D_3$ } & \multicolumn{2}{c}{$D_4$ }& \multicolumn{2}{c}{$D_5$}& \multicolumn{2}{c}{$D_6$}\\ \midrule
Method & Time(s) $\downarrow$ & PD integral & Time(s) $\downarrow$ & PD integral  & Time(s)  & PD integral $\downarrow$ & Time(s) & PD integral $\downarrow$ & Time(s) & PD integral $\downarrow$ & Time(s) & PD integral $\downarrow$\\\cmidrule(r){1-3} \cmidrule(lr){4-5} \cmidrule(l){6-7}\cmidrule(l){8-9}\cmidrule(l){10-11}\cmidrule(l){12-13}
AdaSolver-IL 0.01& \textbf{6.28} & 80.67 & \textbf{73.78} &515.92 & 839.80 & \textbf{7079.13}& 900.00 &\textbf{10691.05} &900.00&\textbf{14452.71}&900.00&\textbf{21792.69}\\
AdaSolver-IL 0.02& 7.09 &87.37 &90.46 &671.65 &872.50 &7392.75 &900.00 &11415.87 &900.00 &14788.13 &900.00 &23364.03\\
AdaSolver-IL 0.03&7.05 &87.98 &90.39 & 672.11 &730.26 &7396.32 &900.00 &11282.07 &900.00 &15107.76 &900.00 &23151.86\\
AdaSolver-IL 0.04&9.05 &98.56 &99.42 &737.68 &869.89 &8088.66 &900.00 &12128.56 &900.00 &15376.10 &900.00 &24009.53\\
AdaSolver-IL 0.05&9.20 &98.94 &107.61 &794.04 &876.45 &8432.83 &900.00 &12479.55 &900.00 &16414.10 &900.00 &23640.26\\ \bottomrule

    \\[-7pt]
    & \multicolumn{12}{ c }{Set Covering} \\
    \\[-3pt]
\end{tabular}
\vspace{-2mm}
}

\end{table*}
\vspace{-1mm}
\begin{table*}[t]
{
\caption{{Sensitivity of the coefficient $\beta$. }}
\vspace{-1mm}
\label{table:beta}
\centering
\resizebox{\textwidth}{!}{
\begin{tabular}{@{}ccccccccccccc@{}}
\toprule
\toprule
& \multicolumn{2}{c}{$D_1$ } & \multicolumn{2}{c}{$D_2$ } & \multicolumn{2}{c}{$D_3$ } & \multicolumn{2}{c}{$D_4$ }& \multicolumn{2}{c}{$D_5$}& \multicolumn{2}{c}{$D_6$}\\ \midrule
$\beta$ & Time(s) $\downarrow$ & PD integral & Time(s) $\downarrow$ & PD integral  & Time(s) $\downarrow$ & PD integral & Time(s) & PD integral $\downarrow$ & Time(s) & PD integral $\downarrow$ & Time(s) & PD integral $\downarrow$\\\cmidrule(r){1-3} \cmidrule(lr){4-5} \cmidrule(l){6-7}\cmidrule(l){8-9}\cmidrule(l){10-11}\cmidrule(l){12-13}
0.001 & 1.82 & 25.53 & 16.59 & 156.92& 182.68 & 329.53 & 823.37 & 1292.05 & 900.00 & 1762.05 & 900.00 & 2379.13\\ 
0.01  & {\textbf{1.63}}  & 22.44 & {\textbf{16.34}}  & 117.13 & {\textbf{178.32}}  & 314.32& 817.57 & {\textbf{1154.79}} & 900.00 & \textbf{1716.55} & 900.00 & \textbf{2107.19}\\
0.1 & 2.31& 26.47 & 23.84& 193.64&181.38 & 367.01& 823.98 & 1332.94 & 900.00 & 1880.19 & 900.00 & 2506.75\\ 
1.0 & 2.52 & 24.80 &25.55 & 287.91 & 200.17 & 406.29 & 845.70 & 1390.83 & 900.00 & 1951.20 & 900.00 & 2663.91\\ \bottomrule

    \\[-7pt]
    & \multicolumn{12}{ c }{Combinatorial Auctions} \\
    \\[-3pt]
\end{tabular}
}
\newline
\vspace{1mm}
\newline
\resizebox{\textwidth}{!}{
\begin{tabular}{@{}ccccccccccccc@{}}
\toprule
\toprule
& \multicolumn{2}{c}{$D_1$ } & \multicolumn{2}{c}{$D_2$ } & \multicolumn{2}{c}{$D_3$ } & \multicolumn{2}{c}{$D_4$ }& \multicolumn{2}{c}{$D_5$}& \multicolumn{2}{c}{$D_6$}\\ \midrule
$\beta$ & Time(s) $\downarrow$ & PD integral & Time(s) $\downarrow$ & PD integral  & Time(s)  & PD integral $\downarrow$ & Time(s) & PD integral $\downarrow$ & Time(s) & PD integral $\downarrow$ & Time(s) & PD integral $\downarrow$\\\cmidrule(r){1-3} \cmidrule(lr){4-5} \cmidrule(l){6-7}\cmidrule(l){8-9}\cmidrule(l){10-11}\cmidrule(l){12-13}
0.001 & 7.30 & 92.75 & 81.53 & 647.23 & 878.35 & 7785.46 & 900.00 & 11395.63 & 900.00 & 14972.49 & 900.00 & 23420.52 \\ 
0.01 & \textbf{6.28} & 80.67 & \textbf{73.78} &515.92 & 839.80 & \textbf{7079.13}& 900.00 &\textbf{10691.05} &900.00&\textbf{14452.71}&900.00&\textbf{21792.69}\\
0.1 & 7.82 & 88.29 & 85.10& 554.67 & 853.72 & 9785.24 & 900.00  & 11536.56 & 900.00 & 15964.53 & 900.00 & 25953.83\\ 
1.0 & 9.97& 104.50 & 97.26 & 928.31 & 900.00 & 10683.34 & 900.00 & 13985.75 & 900.00 & 17434.35 & 900.00 & 26494.63\\ \bottomrule

    \\[-7pt]
    & \multicolumn{12}{ c }{Set Covering} \\
    \\[-3pt]
\end{tabular}
\vspace{-2mm}
}
}
\end{table*}

\vspace{-1mm}
\begin{table*}[t]
{
\caption{{Sensitivity of the coefficient $\tau$. }}
\vspace{-1mm}
\label{table:tau}
\centering
\resizebox{\textwidth}{!}{
\begin{tabular}{@{}ccccccccccccc@{}}
\toprule
\toprule
& \multicolumn{2}{c}{$D_1$ } & \multicolumn{2}{c}{$D_2$ } & \multicolumn{2}{c}{$D_3$ } & \multicolumn{2}{c}{$D_4$ }& \multicolumn{2}{c}{$D_5$}& \multicolumn{2}{c}{$D_6$}\\ \midrule
$\tau$ & Time(s) $\downarrow$ & PD integral & Time(s) $\downarrow$ & PD integral  & Time(s) $\downarrow$ & PD integral & Time(s) & PD integral $\downarrow$ & Time(s) & PD integral $\downarrow$ & Time(s) & PD integral $\downarrow$\\\cmidrule(r){1-3} \cmidrule(lr){4-5} \cmidrule(l){6-7}\cmidrule(l){8-9}\cmidrule(l){10-11}\cmidrule(l){12-13}
0.1 & 1.87 & 23.99 & 17.02 & 123.22 & 188.37 & 337.92 & 832.44& 1253.78&  900.00 & 1792.62& 900.00 & 2253.59\\
0.3& 1.65 & 23.65 & 16.95 & 118.47 & 182.06 & 335.14 & 846.06& 1446.13&  900.00 & 1753.77 & 900.00 & 2164.30\\
0.5 &{\textbf{1.63}}  & 22.44 & {\textbf{16.34}}  & 117.13 & {\textbf{178.32}}  & 314.32& 817.57 & {\textbf{1154.79}} & 900.00 & \textbf{1716.55} & 900.00 & \textbf{2107.19}\\
0.7 & 1.98 & 27.01 & 17.86 & 130.11 & 193.90 & 397.65 & 826.54 & 1476.49 & 900.00 & 1864.20 & 900.00 & 2549.74 \\\bottomrule

    \\[-7pt]
    & \multicolumn{12}{ c }{Combinatorial Auctions} \\
    \\[-3pt]
\end{tabular}
}
\newline
\vspace{1mm}
\newline
\resizebox{\textwidth}{!}{
\begin{tabular}{@{}ccccccccccccc@{}}
\toprule
\toprule
& \multicolumn{2}{c}{$D_1$ } & \multicolumn{2}{c}{$D_2$ } & \multicolumn{2}{c}{$D_3$ } & \multicolumn{2}{c}{$D_4$ }& \multicolumn{2}{c}{$D_5$}& \multicolumn{2}{c}{$D_6$}\\ \midrule
$\tau$ & Time(s) $\downarrow$ & PD integral & Time(s) $\downarrow$ & PD integral  & Time(s)  & PD integral $\downarrow$ & Time(s) & PD integral $\downarrow$ & Time(s) & PD integral $\downarrow$ & Time(s) & PD integral $\downarrow$\\\cmidrule(r){1-3} \cmidrule(lr){4-5} \cmidrule(l){6-7}\cmidrule(l){8-9}\cmidrule(l){10-11}\cmidrule(l){12-13}
0.1 & 8.53 & 106.48& 77.43 & 603.78 & 863.72& 7967.44 &  900.00 & 12532.65 & 900.00& 16937.70 & 900.00 & 25506.73\\
0.3& 7.79 & 88.63 & 74.20 & 581.99 & 852.04& 7407.29 & 900.00 & 12076.84&900.00& 15272.56 & 900.00 & 22587.94\\
0.5& \textbf{6.28} & 80.67 & \textbf{73.78} &515.92 & 839.80 & \textbf{7079.13}& 900.00 &\textbf{10691.05} &900.00&\textbf{14452.71}&900.00&\textbf{21792.69}\\
0.7 & 10.77 & 94.80 & 82.66 & 707.62 & 876.28& 7835.00 & 900.00 &13573.76 &900.00& 17836.41 & 900.00 & 24072.86 \\
\bottomrule

    \\[-7pt]
    & \multicolumn{12}{ c }{Set Covering} \\
    \\[-3pt]
\end{tabular}
\vspace{-2mm}
}
}
\end{table*}


{
\subsubsection{Sensitivity Analysis on $\beta$}
Penalty parameter $\beta$ balances the adversarial reward and the distance between the augmented and original distributions. We conduct sensitivity experiments by varying $\beta$ in Table \ref{table:beta}. When $\beta$ is too small, the augmentation policy may produce overly aggressive perturbations that deviate significantly from the task-relevant structures. Conversely, a very large $\beta$ restricts the diversity of augmented instances. We find that $\beta = 0.01$ consistently yields near-optimal performance.

\subsubsection{Sensitivity Analysis on the Discriminator Threshold}
The discriminator $D$ is designed to filter out unfavorable augmented instances that exhibit excessive deviation from the original training distribution. As described in Section \ref{section: discriminator}, an augmented instance is included in training only if $D(G) > \tau$. We vary the threshold $\tau$ from 0.1 to 0.7 to investigate its effect in Table \ref{table:tau}. A moderate threshold achieves the best performance. A very low threshold allows low-quality instances into the training pool, while an excessively high threshold acts as a strict filter that eliminates most augmented samples, thereby reducing the data diversity. 

}

\begin{table*}[ht]
\caption{{The time (hours) of the four parts during the training process. Most of the additional time is spent in the branching sample collection from augmented instances. }}
\label{table: each part training time}
\centering
\begin{tabular}{@{}>{}l>{}c>{}c>{}c>{}c@{}}
\toprule
\toprule
 & GNN Policy & Augmentation  & Instance &  Sample\\
 & Training &Network Training & Augmentation& Collection\\
 \midrule
 Set Covering & 6.9 & 3.8 & 0.3 & 8.4\\
 Combinatorial Auctions & 4.7 & 2.2 & 0.2 & 4.2\\
 Capacitated Facility Location & 11.4 & 3.6 & 0.3 & 6.0\\
 Maximum Independent Set & 3.2 & 1.4 & 0.1 & 3.4\\
\bottomrule
\end{tabular}
\end{table*}

\begin{figure*}[t]
    \centering
    \begin{subfigure}{0.32\textwidth}
        \includegraphics[width=\textwidth]{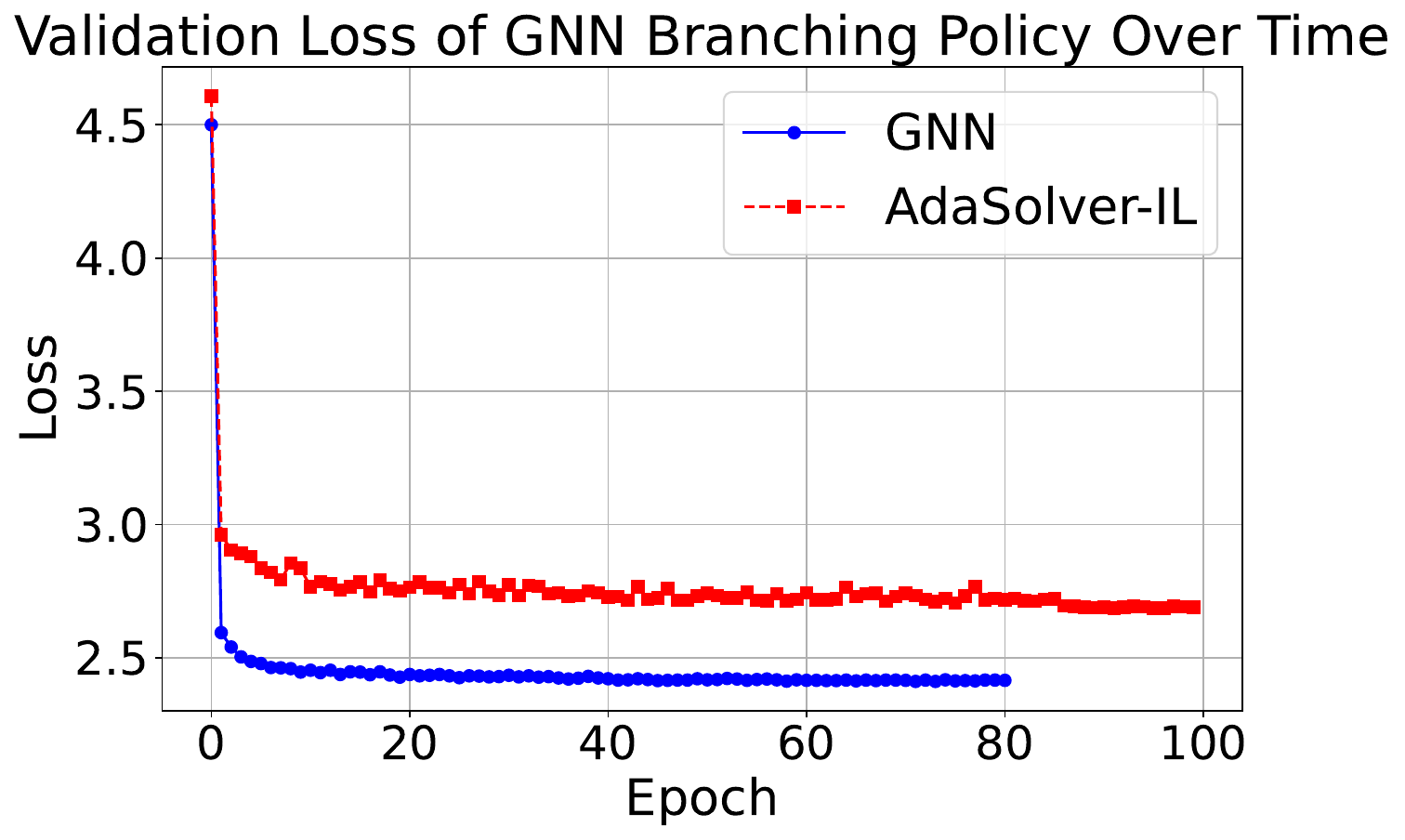}
        \vspace{-5mm}
        \label{fig: training curve IL}
    \end{subfigure}
    \begin{subfigure}{0.32\textwidth}
        \includegraphics[width=\textwidth]{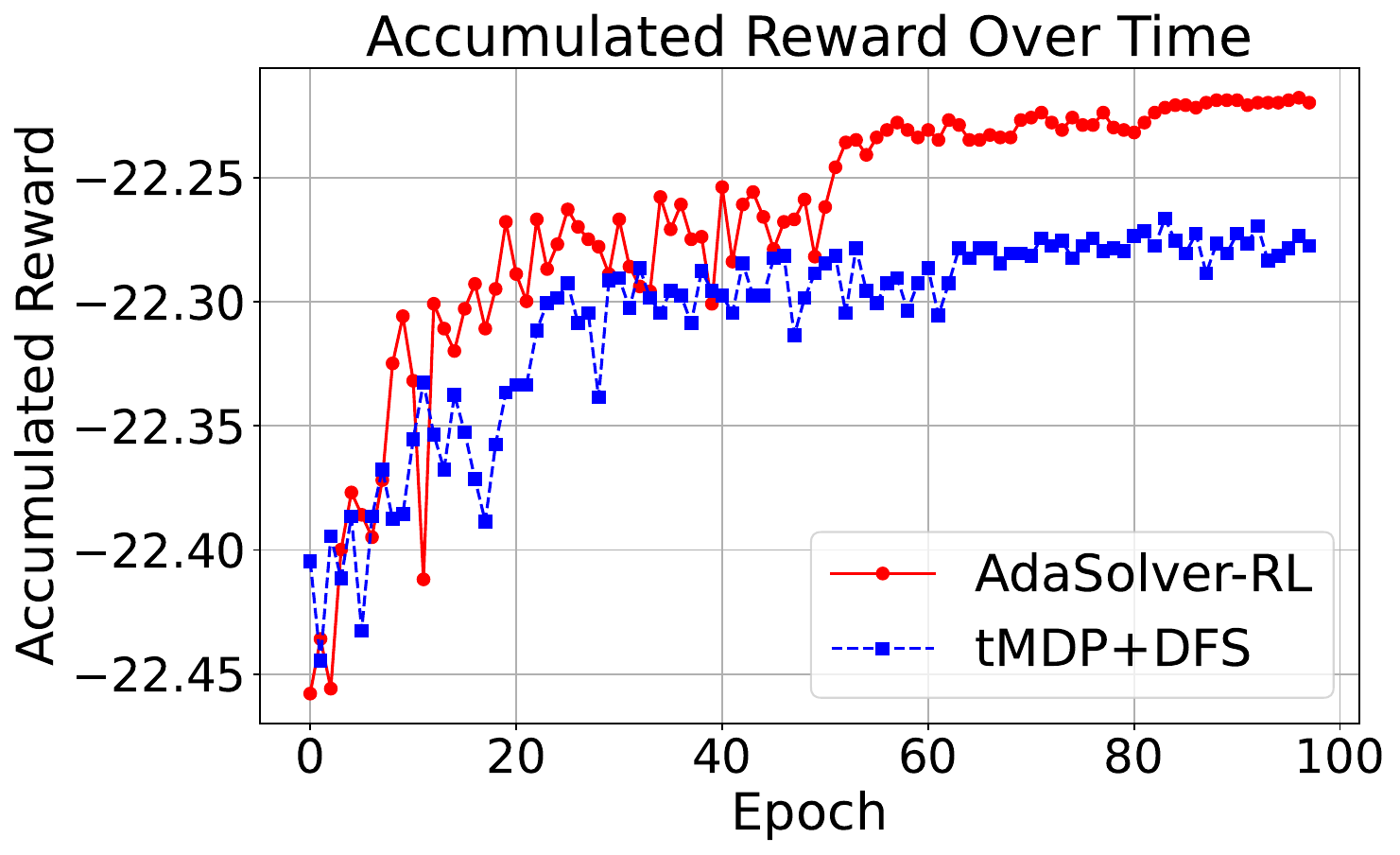}
        \vspace{-5mm}
        \label{fig: training curve RL}
    \end{subfigure}
    \begin{subfigure}{0.32\textwidth}
        \includegraphics[width=\textwidth]{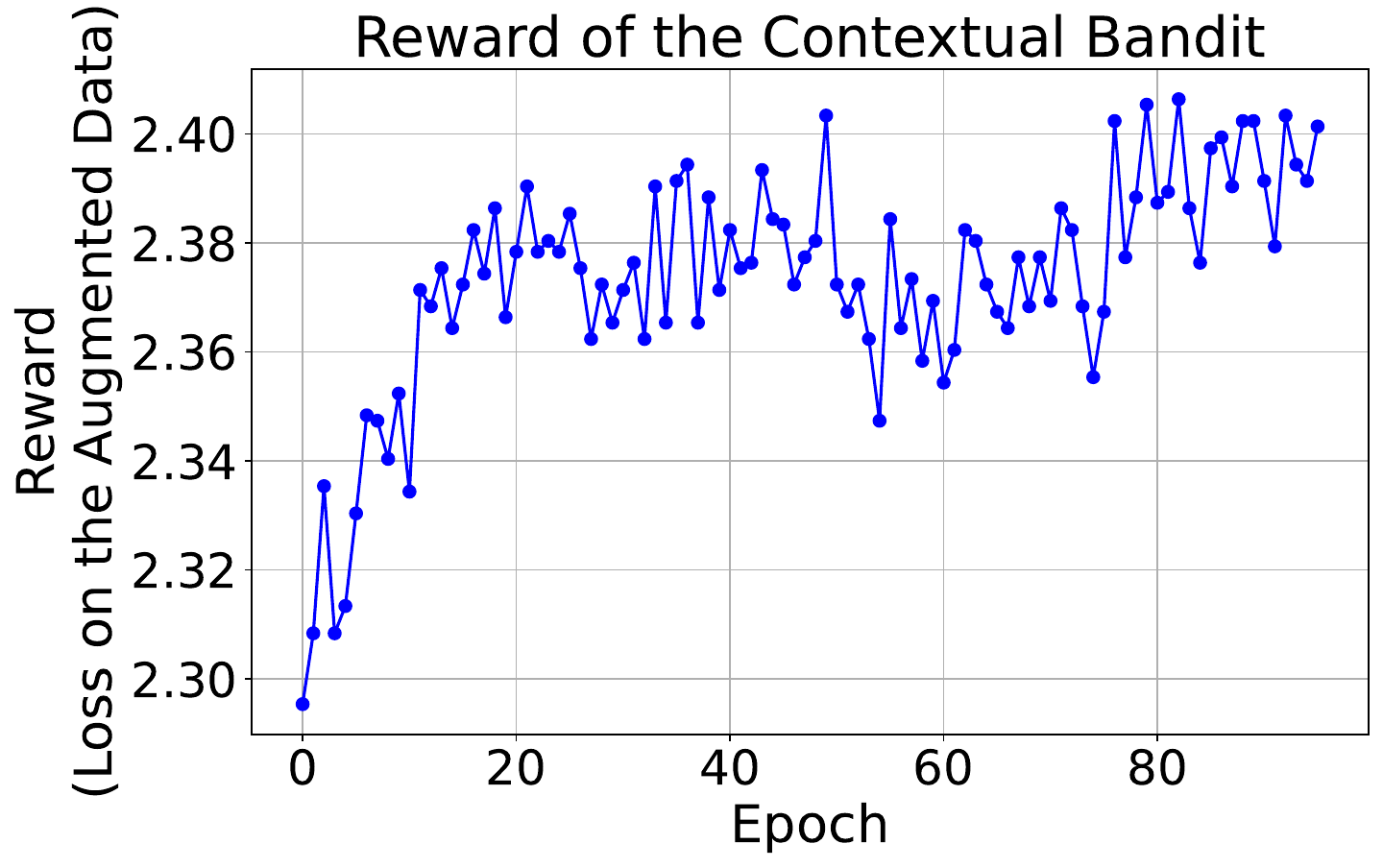}
        \vspace{-5mm}
        \label{fig: training curve bandit}
    \end{subfigure}
    \caption{{Left: the validation loss of GNN branching policy, trained with the baseline method and our AdaSolver algorithm. Middle: the accumulated reward over time in the training process of the RL-based solvers tMDP+DFS and AdaSolver-RL. The accumulated reward is the average primal-dual integral of the validation instances. Right: the reward of the contextual bandit during the training process. The reward is the loss of the GNN branching policy on the augmented data.}}
    \label{figure: stability}
\end{figure*}

\begin{table*}[t]
\caption{{We train the GNN model with random masking data augmentation and AdaSolver-IL for the same training time and compare their performance.
The baseline is denoted by RM (longer training time).
The results demonstrate that longer training time cannot bring improvement to the GNN branching policy, while the adversarial data augmentation brings a significant improvement.} }
\vspace{-1mm}
\label{table: same training time}
\centering
\resizebox{\textwidth}{!}{
\begin{tabular}{@{}>{}c>{}c>{}c>{}c>{}c>{}c>{}c>{}c>{}c>{}c>{}c>{}c>{}c@{}}
\toprule
\toprule
& \multicolumn{2}{c}{$D_1$ } & \multicolumn{2}{c}{$D_2$ } & \multicolumn{2}{c}{$D_3$ } & \multicolumn{2}{c}{$D_4$ }& \multicolumn{2}{c}{$D_5$}& \multicolumn{2}{c}{$D_6$}\\ \midrule
Method & Time(s) $\downarrow$ & PD integral & Time(s) $\downarrow$ & PD integral  & Time(s) $\downarrow$ & PD integral & Time(s) & PD integral $\downarrow$ & Time(s) & PD integral $\downarrow$ & Time(s) & PD integral $\downarrow$\\\cmidrule(r){1-3} \cmidrule(lr){4-5} \cmidrule(l){6-7}\cmidrule(l){8-9}\cmidrule(l){10-11}\cmidrule(l){12-13}
RM (longer training time)  & 1.98 & 29.53 & 19.52 & 148.59 & 193.67 & 399.43 & 836.21 & 1502.86 & 900.00& 1821.29 & 900.00 & 2562.30 \\
AdaSolver-IL & \textbf{1.63}  & {22.44} & \textbf{16.34}  & {117.13} & \textbf{178.32}  & 314.32& 817.57 & \textbf{1154.79} & 900.00 & \textbf{1716.55} & 900.00 & \textbf{2107.19}\\ \bottomrule

    \\[-7pt]
    & \multicolumn{12}{ c }{Combinatorial Auctions} \\
    \\[-3pt]
\end{tabular}
}
\newline
\vspace{1mm}
\newline
\resizebox{\textwidth}{!}{
\begin{tabular}{@{}>{}c>{}c>{}c>{}c>{}c>{}c>{}c>{}c>{}c>{}c>{}c>{}c>{}c@{}}
\toprule
\toprule
& \multicolumn{2}{c}{$D_1$ } & \multicolumn{2}{c}{$D_2$ } & \multicolumn{2}{c}{$D_3$ } & \multicolumn{2}{c}{$D_4$ }& \multicolumn{2}{c}{$D_5$}& \multicolumn{2}{c}{$D_6$}\\ \midrule
Method & Time(s) $\downarrow$ & PD integral & Time(s) $\downarrow$ & PD integral  & Time(s)  & PD integral $\downarrow$ & Time(s) & PD integral $\downarrow$ & Time(s) & PD integral $\downarrow$ & Time(s) & PD integral $\downarrow$\\\cmidrule(r){1-3} \cmidrule(lr){4-5} \cmidrule(l){6-7}\cmidrule(l){8-9}\cmidrule(l){10-11}\cmidrule(l){12-13}
RM (longer training time) & 8.94 & 92.49 & 90.85 & 623.20 & 844.21 & 7192.82 & 900.00 & 11574.76 & 900.00 & 15386.80 & 900.00 & 23438.58\\
AdaSolver-IL & \textbf{6.28} & {80.67} & \textbf{73.78} &{515.92} & 839.80 & \textbf{7079.13}& 900.00 &\textbf{10691.05} &900.00&\textbf{14452.71}&900.00&\textbf{21792.69} \\ \bottomrule

    \\[-7pt]
    & \multicolumn{12}{ c }{Set Covering} \\
    \\[-3pt]
\end{tabular}
}
\end{table*}

{
\begin{table}[t]
\caption{{The training time (hours) of the GNN baseline and our method. 
The adversarial data augmentation process increases the training time of our method, but it is critical to improve the generalization ability.}}
\vspace{-1mm}
\label{table: training time}
\centering
\begin{tabular}{@{}>{}l>{}c>{}c@{}}
\toprule
\toprule
 & GNN & AdaSolver-IL\\
 \midrule
 Set Covering & 7.2 & 19.4\\
 Combinatorial Auctions & 5.7 & 11.3\\
 Capacitated Facility Location & 12.5 & 21.3\\
 Maximum Independent Set & 3.8 & 8.1\\
\bottomrule
\end{tabular}
\end{table}
\subsection{Analysis of the Training Time}
We present the training times for the GNN and our AdaSolver in Table \ref{table: training time}. 
The results indicate that AdaSolver requires twice the training time of the GNN baseline.
This training time encompasses the GNN policy training, augmentation network training, augmented instance generation, and the collection of augmented branching samples. We detail the time allocated to each of these components in Table \ref{table: each part training time}.
The additional time is primarily devoted to collecting branching samples for the augmented instances, while the augmentation network training and instance generation processes are relatively efficient. 

To ensure a fair comparison, we train the GNN baseline using random masking data augmentation for the same duration as AdaSolver (referred to as RM with a longer training time). This approach allows us to explore the baseline's potential for improved generalization with extended training. We maintain consistent training times and select the model that achieves the best validation loss for testing. The results are presented in Table \ref{table: same training time}.
Our findings indicate that increasing training time does not lead to significant improvements in model performance. In contrast, our method clearly outperforms the RM under the same training conditions. 
Existing methods often face performance bottlenecks, especially when training data is limited. 
AdaSolver addresses this challenge by generating difficult training examples throughout the training process, effectively mitigating these issues.
Therefore, the longer training duration combined with adversarial augmentation proves to be beneficial.

\begin{table*}[t]
\caption{{We compare the performance of adversarial augmentation and a fixed data augmentation network (fixed-gen AdaSolver). 
The results demonstrate that the adversarial augmentation brings significant improvement in the model performance and generalization ability. 
The fixed data augmentation policy fails to consistently generate adaptive hard instances for the branching policy. 
}}
\vspace{-1mm}
\label{table: adversarial}
\centering
\resizebox{\textwidth}{!}{
\begin{tabular}{@{}>{}c>{}c>{}c>{}c>{}c>{}c>{}c>{}c>{}c>{}c>{}c>{}c>{}c@{}}
\toprule
\toprule
& \multicolumn{2}{c}{$D_1$ } & \multicolumn{2}{c}{$D_2$ } & \multicolumn{2}{c}{$D_3$ } & \multicolumn{2}{c}{$D_4$ }& \multicolumn{2}{c}{$D_5$}& \multicolumn{2}{c}{$D_6$}\\ \midrule
Method & Time(s) $\downarrow$ & PD integral & Time(s) $\downarrow$ & PD integral  & Time(s) $\downarrow$ & PD integral & Time(s) & PD integral $\downarrow$ & Time(s) & PD integral $\downarrow$ & Time(s) & PD integral $\downarrow$\\\cmidrule(r){1-3} \cmidrule(lr){4-5} \cmidrule(l){6-7}\cmidrule(l){8-9}\cmidrule(l){10-11}\cmidrule(l){12-13}
fixed-gen AdaSolver  & 1.78 & 26.65 & 25.84 & 183.12 & 289.45 & 399.79 & 832.93 & 1557.78 & 900.00 & 2004.13 & 900.00 & 2590.07\\
AdaSolver-IL & \textbf{1.63}  & {22.44} & \textbf{16.34}  & {117.13} & \textbf{178.32}  & 314.32& 817.57 & \textbf{1154.79} & 900.00 & \textbf{1716.55} & 900.00 & \textbf{2107.19}\\ \bottomrule

    \\[-7pt]
    & \multicolumn{12}{ c }{Combinatorial Auctions} \\
    \\[-3pt]
\end{tabular}
}
\newline
\vspace{1mm}
\newline
\resizebox{\textwidth}{!}{
\begin{tabular}{@{}>{}c>{}c>{}c>{}c>{}c>{}c>{}c>{}c>{}c>{}c>{}c>{}c>{}c@{}}
\toprule
\toprule
& \multicolumn{2}{c}{$D_1$ } & \multicolumn{2}{c}{$D_2$ } & \multicolumn{2}{c}{$D_3$ } & \multicolumn{2}{c}{$D_4$ }& \multicolumn{2}{c}{$D_5$}& \multicolumn{2}{c}{$D_6$}\\ \midrule
Method & Time(s) $\downarrow$ & PD integral & Time(s) $\downarrow$ & PD integral  & Time(s)  & PD integral $\downarrow$ & Time(s) & PD integral $\downarrow$ & Time(s) & PD integral $\downarrow$ & Time(s) & PD integral $\downarrow$\\\cmidrule(r){1-3} \cmidrule(lr){4-5} \cmidrule(l){6-7}\cmidrule(l){8-9}\cmidrule(l){10-11}\cmidrule(l){12-13}
fixed-gen AdaSolver  & 7.81 & 89.88 & 93.17 & 688.51 & 871.38 & 7730.19 & 900.00 & 16052.09 & 900.00 & 17963.06 & 900.00 & 24421.63\\
AdaSolver-IL & \textbf{6.28} & {80.67} & \textbf{73.78} &{515.92} & 839.80 & \textbf{7079.13}& 900.00 &\textbf{10691.05} &900.00&\textbf{14452.71}&900.00&\textbf{21792.69}\\ \bottomrule

    \\[-7pt]
    & \multicolumn{12}{ c }{Set Covering} \\
    \\[-3pt]
\end{tabular}
}
\end{table*}

\begin{table*}[t]
\caption{{The comparison of the commercial solver. We use the ML branching interface and build our method on the OptVerse commerical solver (AdaSolver-OptVerse). We use the imitation learning for model training. We find our method can significantly improve the performance of the commercial solver by using our trained branching model.  
}}
\vspace{-1mm}
\label{table:commercial solver}
\centering
\resizebox{\textwidth}{!}{
\begin{tabular}{@{}>{}c>{}c>{}c>{}c>{}c>{}c>{}c>{}c>{}c>{}c>{}c>{}c>{}c@{}}
\toprule
\toprule
& \multicolumn{2}{c}{$D_1$ } & \multicolumn{2}{c}{$D_2$ } & \multicolumn{2}{c}{$D_3$ } & \multicolumn{2}{c}{$D_4$ }& \multicolumn{2}{c}{$D_5$}& \multicolumn{2}{c}{$D_6$}\\ \midrule
Method & Time(s) $\downarrow$ & PD integral & Time(s) $\downarrow$ & PD integral  & Time(s) $\downarrow$ & PD integral & Time(s) & PD integral $\downarrow$ & Time(s) & PD integral $\downarrow$ & Time(s) & PD integral $\downarrow$\\\cmidrule(r){1-3} \cmidrule(lr){4-5} \cmidrule(l){6-7}\cmidrule(l){8-9}\cmidrule(l){10-11}\cmidrule(l){12-13}
OptVerse  &2.54 & 30.92 & 25.64 & 148.98 & 231.38 & 437.16 & 867.20 & 1573.16 & 900.00 & 2141.69 & 900.00 & 2532.76\\
AdaSolver-IL & {1.63}  & {22.44} & {16.34}  & {117.13} & {178.32}  & 314.32& 817.57 & {1154.79} & 900.00 & {1716.55} & 900.00 & {2107.19}\\ 
AdaSolver-OptVerse & 1.59 & 23.58 & 16.21& 121.42 & 176.04 & 355.29 & 831.06 & 1021.63 & 900.00 & 1620.71 & 900.00 & 2008.55 \\ \bottomrule

    \\[-7pt]
    & \multicolumn{12}{ c }{Combinatorial Auctions} \\
    \\[-3pt]
\end{tabular}
}
\newline
\vspace{1mm}
\newline
\resizebox{\textwidth}{!}{
\begin{tabular}{@{}>{}c>{}c>{}c>{}c>{}c>{}c>{}c>{}c>{}c>{}c>{}c>{}c>{}c@{}}
\toprule
\toprule
& \multicolumn{2}{c}{$D_1$ } & \multicolumn{2}{c}{$D_2$ } & \multicolumn{2}{c}{$D_3$ } & \multicolumn{2}{c}{$D_4$ }& \multicolumn{2}{c}{$D_5$}& \multicolumn{2}{c}{$D_6$}\\ \midrule
Method & Time(s) $\downarrow$ & PD integral & Time(s) $\downarrow$ & PD integral  & Time(s)  & PD integral $\downarrow$ & Time(s) & PD integral $\downarrow$ & Time(s) & PD integral $\downarrow$ & Time(s) & PD integral $\downarrow$\\\cmidrule(r){1-3} \cmidrule(lr){4-5} \cmidrule(l){6-7}\cmidrule(l){8-9}\cmidrule(l){10-11}\cmidrule(l){12-13}
OptVerse  & 9.83 & 111.87 & 85.19 & 759.14 & 867.40 & 7137.13 & 900.00 & 11291.57 & 900.00 & 14527.96 & 900.00 & 23160.64 \\
AdaSolver-IL & {6.28} & {80.67} & {73.78} &{515.92} & 839.80 &{7079.13}& 900.00 &{10691.05} &900.00&{14452.71}&900.00&{21792.69}\\
AdaSolver-OptVerse & 6.12 & 78.35 & 72.51 & 498.23 & 849.70 & 7005.02 & 900.00 & 10392.46 & 900.00 & 14262.37 & 900.00 & 20042.03 \\ \bottomrule

    \\[-7pt]
    & \multicolumn{12}{ c }{Set Covering} \\
    \\[-3pt]
\end{tabular}
}
\end{table*}

\subsection{Training Stability}
We present the training curves in Figure \ref{figure: stability} to study the training stability of our proposed method.
First, we assess the training stability of the GNN branching network. On the left side of Figure \ref{figure: stability}, we show the validation loss of the GNN branching policy trained using IL-based methods. Both the GNN baseline and AdaSolver-IL demonstrate a consistently decreasing validation loss until convergence.
It is important to note that the validation dataset exclusively contains instances from the $D_1$ distribution.
Although AdaSolver-IL exhibits a higher validation loss at convergence in $D_1$, it consistently outperforms the GNN baseline across distributions $D_1$ to $D_6$. 
This suggests that AdaSolver mitigates overfitting to a single distribution and enhances generalization across multiple instance distributions.

Next, we report the accumulated reward over time for RL-based methods in the middle of Figure \ref{figure: stability}. The accumulated reward represents the average negative primal-dual integral of the validation instances. Throughout the training process, the accumulated rewards for both the baseline tMDP+DFS and our proposed AdaSolver-RL increase, indicating stable policy improvement. Notably, AdaSolver-RL achieves a higher accumulated reward.

Finally, on the right side of Figure \ref{figure: stability}, we analyze the training stability of the augmented network. We report the rewards from the contextual bandit as training progresses. These rewards reflect the loss of the GNN branching policy on the augmented data. Initially, the rewards increase during training but then stabilize at a high level. The initial rise is attributed to the augmentation policy generating increasingly challenging instances for the branching policy, leading to higher loss values. As training continues, the GNN branching policy has stronger capabilities for handling these hard instances. Ultimately, the branching policy and the augmented network converge to an equilibrium point.
}



{
\subsection{The Effectiveness of Adversarial Training}
To demonstrate the effectiveness of adversarial augmentation, we compare it with a fixed data augmentation policy. Specifically, we implement a variant of AdaSolver called fixed-gen AdaSolver.
The fixed-gen AdaSolver trains an augmentation network to approximate the distribution of hard instances. This augmentation network undergoes five steps of adversarial training. Subsequently, fixed-gen AdaSolver utilizes the instances generated by the fixed augmentation network for training.
We ensure that both methods generate the same amount of training data and compare their performance. The results in Table \ref{table: adversarial} indicate that adversarial augmentation consistently enhances the performance of the solver, outperforming the data generation from a worse distribution.
}

{
\subsection{Comparison with the Commercial Solver}
We built AdaSolver on another commercial MILP solver, OptVerse from Huawei \cite{optverse}.
We do not use alternative commercial solvers, such as Gurobi and Cplex, as they do not provide ML interfaces for users to integrate their trained ML models into the solvers.
The developers of OptVerse provide an interface that allows us to customize the branching process. The results of this comparison are shown in Table \ref{table:commercial solver}, where we evaluate AdaSolver—built on OptVerse—against the default commercial solver on the benchmarks. The findings demonstrate a significant improvement in solving performance.
}

\section{Conclusion}
This paper focuses on the generalization and robustness issues of the learning-based MILP solvers under restrictive training settings.
Learning-based solvers often suffer from severe performance degradation in unseen MILP instances.
To address this challenge, we propose a robust optimization framework (AdaSolver) for learning-based B\&B solvers to promote the diversity of instance graph structures in training distributions through adversarial instance augmentation.
Specifically, we formulate the non-differentiable learning problem of the augmentation policy as a contextual bandit problem, enabling efficient gradient-based adversarial training for the learning-based solvers and the augmentation policy.
Experiments show that AdaSolver significantly improves the generalization and sample efficiency of the learning-based B\&B solvers on five well-recognized benchmarks.


%

\ifCLASSOPTIONcompsoc
  \section*{Acknowledgments}
\else
  \section*{Acknowledgment}
\fi
The authors would like to thank all the anonymous reviewers for their insightful comments. This work was supported by the National Key R\&D  Program of China under contract 2022ZD0119801, National Nature  Science Foundations of China grants U19B2026, U19B2044, 61836011, 62021001, and 61836006.

\ifCLASSOPTIONcaptionsoff
  \newpage
\fi
\bibliographystyle{IEEEtran}
\bibliography{library}




%
%
%

%

\newpage
\begin{IEEEbiography}[{\includegraphics[width=1in,height=1.25in,clip,keepaspectratio]{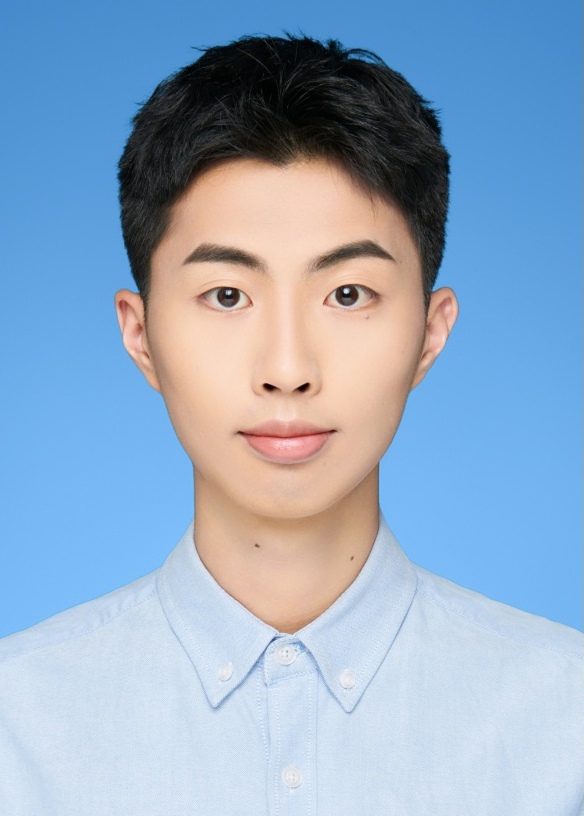}}]{Haoyang Liu}
received the B.Sc. degree in School of Mathematics from University of Science and Technology of China, Hefei, China, in 2023. He is currently a Ph.D. candidate in Department of Electronic Engineering and Information Science at University of Science and Technology of China, Hefei, China. His research interests include reinforcement learning and learning to optimize.
\end{IEEEbiography}

\begin{IEEEbiography}[{\includegraphics[width=1in,height=1.25in,clip,keepaspectratio]{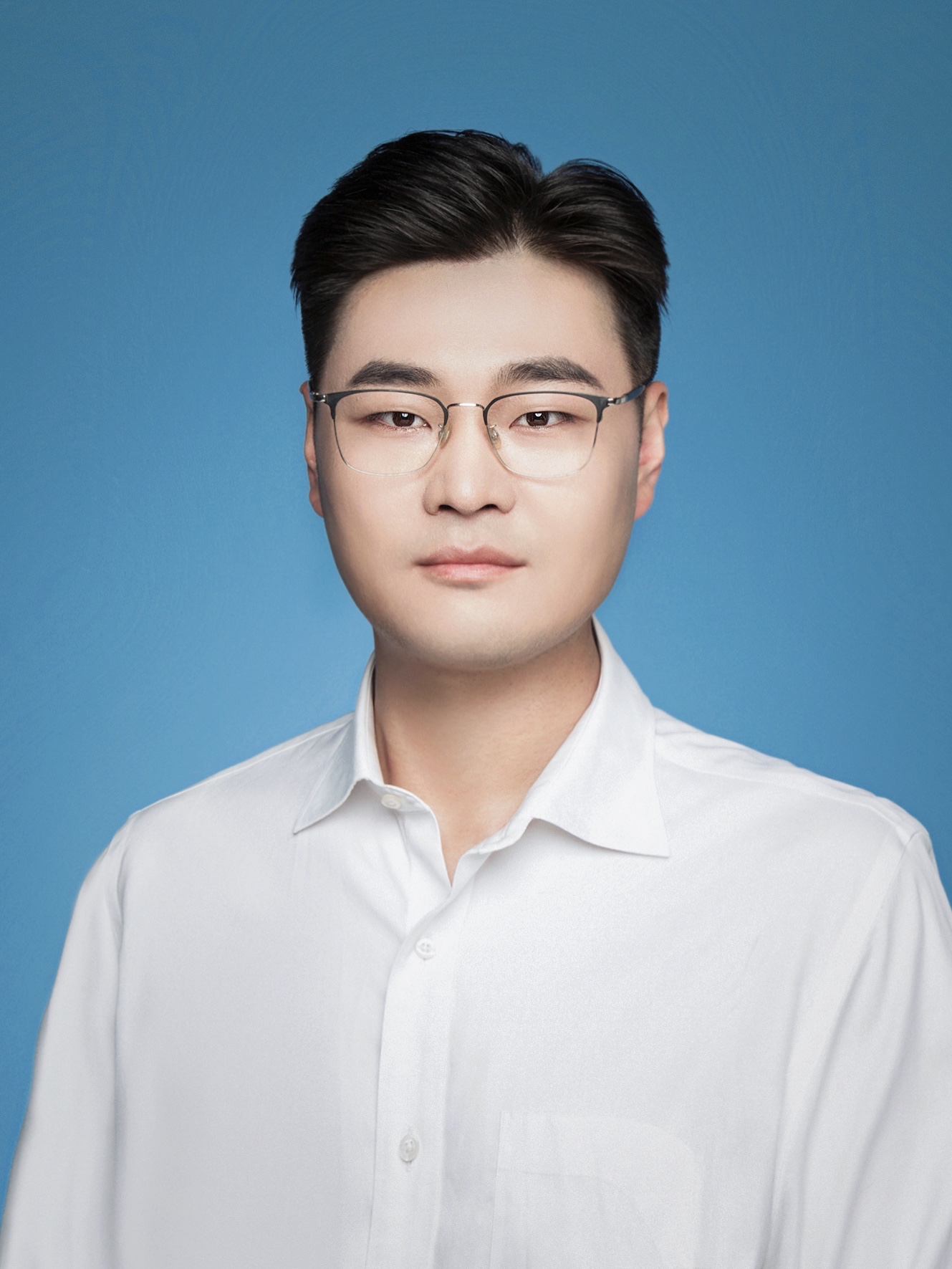}}]{Jie Wang} (Senior Member of IEEE, Associate Editor of IEEE TPAMI)
received the B.Sc. degree in electronic information science and technology from University of Science and Technology of China, Hefei, China, in 2005, and the Ph.D. degree in computational science from the Florida State University, Tallahassee, FL, in 2011. He is currently a Professor in the Department of Electronic Engineering and Information Science at University of Science and Technology of China, Hefei, China. His research interests include reinforcement learning, knowledge graph, large-scale optimization, deep learning, etc. 
\end{IEEEbiography}

\begin{IEEEbiography}[{\includegraphics[width=1in,height=1.25in,clip,keepaspectratio]{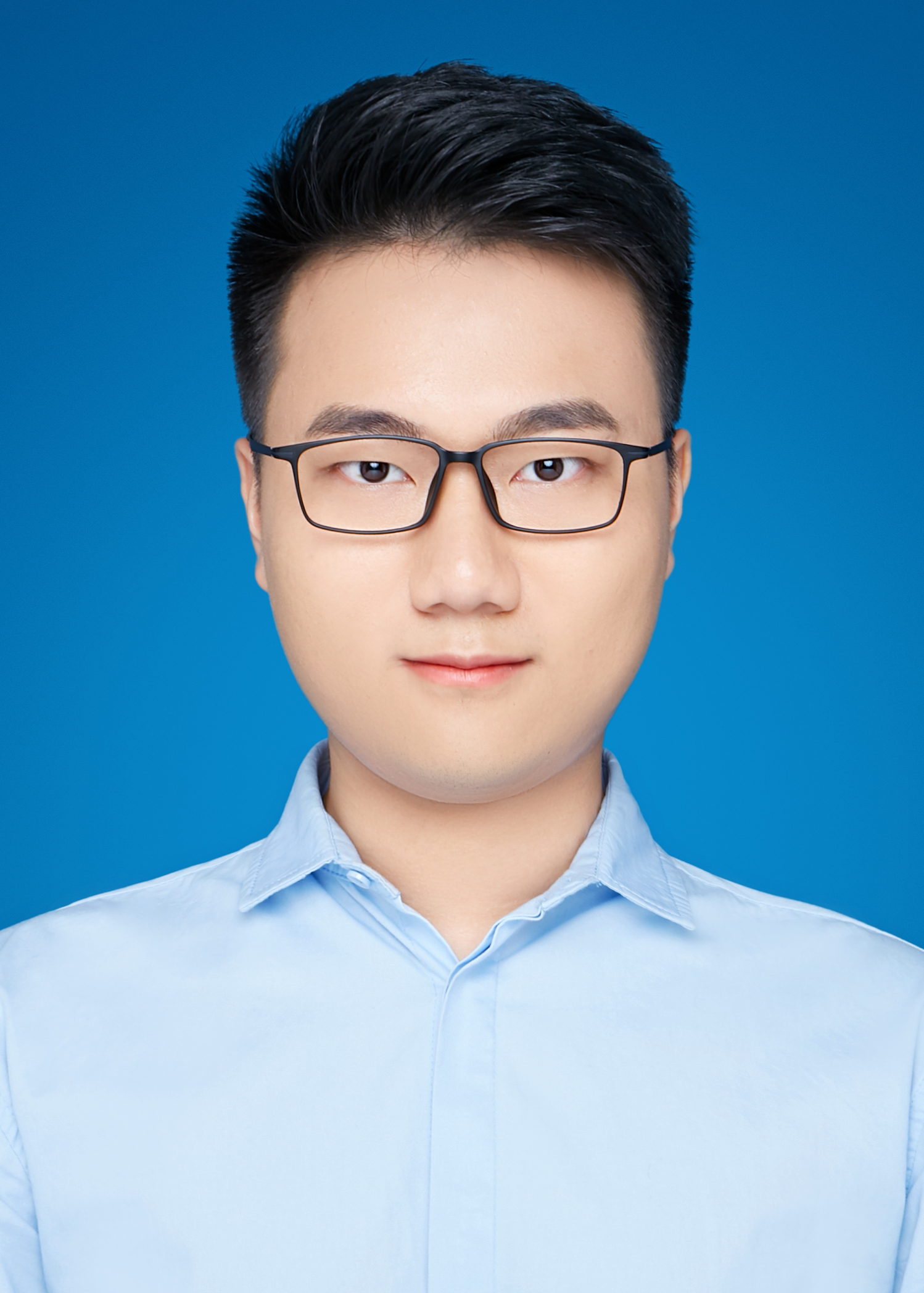}}]{Yufei Kuang}
received the BSc degree in statistics from Nanjing University, Nanjing, China, in 2020. He is currently working toward the PhD degree with the Department of Electronic Engineering and Information Science, University of Science and Technology of China, Hefei, China. His research interests include reinforcement learning, learning to optimize, and large language models for reasoning.
\end{IEEEbiography}

\begin{IEEEbiography}[{\includegraphics[width=1in,height=1.25in,clip,keepaspectratio]{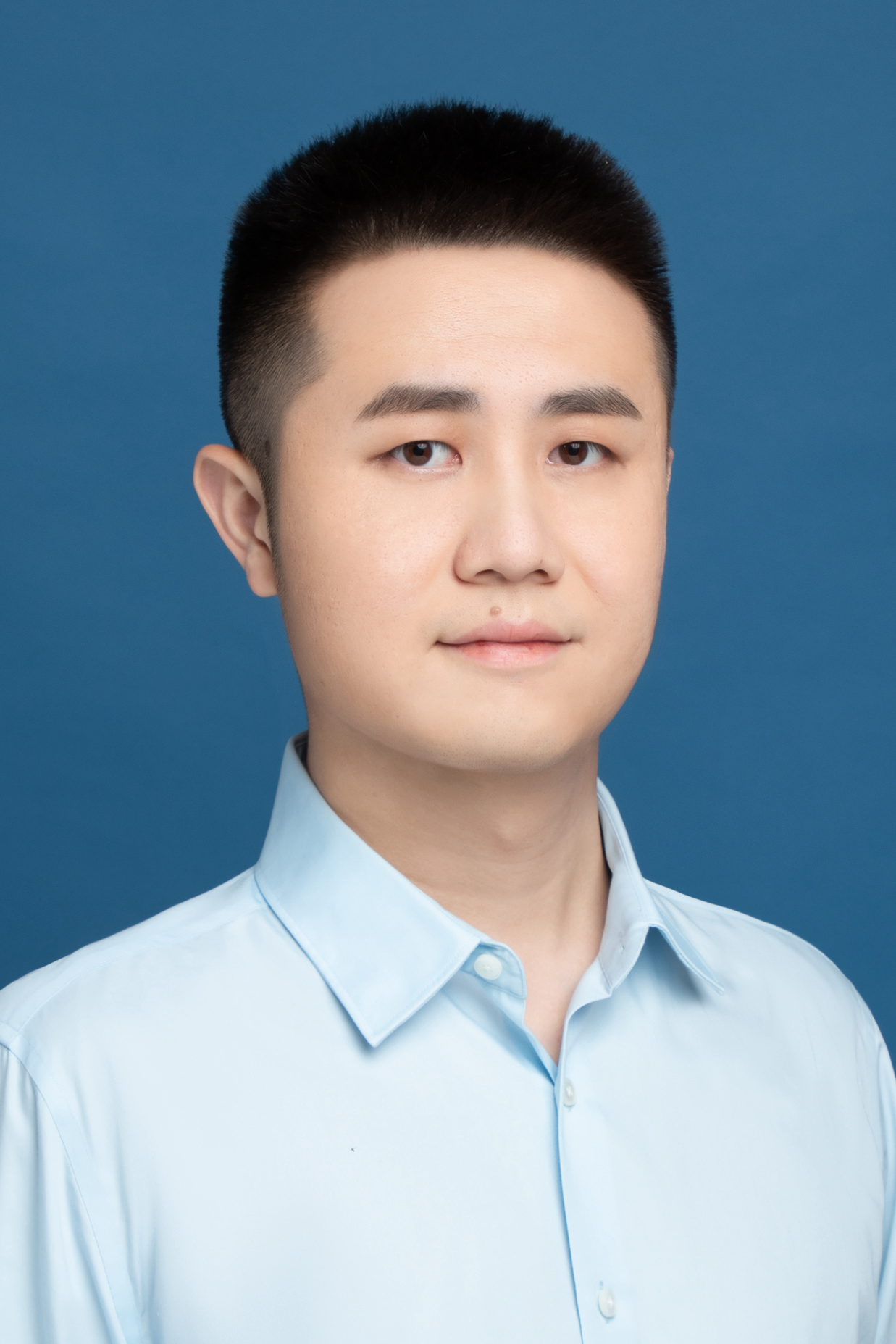}}]{Xijun Li}
is an assistant professor in Shanghai Jiao Tong University and a faculty member of the Shanghai Key Labraotry of Scalable Computing and Systems. Previously, he has served as a princinpal researcher in Huawei Noah’s Ark Lab between 2018 and 2024. He has received his Ph.D. degree from the University of Science and Technology of China in March 2024 (HUAWEI-USTC Joint Ph.D. Program). He has published 30+ papers on top peer-reviewed conferences and journals (such as TPAMI, NeurIPS, ICLR, ICML, KDD, ICDE, SIGMOD, DAC, etc.) and served as Area Chair (AC) in top-tier AI conferences (such as ICLR and NeurIPS). His recent research interests focus on Learning to Optimization (L2O) and Large Language Model for Optimization/Reasoning. Last but not least, he was one of core developers of the Huawei Cloud OptVerse AI Solver and Huawei Pangu Large Language Models.
\end{IEEEbiography}

\begin{IEEEbiography}[{\includegraphics[width=1in,height=1.25in,clip,keepaspectratio]{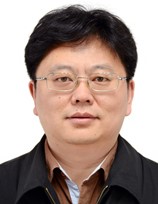}}]{Yongdong Zhang} (Senior Member of IEEE)
received the Ph.D. degree in electronic engineering from Tianjin University, Tianjin, China, in 2002. He is currently a Professor at the University of Science and Technology of China. He has authored more than 100 refereed journal articles and conference papers. His current research interests include multimedia content analysis and understanding, multimedia content security, video encoding, and streaming media technology. He serves as an Editorial Board Member of Multimedia Systems journal and Neurocomputing. He was the recipient of the Best Paper Award in PCM 2013, ICIMCS 2013, and ICME 2010, and the Best Paper Candidate in ICME 2011. 
\end{IEEEbiography}

\begin{IEEEbiography}[{\includegraphics[width=1in,height=1.25in,clip,keepaspectratio]{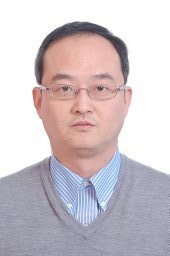}}]{Feng Wu}
(Fellow of IEEE)
received the B.Sc. degree in electronic engineering from Xidian University, Xi’an, China, in 1992, and received the M.Sc. and Ph.D. degrees from the Harbin Institute of Technology, Harbin, China, in 1996 and 1999, respectively.
He is now a Professor and Vice President at the University of Science and Technology of China, Hefei, China. Previously, he was a Principle Researcher and Research Manager with Microsoft Research Asia, Beijing, China.
His research interests include image and video compression, media communication, and media analysis and synthesis. 
He has authored or co-authored over 200 high-quality articles. His 15 techniques have been adopted into international video coding standards. 
He serves or had served as the Editor-in-Chief for IEEE Transactions on Circuits and Systems for Video Technology (TCSVT) and as an Associate Editor for IEEE Transactions on Image Processing (TIP) and IEEE Transactions on Multimedia. He also serves as General Chair in ICME 2019, TPC Chair in MMSP
2011, VCIP 2010, and PCM 2009. He received the IEEE CAS Mac Van Valkenburg Award in 2021, the best paper awards in IEEE TCSVT 2009, VCIP 2016, PCM 2008, and VCIP 2007, and the Best Associate Editor Award of IEEE Transactions on Image Processing (TIP) in 2018. 
\end{IEEEbiography}






\clearpage
\appendices

\section{Notation}
{
To better understand, we summarize and provide some key notations in Table \ref{table: notation}.
\begin{table}[htbp]
\centering
\caption{Notations used in our paper.}
\label{table: notation}
{\begin{tabular}{>{}l>{}l}
\toprule
Notation & Explanation \\
\toprule
$\mathcal{G}$ &The instance graph space \\
$\mathcal{A}$ &The candidate branching variable set \\
$R$ &Reward function in branching MDP \\
$\mathcal{T}$ &Transition function in branching MDP \\
$G$ & The instance graph\\
$G_t$ & The branching sample\\
$W$ & The constraint nodes in the graph\\
$V$ & The variable nodes in the graph\\
$p_{\text{tr}}$ & Training instance distribution\\
$p_{\text{te}}$ & Testing instance distribution\\
$\tilde{p}$ & Augmented instance distribution\\
$\mathcal{D}_{\text{IL}}$ & Dataset for imitation learning\\
$\mathcal{D}_{\text{RL}}$ & Dataset for reinforcement learning\\
\midrule
$\pi$ & The branching policy\\
$\pi_\theta$ & The parameterized policy\\
$c$ & The cost function\\
$\mathcal{W}_{\mathcal{G}}$ & Graph Wasserstein distance\\
$\mathcal{J}_p(\pi)$ & The loss on distribution $p$ of policy $\pi$\\
$\hat{\mathcal{J}}_p(\pi)$ & The empirical risk on distribution $p$ of policy $\pi$\\
\midrule
$\widetilde{\mathcal{A}}$ & Action space in contextual bandit\\
$\tilde{r}$ & Reward in contextual bandit\\
$\phi$ & Augmented network\\
$\Rad$ & Rademacher complexity\\
$\widetilde{\Rad}$ & Rademacher complexity on the augmented data\\
 \hline

\end{tabular}}
\end{table}
}

{
\section{Generalization to Cutting and Other Tasks}
While the experimental evaluation in this paper primarily focuses on the branching module, the proposed adversarial instance augmentation framework is inherently decoupled from specific solver tasks. Given the shared architectural paradigm, integrating AdaSolver with other critical solver components, such as cut and node selection, is straightforward. Most contemporary learning-based policies for branching, cut selection, and node selection employ a GNN backbone to encode the bipartite graph representation of MILP instances into a latent space. Since both the adversarial augmentation policy and the discriminator operate directly on the input graph structure rather than relying on task-specific labels, the augmented instances generated by AdaSolver can be seamlessly utilized to train any GNN-based solver component.

\subsection{Cut Selection}
Cut selection involves identifying a subset of valid inequalities to tighten the linear relaxation of a MILP instance \cite{wanglearning}. By incorporating adversarial instance augmentation during training, variables and constraints can be perturbed to create adversarial scenarios. This forces the cut selection policy to capture robust geometric properties of the cuts, rather than overfitting to spurious correlations within a narrow training distribution. We integrate AdaSolver into the cut selection framework proposed by \cite{wanglearning}. As shown in Table \ref{table:cutting plane}, our approach successfully enhances the performance of the underlying baseline method.

\subsection{Node Selection}
Similarly, adversarial instance augmentation enables node selection policies to generalize more effectively to out-of-distribution instances. We apply AdaSolver to the GNN-based node selection method introduced by \cite{NEURIPS2022_cf5bb188}. The evaluation results, presented in Table \ref{table:node plane}, clearly demonstrate that AdaSolver effectively boosts the generalization capabilities and overall performance of the baseline node selection policy.
}

\begin{table*}[t]
{
\caption{{Experiments on cutting plane selection. }}
\vspace{-1mm}
\label{table:cutting plane}
\centering
\resizebox{\textwidth}{!}{
\begin{tabular}{@{}ccccccccccccc@{}}
\toprule
\toprule
& \multicolumn{2}{c}{$D_1$ } & \multicolumn{2}{c}{$D_2$ } & \multicolumn{2}{c}{$D_3$ } & \multicolumn{2}{c}{$D_4$ }& \multicolumn{2}{c}{$D_5$}& \multicolumn{2}{c}{$D_6$}\\ \midrule
Method & Time(s) $\downarrow$ & PD integral & Time(s) $\downarrow$ & PD integral  & Time(s) $\downarrow$ & PD integral & Time(s) & PD integral $\downarrow$ & Time(s) & PD integral $\downarrow$ & Time(s) & PD integral $\downarrow$\\\cmidrule(r){1-3} \cmidrule(lr){4-5} \cmidrule(l){6-7}\cmidrule(l){8-9}\cmidrule(l){10-11}\cmidrule(l){12-13}
SCIP & 2.84  & 32.20 & 23.20  & 156.62 & 224.66  & {539.25} & 838.77 & 1622.46 & 900.00 & 2313.35 & 900.00 &  2657.51\\
GNN & 2.05 & 28.49 & 19.04 & 135.73 & 190.54& 406.73 & 833.50 & 1403.58 & 900.00 & 1895.70 & 900.00 & 2355.29\\ 
Ours &\textbf{1.76} & 26.05 & \textbf{16.96} & 123.56 & \textbf{188.03} & 391.38 & 828.39  & \textbf{1384.40} & 900.00 & \textbf{1820.37} & 900.00 & \textbf{2317.41} \\ \bottomrule

    \\[-7pt]
    & \multicolumn{12}{ c }{Combinatorial Auctions} \\
    \\[-3pt]
\end{tabular}
}

}
\end{table*}

\begin{table*}[t]
{
\caption{{Experiments on node selection. }}
\vspace{-1mm}
\label{table:node plane}
\centering
\resizebox{\textwidth}{!}{
\begin{tabular}{@{}ccccccccccccc@{}}
\toprule
\toprule
& \multicolumn{2}{c}{$D_1$ } & \multicolumn{2}{c}{$D_2$ } & \multicolumn{2}{c}{$D_3$ } & \multicolumn{2}{c}{$D_4$ }& \multicolumn{2}{c}{$D_5$}& \multicolumn{2}{c}{$D_6$}\\ \midrule
Method & Time(s) $\downarrow$ & PD integral & Time(s) $\downarrow$ & PD integral  & Time(s) $\downarrow$ & PD integral & Time(s) & PD integral $\downarrow$ & Time(s) & PD integral $\downarrow$ & Time(s) & PD integral $\downarrow$\\\cmidrule(r){1-3} \cmidrule(lr){4-5} \cmidrule(l){6-7}\cmidrule(l){8-9}\cmidrule(l){10-11}\cmidrule(l){12-13}
SCIP & 2.84  & 32.20 & 23.20  & 156.62 & 224.66  & {539.25} & 838.77 & 1622.46 & 900.00 & 2313.35 & 900.00 &  2657.51\\
GNN & 2.66 & 32.13 & 20.42& 144.07 & 215.40 & 521.50 & 826.29 & 1541.76 & 900.00 & 2210.06 & 900.00 & 2510.61\\ 
Ours & \textbf{1.99} & 29.51 & \textbf{18.00} & 142.66 & 200.58 & \textbf{517.12} & 819.58 & \textbf{1436.10} & 900.00 & \textbf{2195.28} & 900.00 & \textbf{2486.25}\\ \bottomrule

    \\[-7pt]
    & \multicolumn{12}{ c }{Combinatorial Auctions} \\
    \\[-3pt]
\end{tabular}
}

}
\end{table*}

{
\section{Exploration of Additive Augmentation Operators}
\label{appendix:additive_augmentation}

As discussed in the main text, our primary adversarial graph augmenter relies on a masking operator applied to MILP instances. To explore the full potential of our framework, we further introduce and evaluate an {additive augmentation operator}. Instead of merely removing components, this operator adds new elements into the bipartite graph representation $\mathcal{G} = (\mathcal{V}, \mathcal{C}, \mathcal{E})$ of a MILP instance. Specifically, we append new variable nodes to $\mathcal{V}$ and constraint nodes to $\mathcal{C}$. This is implemented via two dedicated augmentation networks: one for generating the features of the newly added nodes, and another for predicting their corresponding edges. To ensure that the augmented instances remain mathematically solvable and do not deviate excessively from the original data distribution, we strictly constrain the perturbation scale, setting the probability of adding both nodes and edges to $0.01$.

We compare our default masking-only strategy against the pure additive operator ($p=0.01$) and a hybrid strategy (Masking + Addition). The evaluations are conducted on the Combinatorial Auctions and Set Covering datasets, with results summarized in Table \ref{table:additive_operator}. The empirical results demonstrate that the additive augmentation operator yields comparable performance to the masking-only approach. Notably, the hybrid strategy (Masking + Addition) pushes the performance upper bound even further in Set Covering by exposing the model to a richer set of viable structural modifications during training. 

\begin{table*}[htb]
{ 
\caption{{Experiments on different augmentation operators. }}
\vspace{-1mm}
\label{table:additive_operator}
\centering
\resizebox{\textwidth}{!}{
\begin{tabular}{@{}ccccccccccccc@{}}
\toprule
\toprule
& \multicolumn{2}{c}{$D_1$ } & \multicolumn{2}{c}{$D_2$ } & \multicolumn{2}{c}{$D_3$ } & \multicolumn{2}{c}{$D_4$ }& \multicolumn{2}{c}{$D_5$}& \multicolumn{2}{c}{$D_6$}\\ \midrule
Method & Time(s) $\downarrow$ & PD integral $\downarrow$ & Time(s) $\downarrow$ & PD integral $\downarrow$ & Time(s) $\downarrow$ & PD integral $\downarrow$ & Time(s) $\downarrow$ & PD integral $\downarrow$ & Time(s) $\downarrow$ & PD integral $\downarrow$ & Time(s) $\downarrow$ & PD integral $\downarrow$\\\cmidrule(lr){1-1} \cmidrule(lr){2-3} \cmidrule(lr){4-5} \cmidrule(lr){6-7}\cmidrule(lr){8-9}\cmidrule(lr){10-11}\cmidrule(lr){12-13}
Masking  & {1.63}  & {22.44} & {16.34}  & {117.13} & {178.32}  & 314.32& 817.57 & {1154.79} & 900.00 & {1716.55} & 900.00 & {2107.19}\\
Addition & 1.62 & 22.10 & 17.03 & 117.45 & 188.16& 358.04 & 811.37 & 1072.18 & 900.00 & 1701.49 & 900.00 & 2174.20 \\ 
Masking + Addition  &1.67 & 22.57& 16.95 & 117.59 & 177.28 & 299.53 & 804.39 & 1048.52 & 900.00 & 1735.97 & 900.00 & 2094.57\\ 
\bottomrule

    \\[-7pt]
    & \multicolumn{12}{ c }{Combinatorial Auctions} \\
    \\[-3pt]
\end{tabular}
}
\newline
\vspace{1mm}
\newline
\resizebox{\textwidth}{!}{
\begin{tabular}{@{}ccccccccccccc@{}}
\toprule
\toprule
& \multicolumn{2}{c}{$D_1$ } & \multicolumn{2}{c}{$D_2$ } & \multicolumn{2}{c}{$D_3$ } & \multicolumn{2}{c}{$D_4$ }& \multicolumn{2}{c}{$D_5$}& \multicolumn{2}{c}{$D_6$}\\ \midrule
Method & Time(s) $\downarrow$ & PD integral $\downarrow$ & Time(s) $\downarrow$ & PD integral $\downarrow$ & Time(s) $\downarrow$ & PD integral $\downarrow$ & Time(s) $\downarrow$ & PD integral $\downarrow$ & Time(s) $\downarrow$ & PD integral $\downarrow$ & Time(s) $\downarrow$ & PD integral $\downarrow$\\\cmidrule(lr){1-1} \cmidrule(lr){2-3} \cmidrule(lr){4-5} \cmidrule(lr){6-7}\cmidrule(lr){8-9}\cmidrule(lr){10-11}\cmidrule(lr){12-13}
Masking  & {6.28} & 80.67 & {73.78} &515.92 & 839.80 & {7079.13}& 900.00 &{10691.05} &900.00&{14452.71}&900.00&{21792.69}\\
Addition & 6.94& 82.25& 75.21& 558.36& 886.03& 7236.61& 900.00 & 11892.37 & 900.00 & 14459.94 & 900.00 & 21938.27\\ 
Masking + Addition  & 6.13 & 80.14& 71.55& 501.58& 852.69 & 7050.72 & 900.00 & 10492.57 & 900.00 & 14405.53 & 900.00 & 21104.59 \\  \bottomrule

    \\[-7pt]
    & \multicolumn{12}{ c }{Set Covering} \\
    \\[-3pt]
\end{tabular}
\vspace{-2mm}
}
}
\end{table*}

}

\section{More Background Information}
\subsection{Branching Rules}
\label{section:branching_rules}
{
An efficiency branching policy significantly influences the size of B\&B trees, thus the overall solving efficiency.
Commonly used traditional branching rules include strong branching and pseudocost branching.

The strong branching rule is known to produce the smallest B\&B trees among the branching rules.
Let $P_0$ denote the linear relaxed optimal objective for the parent node.
We call a variable $x_i$ a branching candidate variable if $x_i$ has a fractional value $x_i^*$ in the optimal solution of the parent LP relaxation.
For a branching candidate variable $x_i$, we let $P_{i,1}$ and $P_{i,2}$ be the relaxed optimal objective for the children obtained by adding the lower or upper bound ($x_i\le \lfloor x_i^*\rfloor, x_i\ge \lceil x_i^*\rceil$), respectively.
Then strong branching chooses the branching variable $x_{SB}$ in the branching candidate set $\mathcal{C}$ by the following rules,
\begin{align*}
  x_{SB} = \underset{x_i\in\mathcal{C}}{\text{argmax}}\left[\max(P_{i,1}-P_0, \epsilon)\cdot\max(P_{i,2}-P_0, \epsilon)\right],
\end{align*}
where $\epsilon>0$ is a small constant.
Since strong branching needs to solve the LP relaxations of the two sub-problems for each candidate variable, it requires considerable computation.

The pseudocost branching rule is simpler, with much less computational cost than strong branching.
It keeps track of the previous successful branching variables and gives a score for each candidate branching variable using the history information.
The pseudocost branching rule is faster than the strong branching rule with less computational cost.
However, the pseudocost branching rule requires considerable human experience for manual tuning.

To overcome the disadvantages of strong branching and pseudocost branching, some hybrid branching rules are proposed to combine these methods.
The reliability pseudocost branching is one of the variants of these hybrid rules with state-of-the-art performance.
Thus, the reliability pseudocost branching rule is used as the default branching rule in SCIP.
}

\subsection{GNN Encoder for Branching Samples}
In recent studies \cite{gasse2019exact}, researchers use a single graph neural network composed of two interleaved half-convolutions to process the instance bipartite graphs or branching samples.
GNN first performs message passing from the variable nodes to constraint nodes, then from constraint nodes to variable nodes to obtain the variable embeddings, i.e.,
\begin{align*}
 \mathbf{h}_i^W= & \mathrm{MLP}_2\left([\mathbf{w}_i, \sum_{(\mathbf{w}_i,\mathbf{v}_j)\in \mathcal{E}}\mathrm{MLP}_1(\left[\mathbf{w}_i,\mathbf{v}_j,e_{ij}\right])]\right) \\
  \mathbf{h}_j^V= & \mathrm{MLP}_4\left([\mathbf{v}_j, \sum_{(\mathbf{w}_k,\mathbf{v}_j)\in \mathcal{E}}\mathrm{MLP}_3(\left[\mathbf{w}_k,\mathbf{v}_j,e_{kj}\right])]\right),
\end{align*}
where MLP denotes the multi-layer perceptron and $[\cdot]$ is the concatenation operation of vectors.
The variable embeddings $\mathbf{h}_j^V$ can be used for downstream tasks such as optimal solution prediction, branching variable selection and so on.

\begin{figure}[H]
  \centering
  \includegraphics[width=0.5\textwidth]{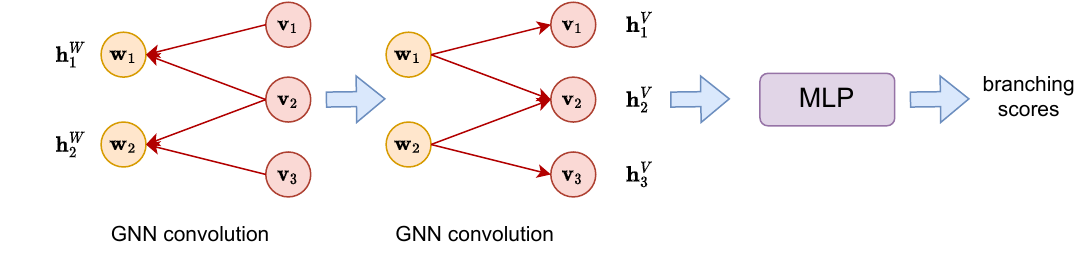}
  \caption{GNN inference for branching score prediction. GNN first performs message passing from the variable nodes to constraint nodes, then from constraint nodes to variable nodes to obtain the variable embeddings. Finally, we apply an MLP to obtain the branching scores.
  }\label{bipartite_conv}
    \vspace{-3mm}
\end{figure}

\subsection{Primal-Dual Integral Metric}
\label{appendix:PD_metric}
When employing the B\&B algorithm, we keep track of the global primal and dual bounds. The global primal bound at time $t$, i.e., $\mathbf{c}^\top \mathbf{x}_t^*$, is the best objective found at time $t$.
The global dual bound is the minimum dual bound $z^*_t$ of all the search tree leaves at time $t$.
By definition, we have the primal-dual gap $\mathbf{c}^\top \mathbf{x}_t^*-\mathbf{z}^*_t>0$, and this gap decreases as the B\&B algorithm proceeds.
B\&B finds the optimal solution if and only if $\mathbf{c}^\top \mathbf{x}_t^*-\mathbf{z}^*_t=0$.

{ The most straightforward way to evaluate a solver's performance within a given time limit is through the primal-dual gap. As noted by L2Dive \cite{diving}, "Unfortunately, measuring the primal-dual gap at time 
$T$ is susceptible to the particular choice of cutoff time." For a more comprehensive assessment of average performance across various cutoff times, the primal-dual integral is a better option.
The primal-dual integral is a critical metric widely used in existing works on ML4CO, including \cite{wanglearning, pmlr-v176-gasse22a, diving, khalil2017learning, NEURIPS2021_cb7c403a, huang2024distributional}. The NeurIPS competition ML4CO \cite{pmlr-v176-gasse22a} (\url{https://www.ecole.ai/2021/ml4co-competition/}) even established a separate task (the configuration task) aimed at improving the primal-dual integral.}
With time limit $T$, the primal-dual integral (PD integral) is defined as
\begin{align*}
    \int_0^T (\mathbf{c}^\top \mathbf{x}_t^*-\mathbf{z}_t^* )dt.
\end{align*}
When solving the same instance, solvers that achieve smaller PD-integral often lead to better feasible solutions given the same solving time.
{ In practical applications, the choice between the primal-dual integral and primal-dual gap depends on the specific scenario. In situations where we seek high-quality solutions as soon as possible, the primal-dual integral is more appropriate, as lower values indicate faster convergence. Conversely, when our focus is solely on the quality of the solution over a longer period, the primal-dual gap is a more suitable metric.}

\section{Theoretical Analysis}
\subsection{More Related Work on Theoretical Results for Data Augmentation}
Data augmentation is an effective method for introducing prior information and promoting data diversity, which is widely used in supervised learning \cite{Shorten2019ASO, Zhong2017RandomED, Perez2017TheEO} and reinforcement learning tasks \cite{yarats2021image, NEURIPS2020_e615c82a, pmlr-v119-laskin20a}.
Recently, augmentation on graph data has attracted more and more attention  \cite{sadeghi2021distributionally, dai2018adversarial, feng2019graph, xue2021cap}.
For theoretical results, previous work \cite{Chen20Grouptheoretic} proposes a generalization bound for data augmentation in a group-theoretical way; \cite{pmlr-v97-dao19b} analyses from the kernel perspective of data augmentation;  \cite{pmlr-v119-wu20g} studies the effects of data augmentation on the ridge estimator in an over-parametrized linear regression setting.

{
\subsection{Preliminaries}
We introduce some preliminaries for the proof, including the McDiarmid’s inequality and Wasserstein distance. 

\textbf{McDiarmid's inequality} is a concentration inequality that generalizes Hoeffding's inequality to functions of dependent random variables. It provides bounds on the deviation of a function from its expected value when the function satisfies a bounded differences condition.
Consider a learning scenario where $\mathcal{D} = \{G_1, \dots, G_n\}$ is an i.i.d. training set with $G_i \sim p$, $\Pi$ is a hypothesis class with $\pi \in \Pi$, and $\ell$ is a loss function bound by $[0,1]$.
We define
\begin{align*}
    \text{Empirical risk:} \quad & \hat{\mathcal{J}_p}(\pi) = \frac{1}{n}\sum_{i=1}^n \ell(\pi(G_i), y) \\
    \text{Expected risk:} \quad & \mathcal{J}_p(\pi) = \mathbb{E}_{G \sim p}[\ell(\pi(G), y)]
\end{align*}
We want to bound the generalization gap
\begin{align*}
\Delta(\pi) = |\hat{\mathcal{J}_p}(\pi) - \mathcal{J}_p(\pi)|
\end{align*}
The following theorem can be found at \cite{MohriRostamizadehTalwalkar18}.
\begin{theorem}
For any $\delta \in (0,1)$, with probability at least $1-\delta$ over the random draw of $\mathcal{D}$,
\begin{align*}
\sup_{\pi \in \Pi} \Delta(\pi) \leq \mathbb{E}_\mathcal{D}\left[\sup_{\pi \in \Pi} \Delta(\pi)\right] + \sqrt{\frac{\log(1/\delta)}{2n}}
\end{align*}
\end{theorem}
We can further use Rademacher complexity $\mathfrak{R}_n(\Pi)$,
\begin{align*}
\mathbb{E}_S\left[\sup_{\pi \in \Pi} \Delta(\pi)\right] \leq 2\mathfrak{R}_n(\Pi)
\end{align*}
Combining these results gives the final bound
\begin{align*}
\sup_{\pi \in \Pi} \Delta(\pi) \leq 2\mathfrak{R}_n(\Pi) + \sqrt{\frac{\log(1/\delta)}{2n}}
\end{align*}

\textbf{The Wasserstein distance} measures the minimum cost to transform one probability distribution into another. For two distributions $p$ and $q$ on a space $\mathcal{G}$, the Wasserstein distance is
\begin{align*}
    \mathcal{W}(p, q) = \left( \inf_{\mu} \mathbb{E}_{(G_1,G_2)\sim \gamma} c(G_1,G_2) \right)
\end{align*}
where $\mu$ is the set of joint distributions with marginals $p$ and $q$, $c(G_1,G_2)$ is the cost function on the space $\mathcal{G}$
The computationally intensive primal form can be reformulated via duality,
\begin{align*}
    \mathcal{W}(p, q) = \sup_{f \in \text{Lip}_1} \left( \mathbb{E}_{G_1 \sim p}[f(G_1)] - \mathbb{E}_{G_2\sim q}[f(G_2)] \right),
\end{align*}
where $\text{Lip}_1$ denotes 1-Lipschitz functions satisfying $|f(G_1)-f(G_2)| \leq c(G_1,G_2)$.

\subsection{Proof of Theorem \ref{thm_IL_bound}}
The generalization bound for Equation \ref{equ_il_bound1} and \ref{equ_il_bound2} can be derived from the classical approach, where we briefly show the proof as follows.
We start by splitting the term to be bounded into two parts,
\begin{align*}
  \mathcal{J}_{p_{\text{te}}}(\hat{\pi})-\hat{\mathcal{J}}_{p_\text{tr}}(\hat{\pi})
  =\left[\mathcal{J}_{p_{\text{te}}}(\hat{\pi})-\mathcal{J}_{p_\text{tr}}(\hat{\pi})\right]+\left[ \mathcal{J}_{p_\text{tr}}(\hat{\pi})-\hat{\mathcal{J}}_{p_\text{tr}}(\hat{\pi})\right].
\end{align*}
The second part can be bounded by the Rademacher complexity using McDiarmid's inequality,
\begin{align*}
  \left|\mathcal{J}_{p_\text{tr}}(\hat{\pi})-\hat{\mathcal{J}}_{p_\text{tr}}(\hat{\pi})\right| \le 2\Rad_{n}(\Pi)+\Omega_n.
\end{align*}
To bound the first part, we see that
\begin{align*}
  &\mathcal{J}_{p_{\text{te}}}(\hat{\pi})-\mathcal{J}_{p_\text{tr}}(\hat{\pi})\\
  =&\mathbb{E}_{G'_t\sim p_{\text{te}}(G'_{1:T}|\pi_E)}\left[\ell(\hat{\pi}(G'_t),\pi_E(G'_t))\right]\\
  &-\mathbb{E}_{G_t\sim p_{\text{tr}}(G_{1:T}|\pi_E)} \left[\ell(\hat{\pi}(G_t),\pi_E(G_t))\right]\\
  \le& \sup_{\|f\|_{\text{Lip}}\le1}\left(\mathbb{E}_{G'_t\sim p_{\text{te}}(G'_{1:T}|\pi_E)}[f(G'_t)]-\mathbb{E}_{G_t\sim p_{\text{tr}}(G_{1:T}|\pi_E)}[f(G_t)]\right)\\
  =& \mathcal{W}_{\mathcal{G}}(p_{\text{tr}}(G_{1:T}|\pi_E), p_{\text{te}}(G_{1:T}|\pi_E)),
\end{align*}
where $\|f\|_{\text{Lip}}\le1$ is the functions with Lipschitz constant less then 1.
The inequality holds as $f(\cdot)=\ell(\hat{\pi}(\cdot),\pi_E(\cdot))$ is bounded and Lipschitz.
The last equation is by the definition of Wasserstein distance. 
We complete the proof of Equation \ref{equ_il_bound1} by combining the bounds for these two parts.
The proof of the generalization bound under instance augmentation (Equation \ref{equ_il_bound2}) is the same.
We next compare the generalization bounds of the two bounds.
\begin{equation*}
\begin{aligned}
  &2\widetilde{\Rad}_n(\Pi)-2\Rad_{n}(\Pi)\\
  \le&2\mathbb{E}\mathbb{E}_{\sigma}\left[\sup_{\pi\in\Pi}\frac{1}{nT}\left|\sum_{i=1}^n\sigma_{i}\sum_{t=0}^{T-1}\ell(\pi(\tilde{G}^{(i)}_t),\pi_{E}({\tilde{G}^{(i)}_t}))\right|\right]\\
  & \qquad-2\mathbb{E}\mathbb{E}_{\sigma}\left[\sup_{\pi\in\Pi}\frac{1}{nT}\left|\sum_{i=1}^n\sigma_{i}\sum_{t=0}^{T-1}\ell(\pi(G^{(i)}_t),\pi_{E}({G^{(i)}_t})\right|\right]\\
  \le& 2\mathbb{E}\mathbb{E}_{\sigma}\left[\sup_{\pi\in\Pi}\frac{1}{nT}\left|\sum_{i=1}^n\sigma_{i}\sum_{t=0}^{T-1}\ell(\pi(\tilde{G}^{(i)}_t),\pi_{E}({\tilde{G}^{(i)}_t}))\right.\right.\\
  & \qquad-\left.\left.\sum_{i=1}^n\sigma_{i}\sum_{t=0}^{T-1}\ell(\pi(G^{(i)}_t),\pi_{E}({G^{(i)}_t})\right|\right]\\
  \le &2\mathbb{E}\left[\inf_\mu c(G^{(i)}_t,\tilde{G}^{(i)}_t)\right]\\
  \le &2\inf_\mu\mathbb{E}\left[ c(G^{(i)}_t,\tilde{G}^{(i)}_t)\right]\\
  =& 2\mathcal{W}_{\mathcal{G}}(\tilde{p}(G_{1:T}|\pi_E), p_{\text{te}}(G_{1:T}|\pi_E)).
\end{aligned}
\end{equation*}
If $\alpha\le1/3$ and the following inequality holds,
\begin{equation*}
\begin{aligned}
 & 3\mathcal{W}_{\mathcal{G}}(\tilde{p}(G_{1:T}|\pi_E), p_{\text{te}}(G_{1:T}|\pi_E))\\
  \le&\mathcal{W}_{\mathcal{G}}(p_{\text{tr}}(G_{1:T}|\pi_E), p_{\text{te}}(G_{1:T}|\pi_E))
\end{aligned}
\end{equation*}
We thus show Equation \ref{equ_compare_bound1}, then the generalization bound of instance augmentation can be lower.
  \begin{align*}
    &\mathcal{W}_{\mathcal{G}}(\tilde{p}(G_{1:T}|\pi_E), p_{\text{te}}(G_{1:T}|\pi_E))
       +2\widetilde{\Rad}_n(\Pi)\\
       &\le \mathcal{W}_{\mathcal{G}}(p_{\text{tr}}(G_{1:T}|\pi_E), p_{\text{te}}(G_{1:T}|\pi_E))+2\Rad_{n}(\Pi).
  \end{align*}

\subsection{Proof of Theorem \ref{thm_rl_error}}
We start by decomposing the generalization error as we did in the IL setting
\begin{equation}\label{equ_rl_decomposition}
\begin{aligned}
  &\mathcal{J}_{p_{\text{te}}}(\tilde{\pi})-\mathcal{J}_{p_{\text{te}}}(\pi^*)\\
   =&  \left(\mathcal{J}_{p_{\text{te}}}(\tilde{\pi})-\mathcal{J}_{\tilde{p}}(\tilde{\pi})\right)+\left(\mathcal{J}_{\tilde{p}}(\tilde{\pi}\right)-\hat{\mathcal{J}}_{\tilde{p}}(\tilde{\pi}))\\
  &+(\hat{\mathcal{J}}_{\tilde{p}}(\tilde{\pi})-\hat{\mathcal{J}}_{\tilde{p}}(\pi^*))\\
  &+(\hat{\mathcal{J}}_{\tilde{p}}(\pi^*)-\mathcal{J}_{\tilde{p}}(\pi^*))
  +(\mathcal{J}_{\tilde{p}}(\pi^*)-\mathcal{J}_{p_{\text{te}}}(\pi^*))\\
  \le & (\mathcal{J}_{p_{\text{te}}}(\tilde{\pi})-\mathcal{J}_{\tilde{p}}(\tilde{\pi}))+(\mathcal{J}_{\tilde{p}}(\tilde{\pi})-\hat{\mathcal{J}}_{\tilde{p}}(\tilde{\pi}))\\
  &+(\hat{\mathcal{J}}_{\tilde{p}}(\pi^*)-\mathcal{J}_{\tilde{p}}(\pi^*))
  +(\mathcal{J}_{\tilde{p}}(\pi^*)-\mathcal{J}_{p_{\text{te}}}(\pi^*))\\
\end{aligned}
\end{equation}
The inequality in (\ref{equ_rl_decomposition}) holds since the training loss $\hat{\mathcal{J}}_{\tilde{p}}$ reaches its minimum at the policy $\tilde{\pi}$ by definition, and thus we have $\hat{\mathcal{J}}_{\tilde{p}}(\tilde{\pi})-\hat{\mathcal{J}}_{\tilde{p}}(\pi^*)\le 0$.
Then, we apply McDiarmid's inequality to the term $\mathcal{J}_{\tilde{p}}(\tilde{\pi})-\hat{\mathcal{J}}_{\tilde{p}}(\tilde{\pi})$ and $\hat{\mathcal{J}}_{\tilde{p}}(\pi^*)-\mathcal{J}_{\tilde{p}}(\pi^*)$, we have 
\begin{equation}
\label{equ_rl_2}
    \begin{aligned}
        \left|\mathcal{J}_{\tilde{p}}(\tilde{\pi})-\hat{\mathcal{J}}_{\tilde{p}}(\tilde{\pi})\right|\le 2\Rad_n(\Pi)+\Omega_n\\
        \left|\hat{\mathcal{J}}_{\tilde{p}}(\pi^*)-\mathcal{J}_{\tilde{p}}(\pi^*)\right|\le 2\Rad_n(\Pi)+\Omega_n,
    \end{aligned}
\end{equation}
with confidence at least $1-2\delta$.
Finally, we bound the first and the last terms.
Different from the IL setting, the transition function plays an important role in sampling training branching samples.
Thus, the branching samples cannot be viewed as i.i.d samples.

For any policy $\pi\in\Pi$ (including $\tilde{\pi}$ and $\pi^*$), we have the following estimations,
\begin{align*}
  &\mathcal{J}_{\tilde{p}}(\pi)-\mathcal{J}_{p_{\text{te}}}(\pi)\\
  =& \mathbb{E}_{\tilde{G}_t\sim \tilde{p}(\tilde{G}_{1:T}|\pi_\theta)} \left[\sum_{t=0}^{T-1}-\gamma^t R(\tilde{G}_t)\right]\\
  &-\mathbb{E}_{G_t\sim p_{\text{te}}(G_{1:T}|\pi_\theta)} \left[\sum_{t=0}^{T-1}-\gamma^t R(G_t)\right]\\
  \le & \mathbb{E}_{\tilde{G}_t, G_t}\left[\sum_{t=0}^{T-1}\gamma^t\left|R(\tilde{G}_t)-R({G}_t)\right|\right]\\
  \le & \sum_{t=0}^{T-1}\gamma^tL_3\mathbb{E}_{\tilde{G}_t, G_t}\left[c(\tilde{G}_t,G_t)\right].
\end{align*}
Next, we bound the discrepancy between states from the original instances and the augmented instances at each step recursively.
\begin{align*}
  & c(\tilde{G}_t,G_t) \\
  \le& c(\mathcal{T}(\tilde{G}_{t-1},\pi(\tilde{G}_{t-1}),\mathcal{T}(G_{t-1},\pi(G_{t-1})))\\
  \le & L_1 c(\tilde{G}_{t-1},G_{t-1})\\
  \le&L_1^tc(\tilde{G}, G).
\end{align*}
We thus obtain the bound
\begin{equation}\label{equ_rl_3}
\begin{aligned}
  \mathcal{J}_{\tilde{p}}(\pi)-\mathcal{J}_{p_{\text{te}}}(\pi)\le \sum_{t=0}^{T-1}\gamma^tL_3L_1^t\mathcal{W}_{\mathcal{G}}(\tilde{p}, p_{\text{te}}).
\end{aligned}
\end{equation}
So far, combining Equation \ref{equ_rl_2} and \ref{equ_rl_3}, we complete the proof.

\begin{align*}
    \mathcal{J}_{p_{\text{te}}}(\tilde{\pi})-\mathcal{J}_{p_{\text{te}}}&(\pi^*)\le4\Rad_n(\Pi)+2\Omega_n\\
    &+2\sum_{t=0}^{T-1}\gamma^tL_3L_1^t\mathcal{W}_{\mathcal{G}}(\tilde{p}, p_{\text{te}}).
\end{align*}
}

\section{Details of the Benchmarks}
\label{appendix:benchmarks}
\subsection{Synthesis Benchmarks for Evaluations}
We evaluate the performance of our proposed methods on four classical benchmarks of NP-hard problem families, namely combinatorial auctions, minimum set covering, capacitated facility location and maximum independent set.
We use these four synthesis benchmarks mainly for the following reasons.
First, these four benchmarks are widely-used testing beds for MILP solvers \cite{gasse2019exact,gupta2020hybrid,gupta2022lookback}.
Second, we can generate a series of transfer distributions with much larger sizes to evaluate solvers' generalization ability.
Third, these benchmarks represent a wide collection of MILP problems in practice.

We use the code in \url{https://github.com/ds4dm/learn2branch} for data generation.
(1) \textbf{Combinatorial auctions} is comprised of instances generated following the arbitrary relationships procedure of \cite{leyton2000towards}.
(2) \textbf{Set covering} is comprised of instances generated following \cite{balas1980set} with default density 0.05.
(3) \textbf{Capacitated facility location} is comprised of instances generated following \cite{Cornujols1991ACO}.
(4) \textbf{Maximum independent set} is comprised of instances on Erd\H{o}s-R\'enyi random graphs, generated following the procedure of \cite{Bergman2015DecisionDF}.
For all benchmarks, we classify six distributions $D_1$ to $D_6$ by increasing instance sizes or difficulty levels, where $D_1$ is the easiest with the smallest sizes and $D_6$ is the most difficult with the largest sizes that are \textbf{five to ten times larger} than those in $D_1$.
The baselines can solve all the instances of $D_1$ and $D_2$ within 900s but reach the time limit in $D_4-D_6$.
Table \ref{table:data_generation} summarizes the generation hyperparameters for generation algorithms.

We generate 10,000 $D_1$ instances for training and 2,000 $D_1$ instances for validation, and we evaluate on $32$ instances on $D_1$ to $D_6$, respectively.
For the GNN and AdaSolver-IL methods, we solve the training instances with SCIP 7.0.3 and record branching samples and branching decisions of a strong branching expert.
We collect 100,000 samples for training and 20,000 for validation in total.
For RL-based solvers, we limit the processing time of each episode to no more than 1000s.

\begin{table}[h!]
\caption{The hyperparameters for generating the instances.}
\centering
\label{table:data_generation}
\small{
\vspace{-1mm}
\resizebox{0.4\textwidth}{!}{
\begin{tabular}{@{}cccccccc@{}}
\toprule
\toprule
Distributions & $D_1$ & $D_2$ & $D_3$ & $D_4$ & $D_5$ & $D_6$ \\ \midrule
Row & 500 & 1000 & 2000 & 3000 & 4000 & 8000\\
Column & 1000 & 1000 & 1000 & 1000 & 1000 & 1000\\
\bottomrule
\multicolumn{7}{ c }{Set Covering} \\
    \\[-3pt]
\end{tabular}
}
\resizebox{0.4\textwidth}{!}{
\begin{tabular}{@{}cccccccc@{}}
\toprule
\toprule
Distributions & $D_1$ & $D_2$ & $D_3$ & $D_4$ & $D_5$ & $D_6$ \\ \midrule
Items & 100 & 200 & 300 & 400 & 500 & 600\\
Bids & 500 & 1000 & 1500 & 2000 & 2500 & 3000\\
\bottomrule
\multicolumn{7}{ c }{Combinatorial Auctions}  \\
    \\[-3pt]
\end{tabular}
}
\resizebox{0.4\textwidth}{!}{\begin{tabular}{@{}cccccccc@{}}
\toprule
\toprule
Distributions & $D_1$ & $D_2$ & $D_3$ & $D_4$ & $D_5$ & $D_6$ \\ \midrule
Facilities & 100 & 200 & 400 & 600 & 800 & 1600\\
Customers & 100 & 100 & 100 & 100 & 100 & 100\\
\bottomrule
\multicolumn{7}{ c }{Capacitated Facility Location} \\
    \\[-3pt]
\end{tabular}
}
\resizebox{0.4\textwidth}{!}{\begin{tabular}{@{}cccccccc@{}}
\toprule
\toprule
Distributions & $D_1$ & $D_2$ & $D_3$ & $D_4$ & $D_5$ & $D_6$ \\ \midrule
Nodes & 500 & 1000 & 1500 & 2000 & 2500 & 3000\\
Affinity & 4 & 4 & 4 & 4 & 4 & 4\\
\bottomrule
\multicolumn{7}{ c }{Maximum Independent Set} \\
    \\[-3pt]
\end{tabular}
}
}
\end{table}

\subsection{Real-world Benchmarks for Evaluations}
The real-world dataset we use in this paper is the Anonymous dataset from the NeurIPS 2021 ML4CO competition \cite{pmlr-v176-gasse22a}.
It consists of 98 training and 20 validation instances.
Similarly to what we do in the synthesis benchmarks, we collect $100,000$ training samples and $20,000$ validation samples for GNN and AdaSolver-IL.
For RL-based solvers, we limit the processing time of each episode to no more than 1,000s.

\subsection{Datasets for the Visualization}
\label{data:vis}
We generate ten distributions of the set covering benchmark for visualization using the combinations of $\{500,1000\}$ rows, 1000 columns and $\{0.01, 0.05, 0.1, 0.2, 0.5\}$ density.
The density hyperparameter for data generation controls the proportion of nonzero entries of the constraint matrices.
We change the density of the instances to obtain graphs with different local structures to simulate the environmental perturbations in real-world applications.

We generate ten distributions of the combinatorial auctions benchmark for visualization using the combinations of $\{100,150,200,250,300\}$ items and $\{500, 600\}$ bids to observe the embeddings for the GNN branching policy.

\section{Algorithm}
{For simplicity, we unify the notation $\mathcal{D}$ to represent the branching sample dataset used for IL- and RL-based methods.}

\begin{algorithm2e}[h]
	\caption{Training of AdaSolver.}
	\label{alg-training}
{
\textbf{INITIALIZE:} learning-based branching policy $\pi_{\theta}$, adversarial graph augmentation policy $(\phi_\eta, D_{\psi})$, training instances $\{G^{(i)}\}_{i=1}^n$\\
 \textcolor{gray}{\# For IL}\\
 Use $\{G^{(i)}\}_{i=1}^n$ to collect branching sample dataset ${\mathcal{D}}$ \\
	\For{$N$ \text{epochs}}{
         \textcolor{gray}{\# Data augmentation} \\
        Perform augmentation on $\{G^{(i)}\}_{i=1}^n$ to obtain the augmented instance $\{\phi(G^{(i)})\}_{i=1}^n$ with policy $\phi_\eta$\\
         \textcolor{gray}{\# For IL}\\
        Use $\{\phi(G^{(i)})\}_{i=1}^n$ to collect branching sample dataset $\tilde{\mathcal{D}}$\\
         \textcolor{gray}{\# For RL}\\
        Use $\{\phi(G^{(i)})\}_{i=1}^n$ as the augmented dataset $\tilde{\mathcal{D}}$
        \\
        \textcolor{gray}{\# Train the GNN branching policy} \\
        Compute the training objective of GNN branching policy $\hat{\mathcal{J}}_{\tilde{p}}(\pi_\theta)$ (IL or RL) over over $\mathcal{D}\cup \tilde{\mathcal{D}}$\\
        Update the parameters $\theta$\\
        \textcolor{gray}{\# Train the data augmentation network} \\
        Record $\{(G^{(i)},\tilde{r}(\tilde{a}_i|G^{(i)}), \tilde{a}^{(i)}),p_{\eta}(\tilde{a}^{(i)}|G^{(i)}))\}_{i=1}^n$ \\
        Compute the losses $\mathcal{L}(\eta)$ and $\mathcal{L}(\theta_3)$\\
        Update the parameters $\eta$ and $\theta_3$\\
	}
	\textbf{OUTPUT:} parameters of branching policy $\theta^*$
 }
\end{algorithm2e}

\section{Implementation Details}
\label{appendix:implementation}
\subsection{Hard-ware Specification and Experiment Settings}
We conducted all the experiments on a single machine with NVidia GeForce GTX 3090 GPUs and Intel(R) Xeon(R) E5-2667 v4 CPUs @ 3.20GHz.
During the evaluation, we run three random seeds $\{0,1,2\}$ over each instance.

\subsection{Details of the Baselines}
In this part, we introduce the details of the baselines used in this paper.

\noindent\textbf{(1) GNN.} We use the same implementation of GCNN in \cite{gasse2019exact}.
First, we apply two interleaved half-convolutions to the input bipartite graphs.
The graph convolution has two successive passes with 64 units, one from variable to constraints and one from constraints to variables.
Then we apply a final 2-layer perceptron on the embeddings of variable nodes, combined with a masked softmax activation to produce a probability distribution over the branching candidates.
We also use the prenorm layers proposed in \cite{gasse2019exact} and conduct a pre-train stage for them.

\noindent\textbf{(2) SVMRank and LMART.} We implement the SVMRank and LMART model for branching policy in SCIP as \cite{gasse2019exact}.
The two models take inputs as the 91-dimensional branching features described in \cite{khalil2016learning} and are trained on pairs of variable's state-rank samples for ranking.
The SVMRank is trained by minimizing a pairwise loss of variables and the LMART is trained by minimizing normalized cumulative gain.
As mentioned in \cite{gasse2019exact}, we reduce the dataset with 250,000 candidate variables for training and 100,000 for validation because of the prohibitive memory and time cost.

\noindent\textbf{(3) Trees.} The Trees model is implemented following that in \cite{gasse2019exact}, where we obtain the input variable features by concatenating the bipartite graph variable features, the minimum, mean and maximum of the edge-constraint features over its neighborhood.
The model predicts a branching score for each candidate variable and is trained by minimizing the mean-squared error between the predicted scores and ground truth strong branching scores.
We also reduce the size of the dataset for Trees.

\noindent\textbf{(4) tMDP+DFS.} tMDP+DFS is a reinforcement-learning-based method under the framework of tree Markov Decision Processes (tMDP), a variant of temporal Markov Decision Processes.
We implement the tMDP+DFS model with two interleaved half-convolutions GNN encoder as described in \cite{NEURIPS2022_756d74cd}.
The GNN model architecture is the same as \cite{gasse2019exact}.
tMDP+DFS is trained by REINFORCE with tree policy gradients and depth-first search node selection.

\noindent\textbf{(5) TreeGate.} TreeGate is an imitation learning framework with input features of 25-dimensional branching candidate states and 60-dimensional search tree states \cite{Zarpellon_Jo_Lodi_Bengio_2021}.
TreeGate introduces such tree-state parameterization by first embedding the tree-state inputs and then sending them to the proposed gating layers to extract the tree-based signal.
Combining the variable representations and tree-state representation, TreeGate predicts the branching scores for each branching candidate variable.

\subsection{Representations of instance Bipartite Graph and Branching Samples}
To avoid mixing up, recall that the instance bipartite graphs are the input of the augmentation policy and do not contain solving statistics (since we augment the instance before we solve it with the solver), while the branching samples are the input of branching policy with solving statistics.
We use the \emph{MILP Bipartite} representation for instance graphs implemented in Ecole 0.8 package \cite{prouvost2020ecole} with nine variable features, one constraint feature and one edge feature listed in Table \ref{table:aug_input_features}.
The branching samples use the graph representations described in \cite{gasse2019exact} listed in Table \ref{table:mlp_input_features}.

\begin{table}[H]
\caption{The variable features, constraint features and edge features used for instance bipartite graphs.}
\centering
\tiny{
\begin{tabular}{ccp{4.5cm}}
\toprule\toprule
Index & Variable Feature Name &  Description \\
\midrule
0 & Objective & Objective coefficient\\
 1-4     & Variable type    & Variable type (binary, integer, implicit-integer, continuous)\\
5    & Specified bounds    & Whether the variable has a lower bound\\
6    & Specified bounds     & Whether the variable has an upper bound\\
7    & Lower bound    & Whether the variable reaches its lower bound\\
8    & Upper bound     & Whether the variable reaches its upper bound\\
\bottomrule
\end{tabular}
}
\vspace{1mm}
\newline
\tiny{
\begin{tabular}{ccp{4.5cm}}
\toprule\toprule
Index & Constraint Feature Name &  Description \\
\midrule
0   & Bias    &Normalized right-hand-side of the constraint\\
\bottomrule
\end{tabular}
}
\vspace{1mm}
\newline
\tiny{
\begin{tabular}{ccp{4.5cm}}
\toprule\toprule
Index & Constraint Feature Name &  Description \\
\midrule
0   & Coefficient   &Constraint coefficient\\
\bottomrule
\end{tabular}
}
\label{table:aug_input_features}
\end{table}

\begin{table}[H]
\caption{The variable features, constraint features and edge feature used for branching samples, following those proposed in \cite{gasse2019exact}.}
\centering
\tiny{
\begin{tabular}{ccp{4.5cm}}
\toprule\toprule
Index & Variable Feature Name &  Description \\
\midrule
 0-3     & Variable type    & Variable type\\
4   & Normalized coefficient    &Normalized objective coefficient\\
5    & Specified bounds    & Whether the variable has a lower bound\\
6    & Specified bounds     & Whether the variable has an upper bound\\
7    & Lower bound    & Whether the variable reaches its lower bound\\
8    & Upper bound     & Whether the variable reaches its upper bound\\
 9     & Solution fractionality    & Fractional part of the variable\\
  10-13      &  Categorical   & Variable is at its bound or zero \\
 14     &  Reduced cost  & Amount by which the  objective coefficient of the variable should decrease so that the variable assumes a positive value in the LP solution\\
  15     & Age   & Number of LP iterations\\
 16     & Solution value   & Value of the variable\\
 17    & Incumbent value   & Value of the variable in the current best solution \\
 18    & Average incumbent value    & Average value in the observed feasible primal solutions\\
\bottomrule
\end{tabular}
}
\vspace{1mm}
\newline
\tiny{
\begin{tabular}{ccp{4.5cm}}
\toprule\toprule
Index & Constraint Feature Name &  Description \\
\midrule
 0     & Cosine similarity    & Cosine of the angle between objective coefficients and the coefficients of this constraint\\
1   & Bias    &Normalized right-hand -side of the constraint\\
2    & Age    & Iterations since the last time the constraint
was active\\
3    & Normalized dual value     & Value of dual variable corresponding to the constraint\\
4   & Bounds   & Whether the constraint reaches its bound\\
\bottomrule
\end{tabular}
}
\vspace{1mm}
\newline
\tiny{
\begin{tabular}{ccp{4.5cm}}
\toprule\toprule
Index & Constraint Feature Name &  Description \\
\midrule
0   & Coefficient   &Constraint coefficient\\
\bottomrule
\end{tabular}
}
\label{table:mlp_input_features}
\end{table}

\subsection{Implementation Details of AdaSolver}
\subsubsection{Model Architecture of the augmentation policy}
The augmentation policy consists of three parts: an augmentation policy network, a state-value function network and a discriminator network.
Each part has a GNN graph encoder and MLP decoders for downstream tasks.
The GNN graph encoder encodes the node features of the instance bipartite graphs into 10-dimensional embeddings, then applies an interleaved half-convolution layer to extract 64-dimensional representations for variables $\{\mathbf{h}_i^V\}_{i=1}^n$ and constraints $\{\mathbf{h}_j^W\}_{j=1}^m$, respectively.
The augmentation policy $\phi_\eta$ and state-value function $V_\alpha$ share the same instance bipartite graph encoder.
For the augmentation policy, we use three single-layer networks, $\text{MLP}_1(\mathbf{h}_i^V)$, $\text{MLP}_2(\mathbf{h}_j^W)$ and $\text{MLP}_3([\mathbf{h}_i^V, \mathbf{h}_j^W])$ to obtain the masking probability for the $i^{th}$ variable, the $j^{th}$ constraint and edge $\mathbf{e}_{i,j}$.
Then we choose the nodes and edges to mask according to the predicted probability with a predefined masking proportion.
The state-value function uses one single-layer network to predict the current state value.
The discriminator uses one single-layer network to predict the probability of an augmented instance that is from the original distribution.

\subsubsection{Training and Optimization}
we use Adam optimizer \cite{kingma2014adam} to train the networks in all experiments.

In the IL setting, we set an initial learning rate of $1\times 10^{-3}$ and batch size of 8 for GNN and set a learning rate of $1\times 10^{-3}$ and batch size of 4 for the augmentation policy.
We divide the learning rate for GNN by 5 when the GNN validation loss does not improve for 10 epochs, and stop training if it does not improve for 20.
The instance graph augmentation begins at Epoch 10.
In each epoch after Epoch 10, we choose $k_1$ instances to perform augmentation and extract $k_2$ branching samples from the augmented instances.
Here $k_1=50$, $k_2=1000$ in the combinatorial auctions benchmark; $k_1=100$, $k_2=5000$ in other three synthesis benchmarks; $k_1=10$, $k_2=1000$ in the real-world Anonymous dataset.
Then we randomly choose 10000 branching samples from the original instances and 2000 samples from the augmented instances to train the GNN.

In the RL setting, we set an initial learning rate of $1\times 10^{-6}$ and batch size of 8 for tMDP+DFS and set a learning rate of $1\times 10^{-6}$ and batch size of 4 for the augmentation policy.
In each epoch, we choose 10 instances and run the branching agent on them.
We augment 10 instances and run on them every ten epochs.
The maximum exploring time is set to be no more than 1000s on each instance.

\subsubsection{Hyperparameters}
We list the common hyperparameters in Table \ref{table:hyperparamter}.
For augmentation operators and masking proportions, we choose to mask 3\% constraints and 1\% edges on the set covering benchmark, mask 1\% variables on the combinatorial auctions benchmark, mask 1\% constraints and 1\% variables on the capacity facility location benchmark, mask 1\% edges and 5\% constraints on the maximum independent set, and mask 2\% constraints and 1\% variables in the Anonymous dataset.

\begin{table}[t]
\caption{Common hyperparameters for the augmentation policy in AdaSolver.}
\label{table:hyperparamter}
\vspace{-1mm}
\centering
\begin{tabular}{@{}lc@{}}
\toprule
\toprule
Hyperparameter& Value\\
\cmidrule(r){1-2}
Half-convolution layers & 1 for variables and 1 for constraints\\
Graph encoder hidden size & 64\\
Batch size & 4\\
PPO learning rate & 1e-3\\
PPO entropy coefficient  & 1e-2\\
PPO epochs at each iteration & 3\\
PPO buffer size    & 300\\
PPO clip ratio  & 0.2\\
\bottomrule
\\[-7pt]
\end{tabular}
\end{table}

\subsubsection{Variants of AdaSolver }
\noindent\textbf{Random augmentation (RM).}
The branching policy network is the same as GNN.
Different from AdaSolver, RM does not have an augmentation network.
Instead, RM randomly masks the constraints, variables and edges under the same operators and proportions as AdaSolver.


We will release all the codes for training and end-to-end evaluation once the paper is accepted.

\section{More Experimental Results}
In this part, we provide the results for the experiments in the main text in terms of nodes (Nodes) and the primal-dual gap (PD gap).
We first give some explanations for these metrics.

\textbf{Nodes.} The number of explored nodes is a metric that reflects the size of B\&B searching trees.
However, the Nodes metric may be not an appropriate metric to measure the solving efficiency.
First, fewer nodes do not mean shorter solving time since the solver may spend more time on each node, resulting in a longer total solving time.
Second, on the distributions that solvers cannot find the optimal solutions within the time limit, the Nodes metric is not appropriate to reflect the solving performance.
On the one hand, we still cannot estimate the number of explored nodes when solving to the optimal value.
On the other hand, given the time limit, fewer nodes imply that the solvers even explore the harder nodes that require more computation to solve the LP relaxations.

\textbf{PD gap.} (1) On the distributions that solvers can find the optimal solutions with the time limit, the PD gap metric reaches zero and fewer nodes imply a better solving performance.
(2) On the distributions that solvers cannot find the optimal solutions within the time limit, the PD gap is a suitable metric to measure the solving efficiency.
Given the time limit, the lower PD gap implies that the solver has found a better feasible solution so far.

\subsection{Analysis of the Imitation Accuracy}
We report the imitation learning accuracy on different distributions on the set covering and combinatorial auctions benchmarks in Table \ref{table:acc}.
For in-distribution $D_1$ imitation learning accuracy, AdaSolver-IL achieves comparable or higher imitation accuracy compared to the GNN baseline.
An interesting observation is that AdaSolver-IL consistently outperforms the GNN baseline in terms of solving efficiency in harder distributions $D_4-D_6$ of the set covering benchmark, despite having slightly lower imitation learning accuracy. 
This observation suggests that imitation learning accuracy alone is not a definitive factor for solving performance, and higher imitation learning accuracy does not necessarily imply better overall performance.

\begin{table*}[t]
\caption{Imitation learning accuracy on the testing distributions of the set covering and combinatorial auctions benchmarks. The imitation learning accuracy of AdaSolver-IL and GNN are comparable.}
\label{table:acc}
\vspace{-1mm}
\centering
\begin{tabular}{@{}ccccccc@{}}
\toprule
\toprule
&\multicolumn{6}{c}{Set Covering}\\
\cmidrule(r){1-7}
 & \multicolumn{1}{c}{$D_1$ } & \multicolumn{1}{c}{$D_2$  } & \multicolumn{1}{c}{$D_3$  } & \multicolumn{1}{c}{$D_4$  }& \multicolumn{1}{c}{$D_5$ }&\multicolumn{1}{c}{$D_5$ }\\
\cmidrule(r){1-7}
GNN &70.02& 66.36&54.07 & 48.03& 48.05 & 16.08 \\
AdaSolver-IL&70.03 &66.39 &54.02 &52.01 &44.08& 16.00\\
\bottomrule
\\[-7pt]
\end{tabular}
\hspace{1mm}
\centering
\begin{tabular}{@{}ccccccc@{}}
\toprule
\toprule
&\multicolumn{6}{c}{Combinatorial Auctions}\\
\cmidrule(r){1-7}
 & \multicolumn{1}{c}{$D_1$ } & \multicolumn{1}{c}{$D_2$  } & \multicolumn{1}{c}{$D_3$  } & \multicolumn{1}{c}{$D_4$  }& \multicolumn{1}{c}{$D_5$ }& \multicolumn{1}{c}{$D_6$ }\\
\cmidrule(r){1-7}
GNN &67.63 &72.42 &67.86 &72.07 & 76.02& 47.03  \\
AdaSolver-IL& 68.55 &72.57 &66.92 &62.00 &74.02 &44.06 \\
\bottomrule
\\[-7pt]
\end{tabular}
\end{table*}

\subsection{More Results of Main Evaluation}
\label{appendix:more_results_main_evalutation}

We present more results on the comparative experiments.
Table \ref{table:other_metrics} reports the performance of AdaSolver and the baselines in terms of the total number of nodes and PD gap.
The results in Table \ref{table:other_metrics} demonstrate that AdaSolver-IL explores significantly fewer nodes than tMDP and TreeGate consistently on the easier distributions, and slightly more nodes than GNN.
However, AdaSolver-IL achieves faster solving efficiency, which means that AdaSolver-IL spends less time than GNN on each node.
For the distributions that the solvers reach the time limit, AdaSolver-IL outperforms all the baselines in terms of the PD gap.
We also observe that AdaSolver-IL explores more nodes than other baselines within the limited time, which demonstrates that AdaSolver-IL learns a policy that is more efficient for exploring nodes.

\begin{table*}[h]
\caption{More results on main comparative experiments in terms of Nodes and PD gap. AdaSolver-IL explores significantly fewer nodes than tMDP and TreeGate consistently on the easier distributions, and slightly more nodes than GNN. However, AdaSolver-IL achieves a shorter solving time and lower PD integral than GNN, indicating that the Nodes metric may not directly reflect the solving efficiency. For the distributions that the solvers reach the time limit, AdaSolver-IL outperforms all the baselines in terms of the PD gap.}
\small
\vspace{-1mm}
\label{table:other_metrics}
\centering
\begin{center}
\resizebox{\textwidth}{!}{
\begin{tabular}{@{}ccccccccccccc@{}}
\toprule
\toprule
& \multicolumn{2}{c}{$D_1$ } & \multicolumn{2}{c}{$D_2$ } & \multicolumn{2}{c}{$D_3$ } & \multicolumn{2}{c}{$D_4$ }& \multicolumn{2}{c}{$D_5$}& \multicolumn{2}{c}{$D_6$}\\ \midrule
Method &  Nodes & PD gap $\downarrow$ &  Nodes & PD gap $\downarrow$&  Nodes& PD gap $\downarrow$&  Nodes & PD gap $\downarrow$ &  Nodes & PD gap $\downarrow$&  Nodes& PD gap $\downarrow$ \\\cmidrule(r){1-3} \cmidrule(r){4-5} \cmidrule(lr){6-7}\cmidrule(lr){8-9}\cmidrule(lr){10-11}\cmidrule(lr){12-13}
RPB &247.59 &0.00&5866.93  &0.00 &52814.36 &0.05 &25630.00 & 0.10 &18279.17  &0.15  &3910.22 & 0.26 \\
\cmidrule(r){1-3} \cmidrule(r){4-5} \cmidrule(lr){6-7}\cmidrule(lr){8-9}\cmidrule(lr){10-11}\cmidrule(lr){12-13}
SVMRank &332.60 &0.00&5478.59  &0.00 &20326.22 &0.03  &5152.89 &0.13 &2382.83  &0.19  &798.75 &0.28\\
Trees &396.46 &0.00&5960.50  &0.00 &17705.65 &0.07 &6323.89 &0.13 &3552.26  &0.18  &1021.86 & 0.28 \\
LMART&327.42 &0.00&5928.39  &0.00 &28178.68 &0.03 &7501.61 &0.13 & 3472.22 &0.18  &893.93 &0.28 \\
tMDP+DFS&2324.27 &0.00&35037.37  &0.02& 34062.76 &0.04 &11145.84&0.17&5365.54&0.22&1519.11&0.31\\
TreeGate&1242.64 &0.00&12826.16  &0.004  &19133.68 &0.03 &6901.62 &0.15 &3144.67  &0.20  &568.44 &0.30  \\
GNN &259.61 &0.00&3913.11  &0.00 &28926.49 &0.05 &14676.73 &0.11 &7162.61  &0.16  &1120.95 & 0.27  \\
AdaSolver-RL&2227.30 &0.00&35936.70  & 0.02 & 32528.50&0.12 &15368.45 &  0.17& 8476.60 & 0.21& 1653.05&0.32 \\
AdaSolver-IL &270.69 &0.00&4361.15  &0.00 &40711.45 &0.05 &30649.53 &0.10 &19573.54  &0.14  &2367.05 &0.27  \\\bottomrule

    \\[-7pt]
    & \multicolumn{12}{ c }{Set Covering} \\
    \\[-3pt]
\end{tabular}
}
\newline
\vspace{1mm}
\newline
\resizebox{\textwidth}{!}{\begin{tabular}{@{}ccccccccccccc@{}}
\toprule
\toprule
& \multicolumn{2}{c}{$D_1$ } & \multicolumn{2}{c}{$D_2$ } & \multicolumn{2}{c}{$D_3$ } & \multicolumn{2}{c}{$D_4$ }& \multicolumn{2}{c}{$D_5$}& \multicolumn{2}{c}{$D_6$}\\ \midrule
Method &  Nodes & PD gap $\downarrow$ &  Nodes & PD gap $\downarrow$&  Nodes& PD gap $\downarrow$&  Nodes & PD gap $\downarrow$ &  Nodes & PD gap $\downarrow$&  Nodes& PD gap $\downarrow$ \\\cmidrule(r){1-3} \cmidrule(r){4-5} \cmidrule(lr){6-7}\cmidrule(lr){8-9}\cmidrule(lr){10-11}\cmidrule(lr){12-13}
RPB &42.00 &0.00&1616.12  &0.00 &14154.73 &0.00&24072.15 &0.01 &15178.22  &0.02  &9185.16 &0.02  \\
\cmidrule(r){1-3} \cmidrule(r){4-5} \cmidrule(lr){6-7}\cmidrule(lr){8-9}\cmidrule(lr){10-11}\cmidrule(lr){12-13}
SVMRank &93.02 &0.00&1604.26  &0.00 &11150.03 &0.004  &9843.47 &0.01 &7564.52  &0.02  &6019.93 &0.02\\
Trees &102.49 &0.00&3487.29  &0.00 &29083.64 &0.005 &16295.54 & 0.01 &12398.58  &0.02  &10584.14 &0.02  \\
LMART&87.58 &0.00&1587.87  &0.00 &14109.08 &0.003 &15478.90 &0.01 &11426.58  & 0.02 &8965.19 &0.02 \\
tMDP+DFS&141.69 &0.00&3290.90  &0.00 &27826.58 &0.003 &47398.77 &0.01 &28752.17  &0.02  &16827.58 &0.02 \\
TreeGate&527.47&0.00&4473.12&0.00&26772.53&0.003&21841.75&0.02&15610.67&0.02&6948.79& 0.03 \\
GNN &84.99&0.00&1306.10&0.00&11710.30&0.00&32724.97&0.009&21995.83&0.01&15017.98&0.02  \\
AdaSolver-RL&191.85 &0.00& 2308.95 & 0.00&31815.45 & 0.00&58011.39 &0.01 & 38012.95 & 0.02 & 22258.45&0.02  \\
AdaSolver-IL &102.71&0.00&1953.57&0.00&39556.67&0.00&36135.36&0.008&27599.51&0.01&17599.67&0.01 \\\bottomrule

    \\[-7pt]
    & \multicolumn{12}{ c }{Combinatorial Auctions}  \\
    \\[-3pt]
\end{tabular}
}
\newline
\vspace{1mm}
\newline
\resizebox{\textwidth}{!}{\begin{tabular}{@{}ccccccccccccc@{}}
\toprule
\toprule
& \multicolumn{2}{c}{$D_1$ } & \multicolumn{2}{c}{$D_2$ } & \multicolumn{2}{c}{$D_3$ } & \multicolumn{2}{c}{$D_4$ }& \multicolumn{2}{c}{$D_5$}& \multicolumn{2}{c}{$D_6$}\\ \midrule
Method &  Nodes & PD gap $\downarrow$ &  Nodes & PD gap $\downarrow$&  Nodes& PD gap $\downarrow$&  Nodes & PD gap $\downarrow$ &  Nodes & PD gap $\downarrow$&  Nodes& PD gap $\downarrow$ \\\cmidrule(r){1-3} \cmidrule(r){4-5} \cmidrule(lr){6-7}\cmidrule(lr){8-9}\cmidrule(lr){10-11}\cmidrule(lr){12-13}
RPB &387.31 &0.00&413.93  &0.00 &145.20 &0.002 &11.78 &0.002 &2.38  &0.02  &1.20 &0.07  \\
\cmidrule(r){1-3} \cmidrule(r){4-5} \cmidrule(lr){6-7}\cmidrule(lr){8-9}\cmidrule(lr){10-11}\cmidrule(lr){12-13}
SVMRank &491.71 &0.00&533.21  &0.00 &337.59 &0.002  &34.63 &0.004 &10.46  &0.02  &2.46 &0.09\\
Trees &508.60 &0.00&588.19  &0.00 &317.75 &0.002 &31.00 &0.004 &8.22  &0.02  &2.78 &0.09  \\
LMART&445.69 &0.00&520.89  &0.00 &338.38 &0.002 &34.79 &0.004 &10.51  &0.02  &2.52 &0.09 \\
tMDP+DFS&723.95 &0.00&788.96  &0.00 &361.68 &0.002 &42.21 &0.002 &7.68  &0.02  &2.89 & 0.09\\
TreeGate&584.64 &0.00&631.03  &0.00 &299.50 & 0.003&46.27 &0.004   &2.93  &0.007  &1.44& 0.05 \\
GNN &458.11 &0.00&592.82  &0.00 &290.25 &0.002 &73.76 &0.003  &16.22  &0.005   &3.17 & 0.07  \\
AdaSolver-RL &1024.40 & 0.00& 839.95 &0.00&392.70 &0.002 &116.00 & 0.004 & 36.08 & 0.004&  8.13& 0.04\\
AdaSolver-IL &529.61 &0.00&659.79  &0.00 &487.49 &0.002 &190.46 &0.002 &59.84  &0.003  &3.72 &0.08  \\\bottomrule

    \\[-7pt]
    & \multicolumn{12}{ c }{Capacitated Facility Location} \\
    \\[-3pt]
\end{tabular}
}
\newline
\vspace{1mm}
\newline
\resizebox{\textwidth}{!}{\begin{tabular}{@{}ccccccccccccc@{}}
\toprule
\toprule
& \multicolumn{2}{c}{$D_1$ } & \multicolumn{2}{c}{$D_2$ } & \multicolumn{2}{c}{$D_3$ } & \multicolumn{2}{c}{$D_4$ }& \multicolumn{2}{c}{$D_5$}& \multicolumn{2}{c}{$D_6$}\\ \midrule
Method &  Nodes & PD gap $\downarrow$ &  Nodes & PD gap $\downarrow$&  Nodes& PD gap $\downarrow$&  Nodes & PD gap $\downarrow$ &  Nodes & PD gap $\downarrow$&  Nodes& PD gap $\downarrow$ \\\cmidrule(r){1-3} \cmidrule(r){4-5} \cmidrule(lr){6-7}\cmidrule(lr){8-9}\cmidrule(lr){10-11}\cmidrule(lr){12-13}
RPB &594.46 &0.00&16097.93  &0.00 &17730.83 &0.03 &900.00 &5387.45 &900.00  &1874.07  &900.00 &854.44  \\
\cmidrule(r){1-3} \cmidrule(r){4-5} \cmidrule(lr){6-7}\cmidrule(lr){8-9}\cmidrule(lr){10-11}\cmidrule(lr){12-13}
SVMRank &289.60 &0.00&5757.21  &0.01  &5097.74 &0.02 &1536.43 &0.03 &909.91  & 0.04 &507.74 &0.05  \\
Trees &653.03 &0.00&13349.63  & 0.004 &7205.15 &0.02 &2037.52 &0.03 &1444.09  &0.04  &912.21 &0.05\\
LMART&425.62 &0.00&12268.72  &0.005  &8108.65 &0.02 &1862.61 &0.03 &1084.89  & 0.04 &605.03 &0.05 \\
tMDP+DFS&459.08 &0.00&21183.04  &0.003 &23298.80 &0.02 &7350.35 &0.03 &3615.30  &0.04  &1647.60 &0.05 \\
TreeGate&2136.85 &0.003 &8585.86  &0.01  &3516.26 &0.01 &1198.28 & 0.04 &925.65  &0.05  &999.87 &0.05 \\
GNN &285.58 &0.00&10623.60  &0.001  &16520.12 &0.02 &3809.20 &0.03 &1392.95  &0.04  &679.15 & 0.04  \\
AdaSolver-RL & 351.50&  0.00 & 22448.85 &0.00& 30533.75& 0.01 &12887.25 &0.03  & 7027.05 &0.04 & 4016.80 &0.05\\
AdaSolver-IL &241.12 &0.00&10664.04  &0.004& 20861.74  &0.01 &19596.78 &0.03 &12479.65  &0.04  &6548.56 & 0.05 \\\bottomrule

    \\[-7pt]
    & \multicolumn{12}{ c }{Maximum Independent Set} \\
    \\[-3pt]
\end{tabular}
}
\end{center}

\vspace{-12mm}
\end{table*}

\subsection{More Results of Ablation Studies}
We compare the Nodes and PD gap metrics in the ablation studies.
The results in Table \ref{table:ablation_other_metrics} show that AdaSolver outperforms RM in terms of solving time and PD gap, demonstrating the effectiveness of our proposed adversarial instance augmentation policy.

\begin{table*}[h]
\caption{More results on the ablation studies in terms of Nodes and PD gap. AdaSolver-IL and AdaSolver-RL improve the PD gap of the GNN and tMDP+DFS baselines. Though slightly more explored nodes in $D_1$ and $D_2$, AdaSolver-IL and AdaSolver-RL still achieve a shorter solving time.}
\small
\vspace{-1mm}
\label{table:ablation_other_metrics}
\centering
\begin{center}
\resizebox{\textwidth}{!}{
\begin{tabular}{@{}ccccccccccccc@{}}
\toprule
\toprule
& \multicolumn{2}{c}{$D_1$ } & \multicolumn{2}{c}{$D_2$ } & \multicolumn{2}{c}{$D_3$ } & \multicolumn{2}{c}{$D_4$ }& \multicolumn{2}{c}{$D_5$}& \multicolumn{2}{c}{$D_6$}\\ \midrule
Method &  Nodes & PD gap $\downarrow$ &  Nodes & PD gap $\downarrow$& Nodes & PD gap $\downarrow$& Nodes & PD gap $\downarrow$ & Nodes & PD gap $\downarrow$& Nodes& PD gap $\downarrow$\\\cmidrule(r){1-3} \cmidrule(r){4-5} \cmidrule(lr){6-7}\cmidrule(lr){8-9}\cmidrule(lr){10-11}\cmidrule(lr){12-13}
RM-IL & 269.83 & 0.00& 5676.33  & 0.00& 44928.63  & 0.06 & 14929.49 & 0.18 & 8362.64 & 0.23 & 3301.91 &  0.31 \\
AdaSolver-IL &270.69 &0.00&4361.15  &0.00 &40711.45 &0.05 &30649.53 &0.10 &19573.54  &0.14  &2367.05 &0.27\\\midrule
RM-RL& 2227.10 & 0.00& 36310.10 & 0.02 & 31287.20 & 0.12 & 10865.95 &  0.18& 5084.90 & 0.22 & 815.05 & 0.33\\
AdaSolver-RL&2227.30 &0.00&35936.70  & 0.02 & 32528.50&0.12 &15368.45 &  0.17& 8476.60 & 0.21& 1653.05&0.32 \\\bottomrule

    \\[-7pt]
    & \multicolumn{12}{ c }{Set Covering} \\
    \\[-3pt]
\end{tabular}
}
\newline
\vspace{-1mm}
\newline
\resizebox{\textwidth}{!}{\begin{tabular}{@{}ccccccccccccc@{}}
\toprule
\toprule
& \multicolumn{2}{c}{$D_1$ } & \multicolumn{2}{c}{$D_2$ } & \multicolumn{2}{c}{$D_3$ } & \multicolumn{2}{c}{$D_4$ }& \multicolumn{2}{c}{$D_5$}& \multicolumn{2}{c}{$D_6$}\\ \midrule
Method &  Nodes & PD gap $\downarrow$ &  Nodes & PD gap $\downarrow$& Nodes$\downarrow$ & PD gap $\downarrow$& Nodes & PD gap $\downarrow$ & Nodes & PD gap $\downarrow$& Nodes& PD gap $\downarrow$\\\cmidrule(r){1-3} \cmidrule(r){4-5} \cmidrule(lr){6-7}\cmidrule(lr){8-9}\cmidrule(lr){10-11}\cmidrule(lr){12-13}
RM-IL & 101.24 & 0.00& 2491.01 &0.00& 43006.75 & 0.00& 29486.44 & 0.009 & 21203.95  & 0.01 & 14441.14 &  0.02 \\
AdaSolver-IL &102.71&0.00&1953.57&0.00&39556.67&0.00&36135.36&0.008&27599.51&0.01&17599.67&0.01 \\\midrule
RM-RL& 199.05 &  0.00 & 2237.20 & 0.00& 30545.60 &0.00& 58141.90 & 0.01   &31930.00 &0.02 &15079.60&0.03\\
AdaSolver-RL&191.85 &0.00& 2308.95 & 0.00&31815.45 & 0.00&58011.39 &0.01 & 38012.95 & 0.02 & 22258.45&0.02  \\\bottomrule

    \\[-7pt]
    & \multicolumn{12}{ c }{Combinatorial Auctions}  \\
    \\[-3pt]
\end{tabular}
}
\newline
\vspace{-1mm}
\newline
\resizebox{\textwidth}{!}{\begin{tabular}{@{}ccccccccccccc@{}}
\toprule
\toprule
& \multicolumn{2}{c}{$D_1$ } & \multicolumn{2}{c}{$D_2$ } & \multicolumn{2}{c}{$D_3$ } & \multicolumn{2}{c}{$D_4$ }& \multicolumn{2}{c}{$D_5$}& \multicolumn{2}{c}{$D_6$}\\ \midrule
Method &  Nodes & PD gap $\downarrow$ &  Nodes & PD gap $\downarrow$&  Nodes& PD gap $\downarrow$& Nodes & PD gap $\downarrow$ & Nodes & PD gap $\downarrow$& Nodes& PD gap $\downarrow$\\\cmidrule(r){1-3} \cmidrule(r){4-5} \cmidrule(lr){6-7}\cmidrule(lr){8-9}\cmidrule(lr){10-11}\cmidrule(lr){12-13}
RM-IL & 473.79  & 0.00& 608.75 & 0.00& 438.14 & 0.002 &89.51 & 0.004 & 29.80 &0.005 & 9.52&0.05 \\
AdaSolver-IL &529.61 &0.00&659.79  &0.00 &487.49 &0.002 &190.46 &0.002 &59.84  &0.003  &3.72 &0.08 \\\midrule
RM-RL& 1018.60 & 0.00& 854.90 & 0.00& 468.90 & 0.002 &98.50 &0.003 &7.90  & 0.008 &2.15& 0.11 \\
AdaSolver-RL &1024.40 & 0.00& 839.95 &0.00&392.70 &0.002 &116.00 & 0.004 & 36.08 & 0.004&  8.13& 0.04\\\bottomrule

    \\[-7pt]
    & \multicolumn{12}{ c }{Capacitated Facility Location} \\
    \\[-3pt]
\end{tabular}
}
\newline
\vspace{-1mm}
\newline
\resizebox{\textwidth}{!}{\begin{tabular}{@{}ccccccccccccc@{}}
\toprule
\toprule
& \multicolumn{2}{c}{$D_1$ } & \multicolumn{2}{c}{$D_2$ } & \multicolumn{2}{c}{$D_3$ } & \multicolumn{2}{c}{$D_4$ }& \multicolumn{2}{c}{$D_5$}& \multicolumn{2}{c}{$D_6$}\\ \midrule
Method &  Nodes & PD gap $\downarrow$ &  Nodes & PD gap $\downarrow$& Nodes & PD gap $\downarrow$& Nodes & PD gap $\downarrow$ & Nodes & PD gap $\downarrow$& Nodes& PD gap $\downarrow$\\\cmidrule(r){1-3} \cmidrule(r){4-5} \cmidrule(lr){6-7}\cmidrule(lr){8-9}\cmidrule(lr){10-11}\cmidrule(lr){12-13}
RM-IL & 226.79 & 0.00& 40493.42 & 0.00& 31616.33 & 0.01 &14390.99 & 0.03 & 7323.71 & 0.04 & 4721.40 & 0.05  \\
AdaSolver-IL &241.12 &0.00&10664.04  &0.004& 20861.74  &0.01 &19596.78 &0.03 &12479.65  &0.04  &6548.56 & 0.05\\\midrule
RM-RL& 323.60 & 0.00& 23020.05 &0.007  & 27906.65 & 0.01 &9073.45 &0.03 & 2974.65 &0.04  &1434.90 & 0.05\\
AdaSolver-RL & 351.50&  0.00 & 22448.85 &0.00& 30533.75& 0.01 &12887.25 &0.03  & 7027.05 &0.04 & 4016.80 &0.05\\\bottomrule

    \\[-7pt]
    & \multicolumn{12}{ c }{Maximum Independent Set} \\
    \\[-3pt]
\end{tabular}
}
\end{center}

\vspace{-6mm}
\end{table*}

%
%
%




\subsection{More Results on the Sample Efficiency}
We compare the Nodes and PD gap metrics in the experiments of Section \ref{subsection:sample_efficiency}.
The results in Table \ref{table:more_sample_efficiency_under_less_data} show that AdaSolver significantly outperforms the underlying baselines in terms of PD gap with only 1\% of training instances.
AdaSolver-IL consistently outperforms the GNN baseline in terms of the PD gap.
Additionally, in the distributions where the solvers can solve the instances within the time limit, AdaSolver-IL explores significantly fewer nodes than the GNN baseline.

\begin{table*}[h!]
\caption{Results of the sample efficiency experiments in terms of the Nodes and PD gap metrics. \textit{Method ($n$ instances)} means that the \textit{Method} is trained on $n$ instances. The GNN baseline faces significant performance degradation when the training instances are few (here we use 1\% of the training instances) in terms of Nodes and PD gap. AdaSolver achieves a significant improvement in Nodes and PD gap with only 1\% of the training instances, demonstrating a significant improvement in sample efficiency.}
\small{
\vspace{-1mm}
\label{table:more_sample_efficiency_under_less_data}
\centering
\resizebox{\textwidth}{!}{
\begin{tabular}{@{}ccccccccccccc@{}}
\toprule
\toprule
Method &  Nodes & PD gap $\downarrow$ &  Nodes & PD gap $\downarrow$&  Nodes& PD gap $\downarrow$&  Nodes & PD gap $\downarrow$ &  Nodes & PD gap $\downarrow$&  Nodes& PD gap $\downarrow$\\\cmidrule(r){1-3} \cmidrule(r){4-5} \cmidrule(lr){6-7}\cmidrule(lr){8-9}\cmidrule(lr){10-11}\cmidrule(lr){12-13}
GNN (10000 instances)&259.61 &0.00&3913.11  &0.00 &28926.49 &0.05 &14676.73 &0.11 &7162.61  &0.16  &1120.95 & 0.27  \\
GNN (100 instances)& 769.60 & 0.00& 15754.15 &0.00& 37577.55 & 0.10 & 26173.75 & 0.15 & 18570.70 &0.20& 4854.60 &0.30\\
AdaSolver-IL (100 instances)&445.25&0.00&10010.15&0.00&49802.45&0.08&38256.80&0.13&28539.60&0.17&7270.45&0.28\\
 \bottomrule
\\[-7pt]
    & \multicolumn{12}{ c }{Set Covering} \\
    \\[-3pt]
\end{tabular}
}
\newline
\vspace{1mm}
\newline
\resizebox{\textwidth}{!}{
\begin{tabular}{@{}ccccccccccccc@{}}
\toprule
\toprule
Method &  Nodes & PD gap $\downarrow$ &  Nodes & PD gap $\downarrow$&  Nodes& PD gap $\downarrow$&  Nodes & PD gap $\downarrow$ &  Nodes & PD gap $\downarrow$&  Nodes& PD gap $\downarrow$\\\cmidrule(r){1-3} \cmidrule(r){4-5} \cmidrule(lr){6-7}\cmidrule(lr){8-9}\cmidrule(lr){10-11}\cmidrule(lr){12-13}
GNN (10000 instances)&84.99&0.00&1306.10&0.00&11710.30&0.00&32724.97&0.009&21995.83&0.01&15017.98&0.02  \\
GNN (100 instances)&217.50 &0.00&3532.60 &0.00&34269.20 &0.004&33319.85 &0.01&23282.75 &0.02&16594.65 &0.03\\
AdaSolver-IL (100 instances)& 107.93 &0.00& 1714.28 &0.00& 14221.71 &0.00& 31613.71 &0.01& 23753.37 &0.01& 16616.03 &0.02\\
\bottomrule
\\[-7pt]
    & \multicolumn{12}{ c }{Combinatorial Auctions}  \\
    \\[-3pt]
\end{tabular}
}
}
\end{table*}

\section{Analysis of the GNN Embeddings in the IL Setting}
\begin{figure}[t]
    \centering
    \begin{subfigure}{0.15\textwidth}
        \includegraphics[width=\textwidth]{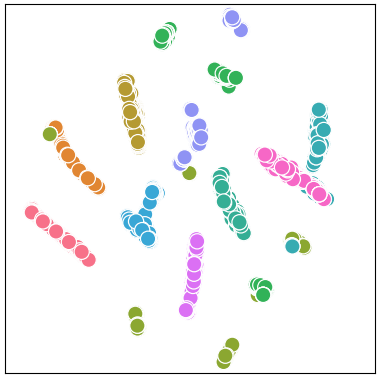}
        \vspace{-5mm}
        \caption{Before training.}
        \label{fig:embeddings_before}
    \end{subfigure}
    \begin{subfigure}{0.15\textwidth}
        \includegraphics[width=\textwidth]{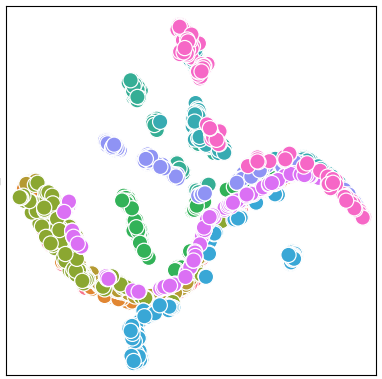}
        \vspace{-5mm}
        \caption{After training.}
        \label{fig:embeddings_after}
    \end{subfigure}
    \begin{subfigure}{0.15\textwidth}
        \includegraphics[width=\textwidth]{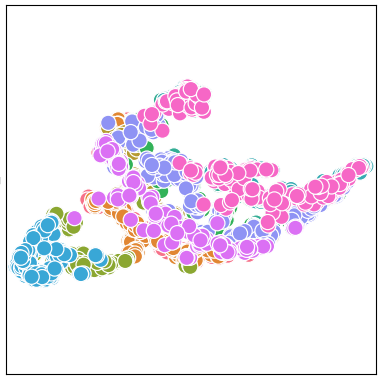}
        \vspace{-5mm}
        \caption{AdaSolver-IL.}
        \label{fig:embeddings_ReSolver}
    \end{subfigure}
    \begin{subfigure}{0.15\textwidth}
        \includegraphics[width=\textwidth]{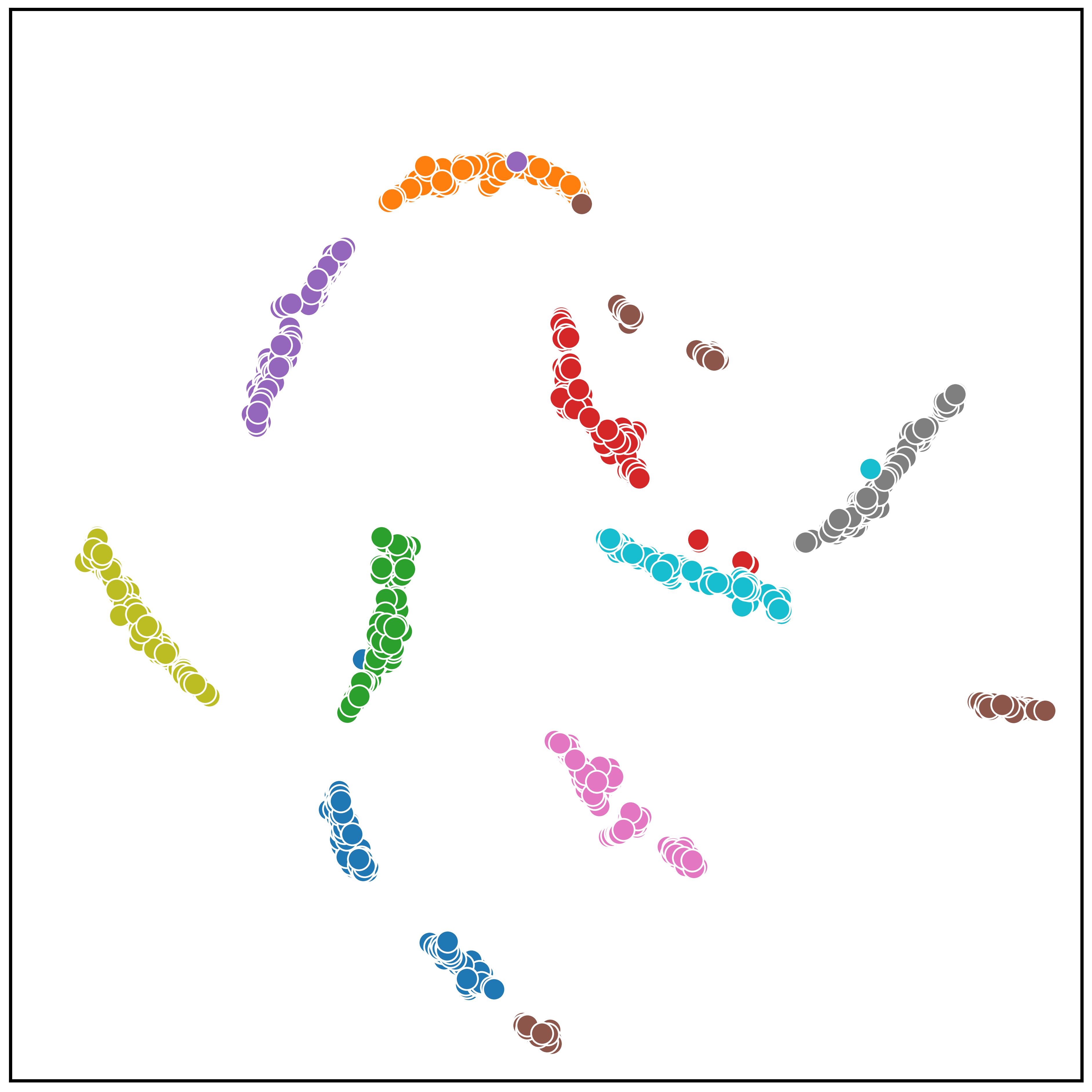}
        \vspace{-5mm}
        \caption{Before training.}
        \label{fig:embeddings_before2}
    \end{subfigure}
    \begin{subfigure}{0.15\textwidth}
        \includegraphics[width=\textwidth]{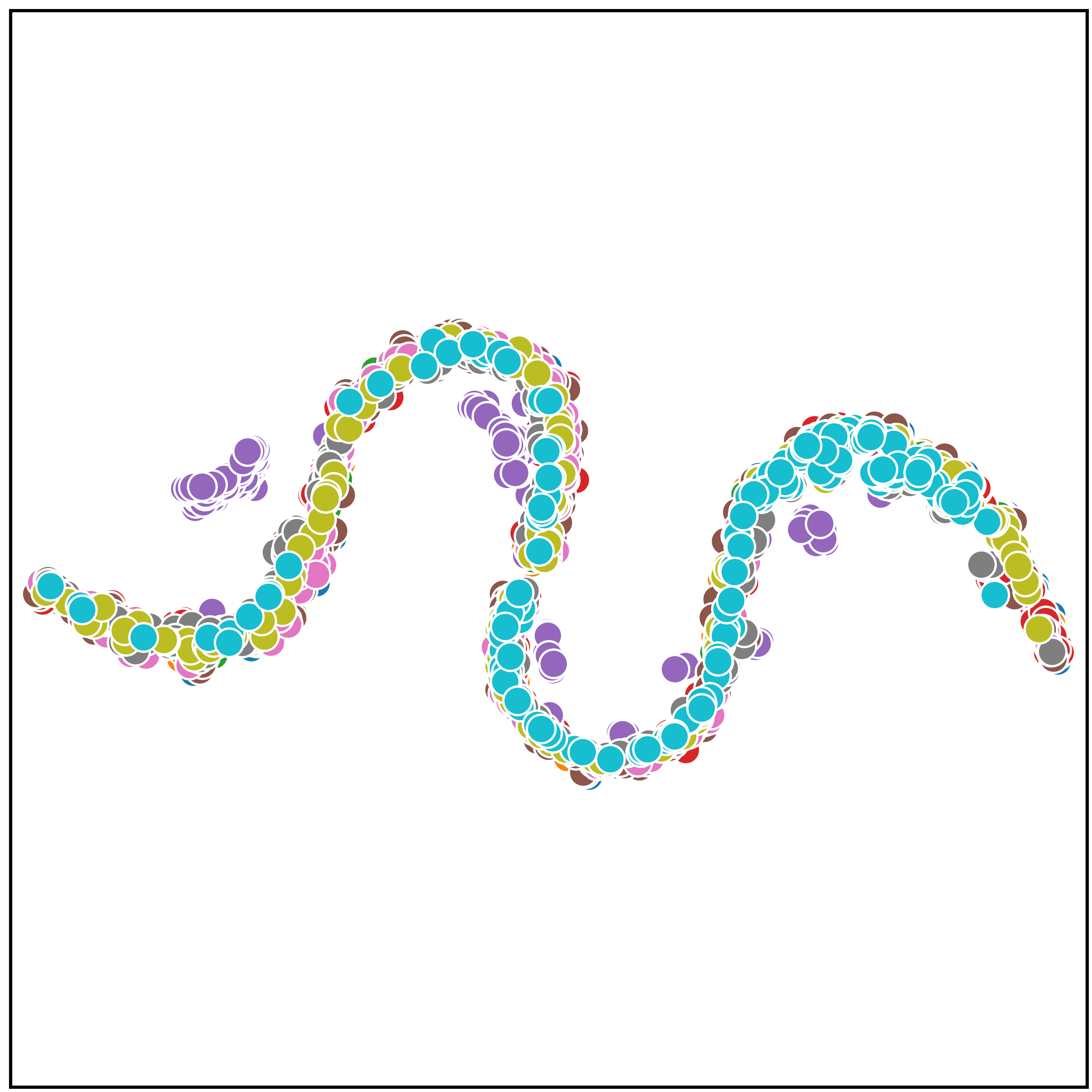}
        \vspace{-5mm}
        \caption{After training.}
        \label{fig:embeddings_after2}
    \end{subfigure}
    \begin{subfigure}{0.15\textwidth}
        \includegraphics[width=\textwidth]{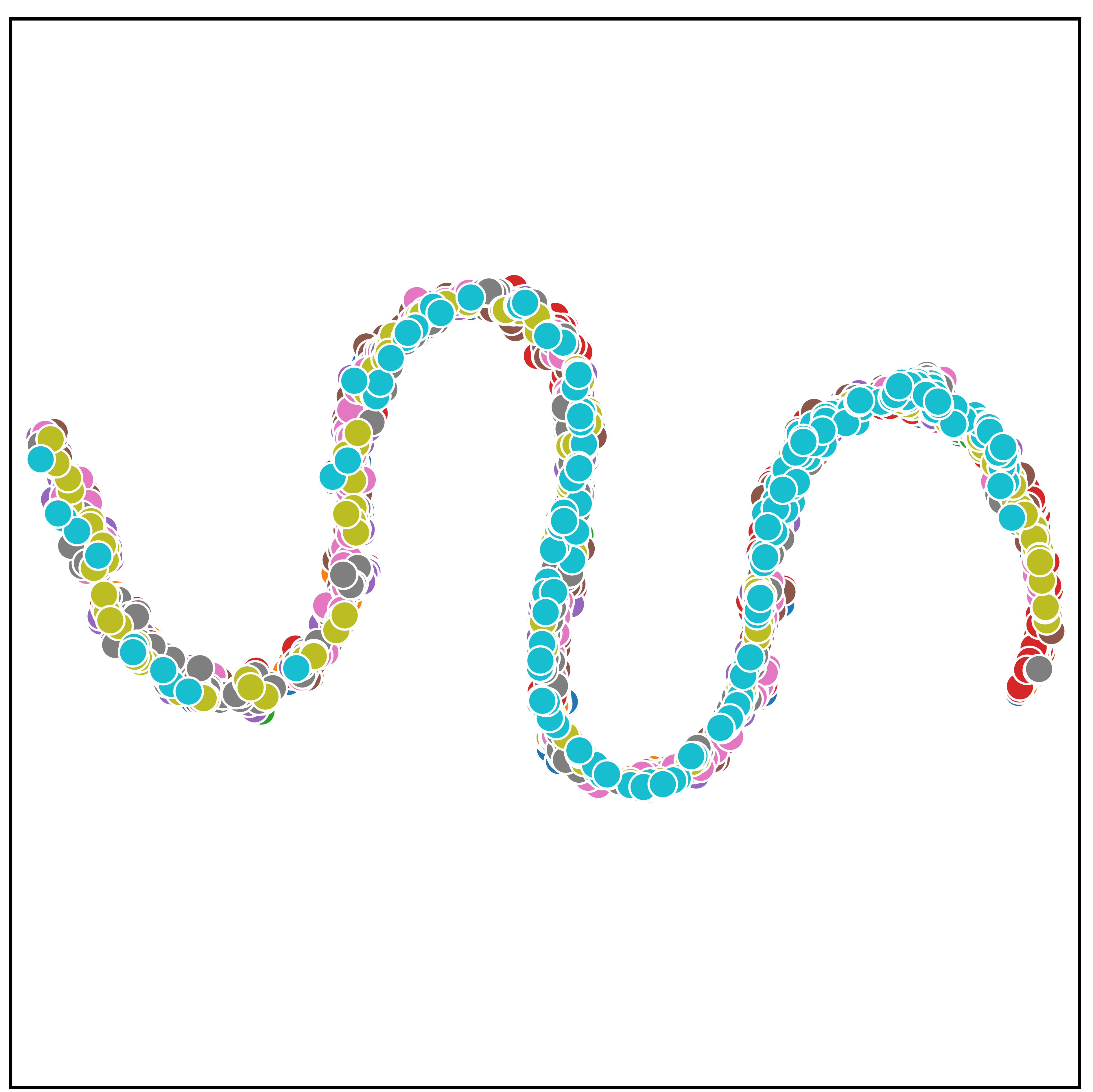}
        \vspace{-5mm}
        \caption{AdaSolver-IL.}
        \label{fig:embeddings_ReSolver2}
    \end{subfigure}
    \caption{T-SNE visualization for embeddings of GNN branching policy. Different colors represent the samples from different distributions. The first row is the results on the combinatorial auctions and the second row is on the set covering benchmark. In each row, the left figure shows the embeddings of the GNN baseline before training, and the middle figure shows those after training. The right figure is the embeddings of AdaSolver-IL, which are more concentrated and stable. }
    \label{embedding}
    \vspace{-4mm}
\end{figure}

We use T-SNE \cite{van2008visualizing} to visualize the learned embeddings of the GNN branching policy with training distribution $D_1$. 
The results on the set covering and combinatorial auctions benchmark are respectively illustrated in the first and second row of Figure \ref{embedding}, where different colors represent samples from various distributions on each benchmark.
In order to investigate the robustness of environmental perturbations, we select ten distributions with different instance sizes and nonzero entry proportions of constraint matrices (please see Appendix \ref{data:vis} for the ten distributions).
As a result, the bipartite graphs of the instances across distributions exhibit varying local graph structures.

We first observe that before training, the embeddings of the GNN baseline on different distributions are separated from each other without overlap (Figure \ref{fig:embeddings_before} and \ref{fig:embeddings_before2})
This indicates that the GNN model fails to extract shared task-relevant features across distributions within the same problem family.
Consequently, the predicted scores using these embeddings become unstable across distributions, resulting in poor generalization ability.
We observe in Figure \ref{fig:embeddings_after} and  \ref{fig:embeddings_after2} that while the embeddings of the GNN baseline exhibit less separation after imitation training, they still appear scattered.
In contrast, our proposed AdaSolver-IL yields more concentrated and stable embeddings, as shown in Figure \ref{fig:embeddings_ReSolver} and \ref{fig:embeddings_ReSolver2}. 
This observation provides evidence that AdaSolver is less sensitive to distributional shifts and perturbations, demonstrating its robustness in handling such changes.

\end{document}